\documentclass[10pt]{article}
\usepackage{latexsym,amsmath,amssymb,graphics,amscd, empheq}
\usepackage{multirow}
\usepackage{soul}
\usepackage{booktabs}
\usepackage{tabularx}
\usepackage[hang,footnotesize]{caption}
\usepackage[pdftex]{graphicx}
\usepackage{graphicx}
\usepackage{overpic}
\usepackage{color}
\usepackage{cancel}
\usepackage[breaklinks,colorlinks,
  citecolor=blue,
  urlcolor=blue]{hyperref}

\addtocounter{footnote}{1}
\makeatletter
\@addtoreset{table}{bsection}
\def\thetable{\thesection.\@arabic\c@table}
\def\fps@table{h, t}
\@addtoreset{equation}{section}

\makeatother

\newtheorem{theorem}{Theorem}[section]

\newtheorem{lemma}[theorem]{Lemma}
\newtheorem{remark}[theorem]{Remark}
\newtheorem{proposition}[theorem]{Proposition}

\newtheorem{corollary}[theorem]{Corollary}

\newtheorem{examples}[theorem]{Examples}

\newcommand{\bfi}{\bfseries\itshape}
\newcommand{\vertiii}[1]{{\left\vert\kern-0.25ex\left\vert\kern-0.25ex\left\vert #1 
    \right\vert\kern-0.25ex\right\vert\kern-0.25ex\right\vert}}
\newsavebox{\savepar}

\makeatletter

\usepackage{scalerel,stackengine}
\stackMath
\newcommand\reallywidehat[1]{%
\savestack{\tmpbox}{\stretchto{%
  \scaleto{%
    \scalerel*[\widthof{\ensuremath{#1}}]{\kern-.6pt\bigwedge\kern-.6pt}%
    {\rule[-\textheight/2]{1ex}{\textheight}}
  }{\textheight}%
}{0.5ex}}%
\stackon[1pt]{#1}{\tmpbox}%
}

\begin{document}

\title{\textbf{Differentiable reservoir computing}}
\author{Lyudmila Grigoryeva$^{1}$ and Juan-Pablo Ortega$^{2, 3}$}
\date{}
\maketitle

\begin{abstract}
Much effort has been devoted in the last two decades to characterize the situations in which a reservoir computing system exhibits the so-called echo state (ESP) and fading memory (FMP) properties. These important features amount, in mathematical terms, to the existence and continuity of global reservoir system solutions. That research is complemented in this paper with the characterization of the  differentiability of reservoir filters for very general classes of discrete-time deterministic inputs. This constitutes a novel strong contribution to the long line of research on the ESP and the FMP and, in particular, links to existing research on the input-dependence of the ESP. Differentiability has been shown in the literature to be a key feature in the learning of attractors of chaotic dynamical systems.
A Volterra-type series representation for reservoir filters with semi-infinite discrete-time inputs is constructed in the analytic case using Taylor's theorem and  corresponding approximation bounds are provided. Finally, it is shown as a corollary of these results that any fading memory filter can be uniformly approximated by a finite Volterra series with finite memory.
\end{abstract}

\bigskip


\makeatletter
\addtocounter{footnote}{1} \footnotetext{%
Department of Mathematics and Statistics. Universit\"at Konstanz. Box 146. D-78457 Konstanz. Germany. {\texttt{Lyudmila.Grigoryeva@uni-konstanz.de} }}
\addtocounter{footnote}{1} \footnotetext{%
Universit\"at Sankt Gallen. Faculty of Mathematics and Statistics. Bodanstrasse 6.
CH-9000 Sankt Gallen. Switzerland. {\texttt{Juan-Pablo.Ortega@unisg.ch}}}
\addtocounter{footnote}{1} \footnotetext{%
Centre National de la Recherche Scientifique (CNRS). France. }
\addtocounter{footnote}{-3}
\makeatother

\medskip

\medskip

\medskip

\section{Introduction}

\paragraph{Context and preliminary discussion.}

Reservoir computing (RC) is a neural approach to the learning of dynamic processes which advocates the use of paradigms in which the supervised estimation of all available interconnection weights is not necessary and only the training of a static memoryless readout suffices to obtain good performances. This computational strategy has been simultaneously inspired by ideas coming from three different fields, namely, recurrent neural networks, dynamical systems, and biologically inspired neural microcircuits. The common thread to these analyses is the use of rich dynamics to process information and to create memory traces. This explains why RC it can be found in the literature under other denominations like {\bfi  Liquid State Machines}~\cite{Maass2000, maass1, Natschlager:117806, corticalMaass, MaassUniversality} and is represented by various learning paradigms, being the  {\bfi  Echo State Networks} introduced in~\cite{jaeger2001, Jaeger04} a particularly important example.

RC has shown superior performance in many forecasting and classification engineering tasks (see \cite{lukosevicius} and references therein) and has shown unprecedented abilities in the learning of the attractors of complex nonlinear infinite dimensional dynamical systems \cite{Jaeger04, pathak:chaos, Pathak:PRL, Ott2018}. Additionally, RC implementations with dedicated hardware have been designed and built (see, for instance,~\cite{Appeltant2011, Rodan2011, SOASforRC, Larger2012, Paquot2012, photonicReservoir2013, swirl:paper, Vinckier2015, Laporte2018}) that exhibit information processing speeds that largely outperform standard Turing-type computers.

Ever since the inception of this methodology, much effort has been devoted to identify the  features that make a RC system capable of retaining relevant memory traces of  the inputs and computationally powerful. The first question has given rise to various notions and computational schemes for the memory capacity of RC systems \cite{Jaeger:2002, White2004, Ganguli2008, Hermans2010, dambre2012, GHLO2014_capacity, linearESN, RC3, tino:symmetric}. Another strand of interesting literature that we will not explore in this work has to do with the Turing computability capabilities of the systems of the type that we just introduced; recent relevant works in this direction are \cite{kilian:1996, siegelmann:1997, cabessa:2015, cabessa:2016}, and references therein.

Regarding computational power, there are three properties that pervade the literature and that are usually declared as necessary to obtain an adequate functioning in a RC system (see, for instance, \cite{DynamicalSystemsMaass, lukosevicius, maass2} and references therein), namely, the {\bfi  fading memory property (FMP)}, the {\bfi  echo state property (ESP)}, and the pairwise {\bfi  separation property (SP)}. The FMP is a notion observed in many modeling situations in which the influence of the input gradually fades out in time. This property is repeatedly invoked in systems theory \cite{volterra:book, wiener:book}, computational neurosciences \cite{corticalMaass}, physics \cite{Coleman1968}, or mechanics (see \cite{Fabrizio2010} and references therein). The ESP \cite{jaeger2001, Yildiz2012, Manjunath:Jaeger} is an existence and uniqueness property for the solutions of a state-space system that guarantees that the past history of the input fully determines the state of the system at any given point in time. Finally, the SP is satisfied by an input/output system if for any two input time series which differed in the past, the network assumes at subsequent time points different states.

Even though these three properties are an essential part of the ``RC jargon", it is not always clear in the literature why they are important. A partial answer to this question has been given in the development of universality theorems for RC machine learning paradigms. Indeed, it has been shown in~\cite{Maass2000, maass1, corticalMaass, MaassUniversality, RC6, RC7} that various families of RC systems that have these three properties are uniform universal approximants in a dynamical context in the presence of uniformly bounded (respectively, almost surely uniformly bounded) deterministic (respectively, stochastic) inputs. Moreover, these properties are exactly what is needed to prove universality statements using the Stone-Weierstrass theorem. Nevertheless, it has also been shown \cite{RC8} that when the uniform approximation criterion is replaced by a $L ^p $ norm defined with the measure induced by the input stochastic process, then the FMP does not play any role anymore.

Additionally, when these properties are invoked, it is not always clear what the actual definition that is being used is and they are even used exchangeably sometimes. The reason for this confusion is that, {\it in the presence of various compactness and contractivity hypotheses, the ESP and the FMP are automatically simultaneously satisfied}. Moreover, the same entanglement occurs when it comes to the actual dynamical implications that these properties entail like the input and state forgetting properties (see later on in the text for detailed definitions).

\paragraph{Important existing results.}
In order to make these remarks explicit, we recall here some results here that will help us later on to introduce the contributions in this paper. Consider the discrete-time nonlinear state-space transformation
\begin{empheq}[left={\empheqlbrace}]{align}
\mathbf{x} _t &=F(\mathbf{x}_{t-1}, {\bf z} _t),\label{reservoir equation}\\
{\bf y} _t &= h (\mathbf{x} _t). \label{readout}
\end{empheq}
In the context of supervised machine learning we will refer to these transformations as {\bfi  reservoir systems} and we will think of them as special types of recurrent neural networks. In that setup, the map $F: \mathbb{R} ^N\times \mathbb{R} ^n\longrightarrow  \mathbb{R} ^N$, $n,N \in \mathbb{N}^+ $, is called the 
{\bfi  reservoir}, it is usually randomly generated and  $h: \mathbb{R}^N \rightarrow \mathbb{R}^d$ is the {\bfi  readout}, which is estimated via a supervised learning procedure. The {\bfi  input} in this system is given by the elements of the infinite sequence
${\bf z}=(\ldots, {\bf z} _{-1}, {\bf z} _0, {\bf z} _1, \ldots) \in (\mathbb{R}^n) ^{\Bbb Z } $ and the {\bfi output} by the components of ${\bf y} \in (\mathbb{R} ^d)^{\Bbb Z } $. 

We say that the reservoir system \eqref{reservoir equation}-\eqref{readout} satisfies the {\bfi  echo state property (ESP)} when for any ${\bf z} \in (\mathbb{R}^n) ^{\Bbb Z } $ there exists a unique $\mathbf{y} \in (\mathbb{R} ^d)^{\Bbb Z } $ that satisfies \eqref{reservoir equation}. When this existence and uniqueness feature is available one can
associate well-defined filters $U ^F: (\mathbb{R}^n) ^{\Bbb Z } \longrightarrow (\mathbb{R}^N) ^{\Bbb Z }$ and $U ^F_h: (\mathbb{R}^n) ^{\Bbb Z } \longrightarrow ({\Bbb R}^d) ^{\mathbb{Z}_{-}}$ to the reservoir map $F$ and the reservoir system \eqref{reservoir equation}-\eqref{readout}, respectively.

Very general situations have been characterized in which the ESP holds. For example, suppose that we restrict ourselves to inputs that are uniformly bounded by a constant $M>0 $, that is, consider the space $K _M $  of semi-infinite sequences given by
\begin{equation}
\label{Kset}
K_{M}:=\left\{ {\bf z} \in \left({\Bbb R}^n\right)^{\mathbb{Z}_{-}} \mid \| {\bf z}_t\| \leq M \quad \mbox{for all} \quad t \in \Bbb Z _{-} \right\}, \quad \mbox{$M>0 $,}
\end{equation}
and assume that the reservoir map $F$ is continuous and a contraction on the first entry that maps $F: \overline{B_{\left\|\cdot \right\|}({\bf 0}, L)} \times \overline{B_{\left\|\cdot \right\|}({\bf 0}, M)} \longrightarrow \overline{B_{\left\|\cdot \right\|}({\bf 0}, L)} $, with $L >0 $ (the symbol $\overline{B_{\left\|\cdot \right\|}(\mathbf{v}, r)} $ denotes the closure of the open ball $B_{\left\|\cdot \right\|}(\mathbf{v}, r)$ with respect to a given norm $\left\|\cdot \right\| $, center $\mathbf{v}$, and radius $r>0$). In that case, it can be shown (see, for instance, \cite[Theorem 3.1]{RC7}) that for any ${\bf z} \in K _M  $ there exists a unique $\mathbf{x} \in K _L:=\{ {\bf x} \in \left({\Bbb R}^N \right)^{\mathbb{Z}_{-}} \mid \| {\bf x}_t\| \leq L \quad \mbox{for all} \quad t \in \Bbb Z _{-} \} $ that satisfies \eqref{reservoir equation}, that is, the ESP holds. This facts allows us
to associate unique filters $U ^F: K _M \longrightarrow K _L$ and $U ^F_h: K _M \longrightarrow ({\Bbb R}^d) ^{\mathbb{Z}_{-}}$ to the reservoir map $F$ and the reservoir system \eqref{reservoir equation}-\eqref{readout}, respectively. 

Moreover, in this situation (see again \cite[Theorem 3.1]{RC7}) the continuity of $F$ and $h$ implies that both $U ^F $ and $U ^F_h  $ are continuous when we consider either the uniform or the product topologies in the domain and target spaces. 
The continuity with respect to the product topology is called in this setup the {\bfi  fading memory property (FMP)} and, as we shall see below, can be characterized using weighted norms in the spaces of input and output sequences, which shows that recent inputs are more represented in the outputs of FMP filters than older ones. Equivalently, the outputs produced by FMP filters associated to inputs that are close in the recent past are close, even when those inputs may be very different in the distant past.

The restriction to uniformly bounded inputs of the type (\ref{Kset}) when using contracting reservoir maps does not only make the ESP and the FMP to simultaneously hold but it also simplifies enormously the characterization of the FMP. Indeed, it has been shown in \cite{sandberg:fmp, RC7} that in that case the fading memory property is not a metric but an exclusively topological property that does not depend on the weighted norm used to define it. Therefore, the FMP does not contain in that situation any information about the rate at which the dependence on the past inputs in the system output declines. This is not the case anymore when we consider unbounded input sets since, as we show later on in Theorem \ref{characterization of fmp unbounded}, {\it reservoir systems have the FMP only with respect to weighting sequences that converge to zero faster than the divergence rate of their outputs}.

\paragraph{Main contributions of the paper.}

The {\it core contributions of this paper are, first, the analysis of the ESP and the FMP in the absence of boundedness hypotheses and, second, the extension of the FMP-related continuity statements in the literature to the study of the differentiability properties of reservoir computers}. In particular, we aim at characterizing the situations in which one can obtain the differentiability of reservoir filters out of the  differentiability properties of the maps that define the corresponding reservoir system.

Regarding the first objective, there are several reasons to study reservoir computing systems with unbounded inputs. First, even though we only deal in this paper with the deterministic setup, any random component in the data generating process of the inputs, like a Gaussian perturbation, would imply unboundedness. Second, when dealing with reservoir systems associated to physical systems, it is certainly reasonable to assume boundedness in the input due to the saturation effects that most of those systems present. Nevertheless, the value of the bounding constant is in general unknown beforehand, which makes uniform boundedness hypotheses unrealistic. Finally, in the study of the differentiability properties of reservoir computers, the differentiability of Fr\'echet type is only defined on open subsets of normed spaces. We shall see that any open set in the Banach space of inputs with a weighted norm contains unbounded sequences, which forces us to deal with that situation.

As to the analysis of the differentiability properties of reservoir systems, this is an important question for several reasons: 

\begin{itemize}
\item The local nature of the differential allows the formulation of conditions that ensure both the local and global existence of differentiable and, in passing, fading memory solutions. These conditions are a novel strong contribution to the long line of research on the ESP and the FMP and, in particular, link to existing research \cite{Manjunath:Jaeger} on the input-dependence of the echo state property.
\item It has been recently shown \cite{Ott2018} how RC applications to the learning of the attractors of chaotic dynamical systems are much related with the notion of {\it Generalized Synchronization} \cite{kocarev1995general, kocarev1996generalized} for which differentiability has been shown to be a relevant feature \cite{hunt:ott:1997}. Indeed, in the absence of differentiability, the synchronization mapping may be ``wild" enough (in the terminology of \cite{hunt:ott:1997}) to create a gap between the information dimensions of the attractors of the input system and the system used to learn it.
\item The metric nature of the differential allows us to measure the speed at which fading memory filters forget inputs.  As we see later on in Theorem \ref{Differential uniform input forgetting property}, we are able to characterize this important piece of information with the differentiability property. 
\item When filters are analytic, they obviously admit a Taylor series expansion which coincides with the so called discrete-time Volterra series representation \cite{volterra:book, schetzen:book, rugh:book, priestley:tsbook} and, moreover, different Taylor remainders can be used to provide bounds on the approximation errors that are committed when those series are truncated. This path has been explicitly explored in \cite{sandberg:time-delay, sandberg:volterra} for analytic filters with respect to the supremum norm and with inputs with a finite past. We extend this work and we characterize the inputs for which an analytic fading memory reservoir   filter  with respect to a weighted norm admits a Volterra series representation with semi-infinite inputs. Additionally, we can use the causality and time-invariance hypotheses to show that the corresponding Volterra series representations have time-independent coefficients (this feature is not available in the case studied in \cite{sandberg:volterra}) that automatically satisfy the convergence conditions spelled out in \cite{sandberg:volterra:2, sandberg:volterra:3}. 
\item These statements can be combined with the results in \cite{RC6} to provide an alternative proof of the following Volterra series universality theorem that was stated for the first time in \cite[Theorems 3 and 4]{Boyd1985}:  {\it  any time-invariant and causal fading memory filter can be uniformly approximated by a finite Volterra series with finite memory}. 
\end{itemize}

\paragraph{Organization of the paper.}
The paper is organized as follows: 
\begin{itemize}
\item Section \ref{Differentiable reservoir computing systems} introduces the Banach sequence spaces where the semi-infinite inputs and  outputs of the reservoir systems that we study are defined. Various elementary facts about weighted and supremum norm topologies are stated, and the notions of fading memory, continuity, and differentiability of maps between sequence spaces are carefully introduced.
\item In Section \ref{Differentiable time-invariant filters and functionals} it is studied in detail the differentiability of causal and time-invariant filters defined on the sequence spaces introduced in Section \ref{Differentiable reservoir computing systems}. Those result  results  are put to work in Section \ref{The fading memory property and remote past input independence} to easily show well-known results that link the continuity of a filter with input and output spaces endowed with weighted norms with its asymptotic independence on the remote past input. A particular attention is paid in Section \ref{Equivalence of FMP and differentiability in filters and functionals}  to the relation between the FMP and the differentiability of causal and time-invariant filters with that of their associated functionals.
\item Starting from Section \ref{The fading memory property in reservoir filters with unbounded inputs} the paper focuses on reservoir filters. The main result in this section is Theorem \ref{characterization of fmp unbounded} that provides a sufficient (but not necessary) condition for the ESP and FMP to hold in the presence of inputs that are not necessarily bounded. This is a significant generalization with respect to the ``standard compactness conditions" imposed in \cite{jaeger2001} or the uniform boundedness in the inputs that was required in similar results in, for instance, \cite{RC7}. An important observation in Theorem \ref{characterization of fmp unbounded} is that for general inputs, the FMP depends on the weighting sequence that is used to define it and establishes that, roughly speaking, {\it reservoir systems have the FMP only with respect to weighting sequences that converge to zero faster than the divergence rate of their outputs}. This newly introduced FMP condition is spelled out for several widely used families of reservoir systems. The above mentioned results involving uniform boundedness hypotheses can be obtained as a corollary (see Corollary \ref{fmp when you restrict to bounded}) of the results in this section. Another statement that we prove (see Theorem \ref{ESP for reservoir maps with compact target}) is that when the target of the reservoir map is a compact set then the echo state property is in that situation guaranteed {\it for no matter what input},  even though the FMP may obviously not hold in that case.
\item Section \ref{Differentiability in reservoir filters} is the core of the paper and studies the differentiability properties of reservoir filters determined by differentiable reservoir maps. The main results are contained in Theorems \ref{Persistence of the ESP and FMP properties} and \ref{characterization of reservoir differentiability}. The first theorem provides an explicit and easy-to-verify sufficient condition for the ESP and the FMP  to hold around a given input for which we know that the reservoir system associated to a differentiable reservoir map has a solution. Theorem  \ref{characterization of reservoir differentiability} is a global extension of the previous result that, unlike Theorems \ref{characterization of fmp unbounded} and \ref{Persistence of the ESP and FMP properties}, fully characterizes the ESP and the differentiability (and hence the FMP) of the reservoir filter associated to a differentiable reservoir map. In Section \ref{The local versus the global echo state property} we show that the global conditions in Theorem \ref{characterization of reservoir differentiability} are much stronger than the local ones in Theorem \ref{Persistence of the ESP and FMP properties} by introducing an example that  shows how {\it the ESP and the FMP are structural features of a reservoir system when considered globally but are mostly input dependent when considered only locally}. This important observation has already been noticed in \cite{Manjunath:Jaeger}  where, using tools coming from the theory of non-autonomous dynamical systems,  sufficient conditions have been formulated (see, for instance, \cite[Theorem 2]{Manjunath:Jaeger}) that ensure the ESP in connection to a given specific input. The differentiability conditions that we impose to our reservoir systems allow us to draw similar conclusions and, additionally,  to automatically establish the FMP of the resulting locally defined reservoir filters. In Section \ref{Remote past input independence and the state forgetting property for unbounded inputs} we show how for globally differentiable reservoir filters we can formulate a non-uniform version of the well-known input forgetting property for FMP filters that we recovered in Section \ref{The fading memory property and remote past input independence} for inputs that are not necessarily bounded. Moreover, a novel uniform differential version of that result is provided in Theorem \ref{Differential uniform input forgetting property}.
\item Section \ref{The Volterra series representation of analytic filters and a universality theorem} contains two main results. First, Theorem \ref{Volterra series representation} shows the availability of discrete-time Volterra series representations for analytic, causal,  time-invariant, and FMP filters. This result extends a similar statement formulated in \cite{sandberg:time-delay, sandberg:volterra} to inputs with a semi-infinite past that are not necessarily bounded. Second, in Theorem \ref{Volterra series are universal}, we combine the previous result with a universality statement in \cite{RC6} to provide an alternative proof of the  Volterra series universality theorem stated for the first time in \cite[Theorems 3 and 4]{Boyd1985}. 
\end{itemize}
The proofs of most results are provided in the appendices at the end of the paper.

\section{The input and output spaces for reservoir systems}
\label{Differentiable reservoir computing systems}

This paper studies input/output systems that are causal, that is, the output depends only on the past history of the input and that, in general, have infinite memory. This makes us consider the spaces of left infinite sequences with values in $\mathbb{R}^n$, that is, $(\mathbb{R}^n) ^{\Bbb Z _ -}=\{{\bf z}=(\ldots, {\bf z} _{-2}, {\bf z} _{-1}, {\bf z} _0) \mid {\bf z} _i \in \mathbb{R}^n, i \in \mathbb{Z}_{-}\}$.  Analogously, $(D_n) ^{\Bbb Z _ -} $ stands for the space of semi-infinite sequences with elements in the subset $D_n\subset \mathbb{R}^n $. The space ${\Bbb R}^n  $ will be considered as a normed space with a norm denoted by $\left\|\cdot \right\|$  which is not necessarily the Euclidean one (even though they are all equivalent), unless it is explicitly mentioned.

We  endow these infinite product spaces with the Banach space structures associated to  one of the  following two norms. First, the {\bfi  supremum norm}  $\| {\bf z}\| _{\infty}:= {\rm sup}_{ t \in \Bbb Z_-} \left\{\| {\bf z} _t
\|\right\}$. The symbol $\ell_{-} ^{\infty}(\mathbb{R}^n) $ is used to denote the Banach space formed by the elements that have a finite supremum norm. Second, given a strictly decreasing sequence with zero limit  $w : \mathbb{N} \longrightarrow (0,1] $ and that $w _0=1 $, we define the  {\bfi  weighted norm} $\| \cdot \| _w $ on $(\mathbb{R}^n)^{\Bbb Z _{-}}$ associated to $w$ by 
$\| {\bf z} \| _w:= \sup_{t \in \Bbb Z_-}\{\| {\bf z}_t w_{-t}\|\}$. It can be shown (see \cite{RC7}) that the set $\ell_{-} ^{w}(\mathbb{R}^n) $ formed by the elements that have a finite $w$-weighted norm is a Banach space.
Moreover, it is easy to show that $\left\| {\bf z} \right\|_{w}\leq \left\| {\bf z} \right\|_{\infty} $, for all $\mathbf{z} \in (\mathbb{R}^n) ^{\Bbb Z _ -} $. This implies that   $\ell_{-} ^{\infty}(\mathbb{R}^n) \subset \ell ^{w}_-({\Bbb R}^n)$ and that the inclusion map $(\ell_{-} ^{\infty}(\mathbb{R}^n), \left\| \cdot \right\|_{\infty}) \hookrightarrow (\ell ^{w}_-({\Bbb R}^n), \left\| \cdot \right\|_{w})$ is continuous.

The Banach spaces $(\ell_{-} ^{\infty}(\mathbb{R}^n), \left\| \cdot \right\|_{\infty})  $  and $(\ell_{-} ^{w}(\mathbb{R}^n), \left\|\cdot \right\| _w) $ are particular cases of weighted Banach sequence spaces $(\ell_{-} ^{p , w}(\mathbb{R}^n), \left\| \cdot \right\|_{p , w})  $ where 
\begin{equation}
\label{definition lpw}
\left\|{\bf z}\right\|_{p, w}:= \left(\sum_{t \in \mathbb{Z}_{-}} \left\|{\bf z}_t\right\|^p w _{-t}\right)^{\frac{1}{p}}, \quad \mbox{with $1\leq p< +\infty $, ${\bf z} \in ({\Bbb R}^n) ^{\mathbb{Z}_-} $,  and $w $ a  sequence.}
\end{equation}
When $p = +\infty $ we set $\left\|\cdot \right\|_{p, w}:=  \left\|\cdot \right\|_ {w} $. We then define
\begin{equation}
\label{definition lpgeneral}
\ell_{-} ^{p , w}(\mathbb{R}^n):= \left\{{\bf z} \in {\Bbb R}^n\mid \left\|{\bf z}\right\|_{p, w}< +\infty\right\}.
\end{equation}
These spaces are defined in the literature (see, for instance, \cite{weightedls, lpasl2}) without the requirement that $w$ is a weighting sequence in the sense of the definition above. Indeed, the standard Banach spaces $(\ell_{-} ^{p}(\mathbb{R}^n), \left\| \cdot \right\|_{p})  $, with $1\leq p\leq +\infty $, are particular cases of $(\ell_{-} ^{p , w}(\mathbb{R}^n), \left\| \cdot \right\|_{p , w})  $ that are obtained by taking as sequence $w$ the constant sequence $w ^ \iota$ given by $w ^{\iota} _t:=1 $, for all $t \in \mathbb{N} $. This observation is used in the paper to obtain many results for  the spaces $\ell_{-} ^{\infty}(\mathbb{R}^n) $ as a particular case of those proved for $\ell_{-}^{w}(\mathbb{R}^n) $.

We emphasize that $w ^ \iota$ is not a weighting sequence and that the spaces $(\ell_{-} ^{w}(\mathbb{R}^n), \left\|\cdot \right\| _w) $ considered in this paper are all based on sequences $w$ of weighting type. It can be proved (see \cite[Theorems 3.3 and 4.1]{weightedls}) that, in that case:
\begin{equation}
\label{inclusions lps}
\ell_{-} ^{p , w}(\mathbb{R}^n) \subset \ell_{-} ^{w}(\mathbb{R}^n), \quad \mbox{for any $1\leq p< + \infty$},
\end{equation}
and that,
\begin{equation}
\label{inclusion lpnormals}
\ell_{-} ^{p}(\mathbb{R}^n) \subset \ell_{-} ^{p , w}(\mathbb{R}^n), \quad \mbox{for any $1\leq p\leq + \infty$.}
\end{equation}

All the results in this paper are formulated  for the weighted spaces $(\ell_{-}^{w}(\mathbb{R}^n), \left\|\cdot \right\| _w) $ even though many of the statements that we provide are also valid for $(\ell_{-}^{\infty}(\mathbb{R}^n), \left\|\cdot \right\| _\infty) $  and $(\ell_{-}^{p,w}(\mathbb{R}^n), \left\|\cdot \right\| _{p,w}) $. That will be explicitly pointed out in the statements or in remarks when it is the case.

\subsection{The topologies induced by weighted and supremum norms}
An important feature of the topology generated by weighted norms is that they coincide with the product topology on subsets made of uniformly bounded sequences like the space $K _M $ in \eqref{Kset}. This fact holds true for any weighting sequence $w$ and has important consequences (see \cite{RC7} for the details). First, the fading memory property that we brought up in the introduction and that we spell out in detail later on is independent of the weighting sequence used to define it. Second, the subsets $ K _M \subset \ell_{-} ^{w}(\mathbb{R}^n) $ are compact in the topology induced by the weighted norms $\left\|\cdot \right\|_w $. We emphasize that these statements are valid exclusively in the context of uniformly bounded subsets which, as we see in the next result, are never open in the weighted topology.

We adopt in the sequel the following notation for product sets and functions:  for any family $\{A _t\}_{t \in \mathbb{Z}_{-}} $, of subsets $A _t \subset {\Bbb R}^n $ the symbol
\begin{equation*}
\prod_{t \in \mathbb{Z}_{-}}A _t:= \left\{{\bf z} \in ({\Bbb R}^n)^{\mathbb{Z}_{-}}\mid {\bf z} _t \in A _t, \quad \mbox{for all $t \in \mathbb{Z}_{-} $}\right\},
\end{equation*}
denotes the Cartesian product of the sets in the family. When all the elements in the family are identical to a given subset $A$, we will exchangeably use the symbols $\prod_{t \in \mathbb{Z}_{-}}A $ and $\left(A\right)^{\mathbb{Z}_{-} } $. A similar notation is adopted for the Cartesian  product of maps: let $V $  be a set  and let $f _t:V \longrightarrow A _t $ be a map, $t \in \mathbb{Z}_{-} $. The symbol $\prod _{t \in \mathbb{Z}_{-}}f _t $ denotes the map
\begin{equation}
\label{product of maps}
\begin{array}{cccc}
\prod _{t \in \mathbb{Z}_{-}}f _t : &V& \longrightarrow & \prod_{t \in \mathbb{Z}_{-}}A _t \\
	&v &\longmapsto & \left(\ldots, f_{-2}(v), f_{-1}(v), f_{0}(v)\right).
\end{array}
\end{equation}

\begin{lemma}
\label{topological lemma balls etc}
Let $w$ be a weighting sequence and $n \in \mathbb{N}^+ $. Then:
\begin{description}
\item [(i)] For any ${\bf z} \in \ell_{-}^{w}(\mathbb{R}^n) $ and $r>0 $,
\begin{equation}
\label{balls for the lwtopology}
B_{\left\|\cdot \right\|_w}({\bf z}, r)= \bigcup_{\delta< r} \left(\prod_{t \in \mathbb{Z}_{-}}B_{\left\|\cdot \right\|}\left(\mathbf{z} _t, \frac{\delta}{w _{-t}}\right)\right).
\end{equation}
In particular, this implies that
\begin{equation}
\label{inclusions balls}
B_{\left\|\cdot \right\|_w}({\bf z}, r)\subset \prod_{t \in \mathbb{Z}_{-}}B_{\left\|\cdot \right\|}\left(\mathbf{z} _t, \frac{r}{w _{-t}}\right)\subset  \overline{B_{\left\|\cdot \right\|_w}({\bf z}, r)}.
\end{equation}
The identity \eqref{balls for the lwtopology} implies that any open ball $B_{\left\|\cdot \right\|_w}({\bf z}, r) $ in $\ell_{-}^{w}(\mathbb{R}^n) $ contains unbounded sequences.
 \item [(ii)] Let $\{A _t\}_{t \in \mathbb{Z}_{-}} $ be a family of subsets $A _t \subset {\Bbb R}^n $ such that there exists a sequence $\{ c _t\}_{t \in \mathbb{Z}_{-}} $ that satisfies
\begin{equation}
\label{condition on ats}
\sup_{\mathbf{z} _t \in A _t}\left\{\left\|\mathbf{z} _t\right\|w _{-t}\right\}< c _t,  \mbox{ for each $t \in \mathbb{Z}_{-} $ and $\sup_{t \in \mathbb{Z}_{-}}\left\{c _t\right\}<+ \infty$,}
\end{equation}
then the product set
\begin{equation*}
\prod_{t \in \mathbb{Z}_{-}}A _t\subset \ell_{-}^{w}(\mathbb{R}^n).
\end{equation*}
\item [(iii)] For every family $\{A _t\}_{t \in \mathbb{Z}_{-}} $ of subsets   $A _t \subset {\Bbb R}^n $ such that  the product set satisfies $\prod_{t \in \mathbb{Z}_{-}}A _t\subset \ell_{-}^{w}(\mathbb{R}^n)$, we have
\begin{equation}
\label{equality with closures}
\overline{\prod_{t \in\mathbb{Z}_{-}}A _t}= \prod_{t \in \mathbb{Z}_{-}}\overline{A _t}.
\end{equation}
\end{description}
These statements, except for the last sentence in part {\rm {\bf (i)}}, are also valid for the space $\ell_{-} ^{\infty}(\mathbb{R}^n) $ and are obtained by taking as sequence $w$ the constant sequence $w ^ \iota$ given by $w ^{\iota} _t:=1 $, for all $t \in \mathbb{N} $.
\end{lemma}

\begin{corollary}
\label{dnn closed and open}
Let $D_n $  be a subset of $\mathbb{R}^n$  and let $w$ be a weighting sequence. Then:
\begin{description}
\item [(i)] If $(D _n) ^{\mathbb{Z}_{-}}\cap \ell_{-}^{w}(\mathbb{R}^n) $ is an open subset of $\ell_{-}^{w}(\mathbb{R}^n)  $ then $D _n= {\Bbb R}^n $, necessarily. 
\item [(ii)] If $(D _n) ^{\mathbb{Z}_{-}}\subset  \ell_{-}^{w}(\mathbb{R}^n) $ is a closed subset of $\ell_{-}^{w}(\mathbb{R}^n)  $ then $D_n $ is necessarily closed in ${\Bbb R}^n $, that is, $D _n= \overline{D _n }$. 
\item [(iii)] The following inclusion always holds 
\begin{equation}
\label{inclusion with bars dn}
\overline{(D _n) ^{\mathbb{Z}_{-}}\cap \ell_{-}^{w}(\mathbb{R}^n)} \subset \left(\overline{D _n}\right) ^{\mathbb{Z}_{-}}\cap \ell_{-}^{w}(\mathbb{R}^n).
\end{equation}
 In particular, if $D_n $ is closed in ${\Bbb R}^n $ then so is $(D _n) ^{\mathbb{Z}_{-}}\cap \ell_{-}^{w}(\mathbb{R}^n) $ in $\ell_{-}^{w}(\mathbb{R}^n) $.
\end{description}
These statements in parts {\rm {\bf (ii)}} and {\rm {\bf (iii)}} are also valid when the space $\ell_{-} ^{w}(\mathbb{R}^n) $ is replaced by $\ell_{-}^{\infty}(\mathbb{R}^n)$.
\end{corollary}

We also recall  (see \cite[Proposition 2.9]{RC7}) that the norm topology in $\ell ^{w}_-({\Bbb R}^n) $ is strictly finer than the subspace topology induced by the product topology in $\left(\mathbb{R}^n\right)^{\mathbb{Z}_{-}} $ on $\ell ^{w}_-({\Bbb R}^n) \subset \left(\mathbb{R}^n\right)^{\mathbb{Z}_{-}}$. We complement this fact by comparing the norm topology on $(\ell_{-} ^{\infty}(\mathbb{R}^n), \left\| \cdot \right\|_{\infty}) $ with the relative topology induced by $(\ell ^{w}_-({\Bbb R}^n), \left\| \cdot \right\|_{w})$ on it.

\begin{corollary} 
The relative topology $\tau_{w, \infty}$ induced by the norm topology $\tau_w$ of $( \ell^{w}_-(\mathbb{R}^n), \|\cdot\|_{w})$  on   $\ell ^{\infty}_-(\mathbb{R}^n)$ is strictly coarser than the norm topology $\tau_{\infty}$ on $( \ell^{\infty}_-(\mathbb{R}^n), \|\cdot\|_{\infty})$, that is, $\tau_{w, \infty} \subsetneq \tau_{\infty}$.
\end{corollary}

\noindent\textbf{Proof.\ \ } 
Since, as we already saw, $\left\| {\bf z} \right\|_{w}\leq \left\| {\bf z} \right\|_{\infty} $, for all $\mathbf{z} \in (\mathbb{R}^n) ^{\Bbb Z _ -} $, we have that   $\ell_{-} ^{\infty}(\mathbb{R}^n) \subset \ell ^{w}_-({\Bbb R}^n)$ (see \eqref{inclusion lpnormals}) and the inclusion $\iota: \ell^{\infty}_-(\mathbb{R}^n) \hookrightarrow \ell^w_-(\mathbb{R}^n)$ is continuous. Consequently, for any open $U\in \tau_w$ the set $\iota^{-1}(U) = U\cap \ell^{\infty}_-(\mathbb{R}^n) \in \tau_{w, \infty}$ is also open in $\tau_{\infty}$. This immediately implies that \begin{equation*}
\tau_{w, \infty} \subset \tau_{\infty}.
\end{equation*}
In order to establish that this inclusion is strict, one needs to notice that, given an arbitrary open ball $B_{\|\cdot\|_\infty} ({\bf z}, r)$, $r>0$, around ${\bf z}\in \ell^{\infty}_-(\mathbb{R}^n)$, all the open balls $B_{\|\cdot\|_w} ({\bf z}, \epsilon)$ for all $\epsilon>0$ contain elements that are not included in $B_{\|\cdot\|_\infty} ({\bf z}, r)$ by Lemma \ref{topological lemma balls etc} {\bf(i)}. $\blacksquare$

\begin{lemma}
\label{inclusions with lw}
Let $w$ be a weighting sequence and $n \in \mathbb{N}^+  $. We denote by $w ^a $, $a \in \mathbb{R} $, the sequence with terms $w _t^a$, $t \in \mathbb{N} $.
Then, the following inclusions are continuous:
\begin{equation}
\label{inclusions continuous}
\left(\ell_{-} ^{\infty}(\mathbb{R}^n), \left\| \cdot \right\|_{\infty}\right) \hookrightarrow \cdots\hookrightarrow\left(\ell ^{w^{\frac{1}{k+1}}}_-({\Bbb R}^n), \left\| \cdot \right\|_{w^{\frac{1}{k+1}}}\right)\hookrightarrow\left(\ell ^{w^{\frac{1}{k}}}_-({\Bbb R}^n), \left\| \cdot \right\|_{w^{\frac{1}{k}}}\right)\hookrightarrow\\ \cdots\hookrightarrow\left(\ell ^{w}_-({\Bbb R}^n), \left\| \cdot \right\|_{w}\right),
\end{equation}
\begin{equation}
\label{inclusions continuous powers}
\left(\ell ^{w}_-({\Bbb R}^n), \left\| \cdot \right\|_{w}\right)\hookrightarrow \cdots\hookrightarrow\left(\ell ^{w^{k}}_-({\Bbb R}^n), \left\| \cdot \right\|_{w^{k}}\right)
 \hookrightarrow
\left(\ell ^{w^{{k+1}}}_-({\Bbb R}^n), \left\| \cdot \right\|_{w^{k+1}}\right) 
 \hookrightarrow \cdots \hookrightarrow (\mathbb{R}^n) ^{\mathbb{Z}_-},
\end{equation}
where $k \in \mathbb{N}^+$ and  in $(\mathbb{R}^n) ^{\mathbb{Z}_-} $ we consider the trivial topology.
Define 
\begin{equation}
\label{definition sw}
S _w:=\bigcap _{k \in \mathbb{N}^+}\ell_{-} ^{w^{\frac{1}{k}}}(\mathbb{R}^n) \quad \mbox{and} \quad S ^w:=\bigcup _{k \in \mathbb{N}^+}\ell_{-} ^{w^{k}}(\mathbb{R}^n).
\end{equation}
Then, in general,
\begin{equation}
\label{properties sw}
\ell_{-} ^{\infty}(\mathbb{R}^n) \subsetneq S _w \quad \mbox{ and} \quad S ^w\subsetneq (\mathbb{R}^n) ^{\mathbb{Z}_-}.
\end{equation}
\end{lemma} 

%

\subsection{Continuity and differentiability of maps on infinite sequence spaces}
\label{Continuity and differentiability of maps on infinite sequence spaces}

Much of this paper is related to the continuity and the differentiability of maps of the type $f: U \subset \ell_{-} ^{w^1}(\mathbb{R}^n) \longrightarrow V \subset \ell_{-} ^{w^2}(\mathbb{R}^N) $, with $ w^1, w^2 $ weighting sequences and  $U$ and $V$ subsets of $\ell_{-} ^{w^1}(\mathbb{R}^n) $ and $\ell_{-} ^{w^2}(\mathbb{R}^N) $, respectively, that in the case of differentiable maps are necessarily open. Maps that are continuous with respect to topologies generated by weighted norms will be generically referred to as {\bfi  fading memory maps} (or we say that they have the {\bfi  fading memory property (FMP)}) while when the topology considered is generated by the supremum norm, we just say that the map is {\bfi  continuous}. Most of the definitions that we provide in what follows for the weighted norms case can be adapted to the supremum  norm case by replacing the weighting sequences by the constant sequence $w ^ \iota$ given by $w ^{\iota} _t:=1 $, for all $t \in \mathbb{N} $.

Suppose now that $U$ and $V$ are open subsets. The map $f: U \subset \ell_{-} ^{w^1}(\mathbb{R}^n) \longrightarrow V \subset \ell_{-} ^{w^2}(\mathbb{R}^N) $  is {\bfi  (Fr\'echet) differentiable} at $\mathbf{u} _0 \in U $ when there exists a bounded linear map $D f (\mathbf{u} _0):\ell_{-} ^{w^1}(\mathbb{R}^n) \longrightarrow \ell_{-} ^{w^2}(\mathbb{R}^N) $ that satisfies
\begin{equation}
\label{definition frechet}
\lim_{\mathbf{u} \rightarrow \mathbf{u} _0}\frac{f (\mathbf{u})-f(\mathbf{u} _0)-D f (\mathbf{u} _0) \cdot (\mathbf{u} - \mathbf{u} _0)}{\left\|\mathbf{u}- \mathbf{u} _0\right\|_{w^1}}= {\bf 0}.
\end{equation} 
We say that $f: U \subset \ell_{-} ^{w^1}(\mathbb{R}^n) \longrightarrow V \subset \ell_{-} ^{w^2}(\mathbb{R}^N)$ is of class $C ^1(U) $ when it is differentiable at any point in $U$ and the induced map $Df:U \longrightarrow L \left(\ell_{-} ^{w^1}(\mathbb{R}^n), \ell_{-} ^{w^2}(\mathbb{R}^N)\right) $ is continuous, where the space of linear maps $L \left(\ell_{-} ^{w^1}(\mathbb{R}^n), \ell_{-} ^{w^2}(\mathbb{R}^N)\right) $ is endowed with the operator norm $\vertiii{ \cdot } _{w^1, w^2}  $ defined by
\begin{equation}
\label{operator norm definition}
\vertiii{ A} _{w^1, w^2} :=\sup_{\mathbf{u} \in \ell_{-} ^{w^1}(\mathbb{R}^n)} \left.\left\{\frac{\left\|A(\mathbf{u})\right\|_{w^2}}{\left\|\mathbf{u}\right\|_{w^1}}\right| \mathbf{u}\neq \boldsymbol{0}\right\}, \quad A \in L \left(\ell_{-} ^{w^1}(\mathbb{R}^n), \ell_{-} ^{w^2}(\mathbb{R}^N)\right).
\end{equation}
When in the domain and the range we use the same weighting sequence $w$, we will write $\vertiii{ A} _{w} $ instead of $\vertiii{ A} _{w^1, w^2}  $. The higher order derivatives $$D ^r f (\mathbf{u} _0):\underbrace{ \ell_{-} ^{w^1}(\mathbb{R}^n)\times \cdots\times \ell_{-} ^{w^1}(\mathbb{R}^n)}_{\mbox{$r$ times}}   \longrightarrow \ell_{-} ^{w^2}(\mathbb{R}^N), \quad r \in \mathbb{N}^+, $$  are inductively defined and the map $f$ is said to be of class $C ^{r}(U)$ when it is $r$-times differentiable at any point in $U$ and the induced map $D^rf:U \longrightarrow L^r \left(\ell_{-} ^{w^1}(\mathbb{R}^n), \ell_{-} ^{w^2}(\mathbb{R}^N)\right) $ into the normed space of $r$-multilinear maps is continuous. We recall that the operator norm $\vertiii{ \cdot } _{w^1, w^2}  $ in $L^r \left(\ell_{-} ^{w^1}(\mathbb{R}^n), \ell_{-} ^{w^2}(\mathbb{R}^N)\right) $ is given by
\begin{equation}
\label{operator norm multi definition}
\vertiii{ A} _{w^1, w^2} :=\sup_{\mathbf{u}_1, \ldots, \mathbf{u}_r \in \ell_{-} ^{w^1}(\mathbb{R}^n)} \left.\left\{\frac{\left\|A(\mathbf{u}_1, \ldots, \mathbf{u}_r)\right\|_{w^2}}{\left\|\mathbf{u}_1\right\|_{w^1} \cdots \left\|\mathbf{u}_r\right\|_{w^1}}\right| \mathbf{u}_1, \ldots, \mathbf{u}_r\neq \boldsymbol{0}\right\}, \quad A \in L^r \left(\ell_{-} ^{w^1}(\mathbb{R}^n), \ell_{-} ^{w^2}(\mathbb{R}^N)\right).
\end{equation}

We recall that differentiable functions are automatically continuous and we denote the class of continuous functions by  $C ^0(U) $.
When $f$ is of class $C ^r(U)$ in $U$ for any $r \in \mathbb{N}^+ $, we say that $f$ is {\bfi  smooth} in $U$ and we denote this class by $C^{\infty}(U) $. When $f  $ is smooth in $U$ we can construct for it a Taylor power series expansion. We say that $f$ is {\bfi  analytic } in $U$ when the convergence domain of that power series includes $U$. The analytic class is denoted by $C ^\omega(U) $.

We emphasize that, as we  pointed out in Lemma \ref{topological lemma balls etc}, for any weighting sequence $w$, any open set in $ \left(\ell_{-} ^{w}(\mathbb{R}^n), \left\|\cdot \right\|_w\right) $ contains unbounded sequences. For instance, let $B_{\left\|\cdot \right\|_w}({\bf 0}, \epsilon)$ be the ball of radius $\epsilon >0$ around the zero sequence and let $\mathbf{v} \in {\Bbb R}^n $ be a vector such that $\left\|\mathbf{v}\right\|=1 $. The divergent sequence ${\bf z} $ defined by ${\bf z} _t:= \epsilon\mathbf{v}/2 w_{-t} $ is such that $\left\|{\bf z}\right\|_w= \epsilon/2  $ and hence ${\bf z} \in B_{\left\|\cdot \right\|_w}({\bf 0}, \epsilon) \subset \ell_{-} ^{w}(\mathbb{R}^n)$.

The following lemma  spells out conditions under which infinite Cartesian products of continuous and differentiable functions are continuous and differentiable when we use weighted and supremum norms.

\begin{lemma}
\label{cr using product maps}
Let  $W \subset V $ with $\left(V, \left\|\cdot \right\|\right)$ a normed space  and let $D_N \subset {\Bbb R}^N$  be a subset of $\mathbb{R}^N$. Let $H_t:W\longrightarrow D _N$, $t \in \mathbb{Z}_{-} $, be a family of maps. Consider the corresponding product map $\mathcal{H} : W \longrightarrow \left(D _N\right)^{\mathbb{Z}_{-}}$,  defined as in \eqref{product of maps}:
\begin{equation}
\label{definition fff_new}
\mathcal{H}:=\prod_{t \in \mathbb{Z}_{-}} H _t:= \left(\ldots , H _{-2}, H  _{-1}, H _0\right),  \mbox{ or equivalently, } \left(\mathcal{H}({\bf z})\right)_t:= H _t({\bf z}), \mbox{ $ {\bf z} \in W$, $t \in \mathbb{Z}_{-}  $.}
\end{equation}

\begin{description}
\item [(i)] Endow  $W \subset V$ with the subspace topology. If $D_N  $ is a compact subset of $\mathbb{R}^N $ then   $(D_N)^{\mathbb{Z}_{-}} \subset \ell_{-}^{w}(\mathbb{R}^N) $ for any weighting sequence $w$. If each of the functions $H _t  $ is  continuous then $\mathcal{H} : W\longrightarrow (D_N)^{\mathbb{Z}_{-}}\subset \ell_{-}^{w}(\mathbb{R}^N)$ is also continuous.
\item [(ii)] Let $w $ be a weighting sequence and suppose that $W$ contains a point ${\bf z}^0 $ such that ${\mathcal H}({\bf z}^0)\in \ell_{-}^{w}(\mathbb{R}^N)$. If each of the functions $H _t  $ is Lipschitz continuous  with Lipschitz constant $c ^0 _t $ and the sequence $c ^0:= (c ^0  _t)_{t \in \mathbb{Z}_{-}} $ formed by these Lipschitz constants satisfies that $c ^0 \in \ell_{-}^{w}(\mathbb{R}) $, then $\mathcal{H} : W\longrightarrow (D_N)^{\mathbb{Z}_{-}}\cap \ell_{-}^{w}(\mathbb{R}^N)$ is Lipschitz continuous with Lipschitz constant $c ^0 _{\mathcal{H}} \leq \left\|c ^0\right\| _w$. 
\item [(iii)] Suppose that $W$  is an open convex subset of $\left(V, \left\|\cdot \right\|\right)$ and that it contains a point ${\bf z}^0 $ such that ${\mathcal H}({\bf z}^0)\in \ell_{-}^{w}(\mathbb{R}^N)$. Suppose also that the maps $H _t $ are of class $C ^r(W ) $, $r  \geq 1 $, and let $c ^r_t $ be finite constants such that  $\sup_{{\bf z} \in W} \left\{\vertiii{D^rH _t({\bf z})}\right\}\leq c ^r_t<+ \infty$. If $c ^r:= \left(c^r _t\right) _{t \in \mathbb{Z}_{-}} \in \ell_{-}^{w}(\mathbb{R})  $ then ${\mathcal H} $ is  differentiable of order $r$ when considered as a map ${\mathcal H}: W\subset \left(V, \left\|\cdot \right\|\right) \longrightarrow \left(\ell_{-} ^{w}(\mathbb{R}^N), \left\|\cdot \right\|_w\right)$ and 
\begin{equation}
\label{norm of r derivative with wrglobal}
\vertiii{D^r {\mathcal H}({\bf z})}\leq \|c ^r\| _w, \mbox{ for any ${\bf z} \in W $.} 
\end{equation}
Additionally, if  $ c ^j \in \ell_{-}^{w}(\mathbb{R}) $ for all $j \in \left\{1, \ldots, r\right\}$, then $ \mathcal{H} $ is of class $C^{r-1}(W) $ and the map
$D ^{r-1} \mathcal{H}: (W, \left\|\cdot \right\|) \longrightarrow \left(L ^{r-1}(V, \ell_{-} ^{w}(\mathbb{R}^N)), \vertiii{\cdot }\right)  $ is Lipschitz continuous with Lipschitz constant $c _{{\mathcal H}} ^r\leq \|c ^r\| _w $.
\item [(iv)] Suppose that $W$  is an open convex subset of $\left(V, \left\|\cdot \right\|\right)$ and that it contains a point ${\bf z}^0 $ such that ${\mathcal H}({\bf z}^0)\in \ell_{-}^{w}(\mathbb{R}^N)$. If the maps $H _t $ are smooth and $\|c ^r\|_{w}< +\infty $, for each $r \in \mathbb{N}^+$, then so is $\mathcal{H}: W \subset \left(V, \left\|\cdot \right\|\right)\longrightarrow (\ell_{-} ^{w}(\mathbb{R}^N) , \left\|\cdot \right\|_{w}) $.  Suppose, additionally, that the maps $H _t $ are analytic and that $\rho _t  >0$ is the radius of convergence of the series expansion of $H _t $. If $\rho:=\inf_{ t \in \mathbb{Z}_{-}}\left\{\rho _t\right\}>0$
then ${\mathcal H} $ is  analytic when considered as a map $\mathcal{H}: W \subset \left(V, \left\|\cdot \right\|\right)\longrightarrow (\ell_{-} ^{w}(\mathbb{R}^N) , \left\|\cdot \right\|_{w}) $ and the radius of convergence $\rho_{{\mathcal H}} $ of its series expansion satisfies that $\rho_{{\mathcal H}}\geq \rho >0$.
\end{description}

\noindent Parts {\bf (ii)}, {\bf (iii)}, and {\bf (iv)}  also hold true when the Banach space $(\ell_{-}^{w}(\mathbb{R}^N), \left\|\cdot \right\| _w) $ is replaced  by $(\ell_{-}^{\infty}(\mathbb{R}^N), \left\|\cdot \right\|_{\infty} ) $.     Part {\bf (i)} is in general false in that situation.
\end{lemma}

\section{Differentiable time-invariant filters and functionals}
\label{Differentiable time-invariant filters and functionals}
Let $D_n \subset \mathbb{R}^n $ and $D_N \subset \mathbb{R}^N $. We  refer to the maps of the type $U: (D _n) ^{\Bbb Z} \longrightarrow (D_N) ^{\Bbb Z} $ as {\bfi  filters} or {\bfi  operators} and to those like $H: (D _n) ^{\Bbb Z} \longrightarrow D_N $ (or $H: (D _n) ^{\Bbb Z_\pm} \longrightarrow D_N $) as $\mathbb{R}^N $-valued  {\bfi  functionals}. These definitions can be easily extended to accommodate situations where the domains and the targets of the filters are not necessarily product spaces but just arbitrary subsets $V _n $ and $V _N $ of $\left({\Bbb R}^n\right)^{\mathbb{Z}}  $ and $\left({\Bbb R}^N\right)^{\mathbb{Z}}  $ like, for instance, $\ell ^{\infty}(\mathbb{R}^n) $ and $\ell ^{\infty}(\mathbb{R}^N) $, or $\ell_{-}^{w}(\mathbb{R}^n) $ and $\ell_{-}^{w}(\mathbb{R}^N) $, for some weighting sequence $w $.  A filter $U: (D _n) ^{\Bbb Z} \longrightarrow (D_N) ^{\Bbb Z} $ is called {\bfi  causal} when for any two elements ${\bf z} , \mathbf{w} \in (D _n) ^{\Bbb Z}  $  that satisfy that ${\bf z} _\tau = \mathbf{w} _\tau$ for any $\tau \leq t  $, for a given  $t \in \Bbb Z $, we have that $U ({\bf z}) _t= U ({\bf w}) _t $. Let  $T^{\Bbb Z}_\tau:({\Bbb R}^n) ^{\Bbb Z} \longrightarrow({\Bbb R}^n) ^{\Bbb Z} $ be the {\bfi  time delay} operator defined by $T^{\Bbb Z}_\tau( {\bf z}) _t:= {\bf z}_{t- \tau}$, $\tau \in \Bbb Z  $. A subset $V _n \subset ({\Bbb R}^n) ^{\Bbb Z}$ is called time-invariant when $T^{\Bbb Z} _\tau(V _n)= V _n $, for all $\tau \in \Bbb Z $. The filter $U$ is called {\bfi  time-invariant}  when it is defined on a time-invariant set and commutes with the time delay operator, that is, $T^{\Bbb Z}_\tau \circ U=U  \circ T^{\Bbb Z}_\tau $,  for any $\tau\in \Bbb Z $ (in this expression, the two  operators $T^{\Bbb Z}_\tau $ have  to be understood as defined in the appropriate sequence spaces). 

We recall that there is a bijection between causal time-invariant filters and functionals on $(D_n)^{\Bbb Z _-} $. Indeed, given a time-invariant filter $U:(D_n) ^{\Bbb Z} \longrightarrow (\mathbb{R}^N) ^{\Bbb Z}$, we can associate to it a functional $H _U: (D_n) ^{\Bbb Z_-} \longrightarrow \mathbb{R} ^N$ via the assignment $H _U ({\bf z}):= U({\bf z} ^e) _0 $, where ${\bf z} ^e \in (\mathbb{R}^n)^{\Bbb Z } $ is an arbitrary extension of ${\bf z} \in (D_n)^{\Bbb Z _-} $ to $ (D_n)^{\Bbb Z } $.  Conversely, for any functional  $H: (D_n) ^{\Bbb Z_-} \longrightarrow \mathbb{R} ^N$, we can define a time-invariant causal filter $U_H:(D_n) ^{\Bbb Z} \longrightarrow (\mathbb{R}^N) ^{\Bbb Z}$ by $U_H({\bf z}) _t:= H((\mathbb{P}_{\Bbb Z_-} \circ T^{\Bbb Z} _{-t}) ({\bf z})) $, where $T^{\Bbb Z} _{-t} $ is the $(-t)$-time delay operator and $\mathbb{P}_{\Bbb Z_-}: (\mathbb{R}^n)^{\Bbb Z} \longrightarrow (\mathbb{R}^n)^{\Bbb Z _-} $ is the natural projection. 
Moreover, when considering  causal  and time-invariant filters $U: (D_n) ^{\Bbb Z} \longrightarrow (D_N) ^{\Bbb Z} $ it suffices to work  just with the restriction $U: (D_n) ^{\Bbb Z_-} \longrightarrow (D_N) ^{\Bbb Z_-} $, that we denote with the same symbol, since the latter uniquely determines the former. Indeed, by definition,  for any ${\bf z} \in ( D _n) ^{\Bbb Z} $ and $t \in \mathbb{N}^+$:
\begin{equation}
\label{why we can restrict to zminus}
U ({\bf z})_t= \left(T^{\Bbb Z}_{-t} \left(U({\bf z})\right)\right)_0= U \left(T^{\Bbb Z}_{-t}({\bf z})\right)_0,
\end{equation}
where the second equality holds by the time-invariance of $U$ and the value in the right-hand side depends only on $\mathbb{P}_{\mathbb{Z}_{-}}\left(T^{\Bbb Z}_{-t}({\bf z})\right) \in (D_n) ^{\Bbb Z_-}$, by causality.

In view of this observation, we restrict our study to filters with domain and target in the spaces of left semi-infinite  sequences. In particular,  we say that a causal and time-invariant filter $U$ has the fading memory property or that it is continuous when the corresponding restricted filter  defined on left semi-infinite inputs has those properties, as we defined them in Section \ref{Continuity and differentiability of maps on infinite sequence spaces}.

Additionally, from now on we   consider most of the time  time delay operators with domain and target in $({\Bbb R}^n) ^{\Bbb Z_-} $ and that we simply denote as $T _{-\tau}: (\mathbb{R}^n)^{\Bbb Z _-} \longrightarrow (\mathbb{R}^n)^{\Bbb Z _-} $. The definition of these restricted time delay operators $T _{-\tau}$ requires considering two cases:
\begin{itemize}
\item $T _{-\tau}: (\mathbb{R}^n)^{\Bbb Z _-} \longrightarrow (\mathbb{R}^n)^{\Bbb Z _-} $ with $\tau$ negative: as before,  $T_{-\tau}( {\bf z}) _t:= {\bf z}_{t+ \tau}$,  for any ${\bf z} \in (\mathbb{R}^n)^{\Bbb Z _-}  $ and $t \in \mathbb{Z}_{-} $. This implies that, in this case,
\begin{equation*}
T _{-\tau}({\bf z})=\mathbb{P}_{\mathbb{Z}_{-}}\circ T ^{\mathbb{Z}}_{-\tau}({\bf z} ^e), \quad {\bf z} \in (\mathbb{R}^n)^{\Bbb Z _-}, \quad \tau<0,
\end{equation*}
where ${\bf z} ^e \in (\mathbb{R}^n)^{\Bbb Z } $ is an arbitrary extension of ${\bf z} \in (\mathbb{R}^n)^{\Bbb Z _-} $ to $ (\mathbb{R}^n)^{\Bbb Z } $.  The map $T _{-\tau}$, $\tau\in \mathbb{Z}_{-}  $,  is surjective, that is, $T _\tau((\mathbb{R}^n)^{\Bbb Z _-})= (\mathbb{R}^n)^{\Bbb Z _-}  $, but it is not injective. The same applies to the restriction of  $T _{-\tau}  $ to any time-invariant set $V _n\subset ({\Bbb R}^n) ^{\mathbb{Z}_{-}} $ which satisfies $T _{-\tau}(V _n)=V _n   $.
\item $T _{-\tau}: (\mathbb{R}^n)^{\Bbb Z _-} \longrightarrow (\mathbb{R}^n)^{\Bbb Z _-} $ with $\tau$ positive: there is in principle not a unique way to define the restricted operators $T _{-\tau} $ since that involves the choice of  vectors $\mathbf{v}_\tau \in (\mathbb{R}^n)^\tau $ such that $T _{-\tau} ({\bf z}):=({\bf z}, \mathbf{v}_\tau)$, for any ${\bf z} \in \left({\Bbb R}^n\right)^{\mathbb{Z}_{-}} $. The choice $\mathbf{v}_\tau= {\bf 0} $ for all $\tau>0 $ is canonical since it is the only one that makes the resulting maps linear and additionally satisfy
$$
T _{-\tau} =\underbrace{T_{-1}  \circ  \cdots\circ T_{-1}}_{\mbox{$\tau$ times}}.
$$
We hence adopt the definition 
\begin{equation*}
T _{-\tau} ({\bf z}):=({\bf z}, \underbrace{{\bf 0}, \ldots, {\bf 0}}_{\mbox{$\tau$ times}}), \quad {\bf z} \in (\mathbb{R}^n)^{\Bbb Z _-}, \quad \tau>0,
\end{equation*}
for the rest of the paper. In this case $T _{-\tau} $ it is injective but not surjective.
\end{itemize}

The following lemma gathers some differentiability properties of projections and time delay operators when restricted to normed sequence spaces and that will be used later on. A key element in this result is what we call, for each weighting sequence $w$, their {\bfi  decay ratio} $D_w $  and  {\bfi  inverse decay ratio} $L _w $, that are defined as:
\begin{equation}
\label{inverse decay ratio}
D_w:=\sup_{t \in \mathbb{N}}\left\{\frac{w _{t+1}}{w _{t}}\right\} \quad \mbox{and } \quad L _w:=\sup_{t \in \mathbb{N}}\left\{\frac{w _t}{w _{t+1}}\right\}.
\end{equation} 
As $w$ is by definition strictly decreasing we necessarily have that  $0<w _{t+1}/w _{t}<1$, for all $t \in \mathbb{N} $, and $1<w _0/w _{1}\leq\sup_{t \in \mathbb{N}}\left\{w _t/w _{t+1}\right\} =L _w$. Consequently:
\begin{equation*}
0<D _w \leq 1 \quad \mbox{and} \quad 1<L _w\leq + \infty.
\end{equation*}
The decay ratios provide a geometric bound for the convergence speed of $w$  and the divergence rate of $w ^{-1} $. Indeed, it is easy to see that
\begin{equation}
\label{convergence and divergence rates}
w _t\leq D_w ^t \quad \mbox{and} \quad 1/ w _t\leq L _w ^t, \quad \mbox{ for any $t \in \mathbb{N} $.} 
\end{equation}
Additionally, the fact that for all $t \in \mathbb{N} $ we have that  $1 < w _{t}/w _{t+1}$ and that $0<w _{t+1}/w _{t}<1$ implies that
\begin{equation*}
1/\sup_{t \in \mathbb{N}} \left\{\frac{w _t}{w _{t+1}}\right\}=\inf_{t \in \mathbb{N}} \left\{\frac{w _{t+1}}{w _{t}}\right\}\leq \sup_{t \in \mathbb{N}} \left\{\frac{w _{t+1}}{w _{t}}\right\}\ \   \mbox{and} \ \ 
1/\sup_{t \in \mathbb{N}} \left\{\frac{w _{t+1}}{w _{t}}\right\}=\inf_{t \in \mathbb{N}} \left\{\frac{w _{t}}{w _{t+1}}\right\}\leq \sup_{t \in \mathbb{N}} \left\{\frac{w _{t}}{w _{t+1}}\right\},
\end{equation*}
which, in both cases, implies that 
\begin{equation}
\label{relation ratios}
L _wD _w\geq 1.
\end{equation}
More generally, in relation with the power weighting sequences that we discussed in Lemma \ref{inclusions with lw}, we have that:
\begin{equation}
\label{inclusion with powers}
0<D _{w^{n}}\leq D _w\leq D _{w^{1/m}}\leq 1 \quad \mbox{and } \quad
1<L _{w^{1/m}}\leq L _w\leq L _{w^n}\leq + \infty,\quad \mbox{for any $m,n \in \mathbb{N}^+ $.}
\end{equation}

\begin{lemma}
\label{smooth projections and time delays}
Let  $w$ be a weighting sequence and $ n \in \mathbb{N}^+  $. Then:
\begin{description}
\item [(i)] The projections $p _t:  (\ell_{-} ^{w}(\mathbb{R}^n), \left\|\cdot \right\|_w) \longrightarrow (\mathbb{R}^n, \left\|\cdot \right\| ) $, $t \in \mathbb{Z}_{-}  $, given by $p _t ({\bf z}):= {\bf z} _t $, ${\bf z} \in \ell_{-} ^{w}(\mathbb{R}^n) $, are linear, smooth, and hence continuous. Moreover, $\vertiii{ p _t} _w= 1 / w _{-t}$.
\item [(ii)] Consider the restriction of the time delay operator $T _{-t} $ to  $\ell_{-} ^{w}(\mathbb{R}^n) $ for any $t \in \Bbb Z $. We  consider two cases. First, if $t <0 $ and the inverse decay ratio $L _w  $ of $w$ is finite, then $T _{-t} $
maps into  $\ell_{-} ^{w}(\mathbb{R}^n) $, that is, $\ell_{-} ^{w}(\mathbb{R}^n) $ is $T _{-t}$-invariant and
$T _{-t}: (\ell_{-} ^{w}(\mathbb{R}^n), \left\|\cdot \right\|_w) \longrightarrow (\ell_{-} ^{w}(\mathbb{R}^n), \left\|\cdot \right\|_w) $ is surjective, open, and  a submersion, that is, $\ker T _{-t} $ is a split subspace of $\ell_{-}^{w}(\mathbb{R}^n) $. 
If $t >0 $, then $\ell_{-} ^{w}(\mathbb{R}^n) $ is always $T _{-t}$-invariant. $T _{-t}: (\ell_{-} ^{w}(\mathbb{R}^n), \left\|\cdot \right\|_w) \longrightarrow (\ell_{-} ^{w}(\mathbb{R}^n), \left\|\cdot \right\|_w) $ is in that case an immersion, that is, it is injective and its image ${\rm Im} \,T _{-t} $ is split. Moreover, for any $t>0 $,
$
T _{t} \circ T _{-t}= \mathbb{I}_{\ell_{-} ^{w}(\mathbb{R}^n)},
$
and in both cases the maps $T _{-t} $ are linear, smooth, and hence continuous. Additionally,
\begin{equation}
\label{norms of tt}
\vertiii{ T _{1}} _w= L _w, \quad \vertiii{ T _{-1}} _w= {D _w}, \quad \vertiii{ T _{-t}} _w\leq L _w^{- t}, \quad \mbox{and} \quad \vertiii{ T _{t}} _w\leq D _w^{-t}, \quad \mbox{for all $t \in \mathbb{Z}_{-} $}.
\end{equation}

\item [(iii)] For any $t _1, t _2  \in \mathbb{Z}_{-} $  we have
\begin{equation}
\label{relation p and T}
p_{t _1+ t _2}= p_{t _1} \circ T_{-t _2}=p_{t _2} \circ T_{-t _1}.
\end{equation}
\end{description}
These statements also hold true when $(\ell_{-} ^{w}(\mathbb{R}^n), \left\|\cdot \right\|_w) $ is replaced by $(\ell_{-} ^{\infty}(\mathbb{R}^n), \left\|\cdot \right\|_\infty) $. In that case one has to take as sequence $w$ the constant sequence $w ^ \iota$ given by $w ^{\iota} _t:=1 $, for all $t \in \mathbb{N} $, and $L _w  $ is replaced by the constant $1$.
\end{lemma}

\begin{remark}
\label{examples for lw}
\normalfont
The  decay ratios are easy to compute for many families of weighting sequences.  Two cases that we frequently encounter are: 
\begin{description}
\item [(i)] Geometric sequence: $w _t:= \lambda ^t $, $t \in \mathbb{N} $, with $0<\lambda<1 $. In this case:
\begin{equation*}
L _w:=\sup_{t \in \mathbb{N}}\left\{\frac{\lambda ^{t}}{\lambda ^{t+1}}\right\}= \frac{1}{\lambda}>1 \quad \mbox{and} \quad D _w:=\sup_{t \in \mathbb{N}}\left\{\frac{\lambda ^{t+1}}{\lambda ^{t}}\right\}= {\lambda}<1.
\end{equation*}
\item [(ii)] Harmonic sequence: $w _t:= 1/(1+td) $, $t \in \mathbb{N} $, with $d>0$. In this case $D_w=1 $ and $L _w= 1+d $.
\end{description}
We emphasize that the finiteness of the inverse decay ratio is not  guaranteed for all weighting sequences. An example that illustrates this fact is the sequence $w _t:=\exp(-t^2) $. It is easy to verify that in that case $L _w=+ \infty $ and $D _w=1/e $.
\end{remark}

\begin{remark}
\normalfont
The inequalities \eqref{norms of tt} can be combined with Gelfand's formula \cite[page 195]{lax:functional:analysis} to provide bounds for the spectral radii  $\rho(T _{-t}) $ and $\rho(T _{t}) $ for all $t \in \mathbb{Z}_{-} $. Indeed,
\begin{equation*}
\rho(T _{-t})=\lim\limits_{n \rightarrow \infty}\vertiii{T _{-t}^n}_w ^{1/n} \leq \lim\limits_{n \rightarrow \infty}(L _w^{-tn}) ^{1/n}=L _w^{-t}, \quad \mbox{with $t \in \mathbb{Z}_{-} $.}
\end{equation*}
Analogously, one shows that $\rho(T _{t})\leq D_w^{-t}$.
\end{remark}

\begin{remark}
\label{norms for pw case}
\normalfont
Lemma \ref{smooth projections and time delays} remains valid when instead of the spaces $\ell_{-}^{w}(\mathbb{R}^n) $ we use the spaces $\ell_{-}^{p,w}(\mathbb{R}^n) $ that we introduced in Section \ref{Differentiable reservoir computing systems}, for any $1\leq p<+ \infty $. In that case, and for any $t \in \mathbb{Z}_{-} $,
\begin{equation}
\label{norm of p for p}
\vertiii{p _t} _{p,w} =  \frac{1}{w _{-t}^{1/p}},
\end{equation}
\begin{equation}
\label{norm of t for p}
\vertiii{ T _{1}} _{p,w}= L _{w,p}, \  \vertiii{ T _{-1}} _{p,w}= {D _{w,p}}, \  \vertiii{ T _{-t}} _{p,w}\leq L _{w,p}^{- t}, \  \mbox{and} \  \vertiii{ T _{t}} _{p,w}\leq D _{w,p}^{-t}, \  \mbox{for all $t \in \mathbb{Z}_{-} $}.
\end{equation}
where $D _{w,p} $ and $L _{w,p} $ are adapted versions to these norms of the decay ratio $D_w  $ and $L _w $, respectively, given by
\begin{equation}
\label{inverse decay ratio for p}
D _{w,p}:=\sup_{t \in \mathbb{N}}\left\{\left(\frac{w _{t+1}}{w _{t}}\right)^{1/p}\right\}\quad \mbox{and} \quad
L _{w,p}:=\sup_{t \in \mathbb{N}}\left\{\left(\frac{w _t}{w _{t+1}}\right)^{1/p}\right\}.
\end{equation}
Indeed, \eqref{norm of p for p} can be obtained by noting that for any $\mathbf{u} \in \ell_{-}^{p,w}(\mathbb{R}^n) $,
\begin{equation*}
\left\|p _t(\mathbf{u})\right\|^p= \left\|\mathbf{u} _t\right\|^p= \frac{\left\|\mathbf{u} _t\right\|^p w _{-t}}{w _{-t}}\leq \frac{1}{w _{-t}}\sum_{s \in \mathbb{Z}_{-}}\left\|\mathbf{u} _s\right\|^p w_{-s}= \frac{\left\|\mathbf{u}\right\|^p_{p,w}}{w _{-t}},
\end{equation*}
which shows that $\vertiii{p _t} _{p,w} \leq  {1}/{w _{-t}^{1/p}} $. Consider now the vector $\mathbf{v} \in \ell_{-}^{p,w}(\mathbb{R}^n) $ given by $\mathbf{v} _s:= \delta_{s,t} \widetilde{\mathbf{v}}/ w _{-t}^{1/p} $, where $s \in \mathbb{Z}_{-} $, the symbol $\delta_{s,t} $ stands for Kronecker's delta, and $\widetilde{\mathbf{v}}\in {\Bbb R}^n $ is such that $\left\|\widetilde{\mathbf{v}}\right\|=1 $. Notice that $\left\|\mathbf{v}\right\| ^p_{p,w}= \sum_{s \in \mathbb{Z}_{-}}\left\|\mathbf{v} _s\right\|^p w_{-s}= \left\|\widetilde{\mathbf{v}}\right\|=1$ and given that $\left\|p _t(\mathbf{v})\right\|= \left\|\widetilde{\mathbf{v}}\right\|/ w _{-t}^{1/p}=1/ w _{-t}^{1/p}$, this implies that
\begin{equation*}
\vertiii{p _t}_{p,w}=\sup_{\left\|\mathbf{u}\right\|_{p,w}=1}\left\{\left\|p _t(\mathbf{u})\right\|\right\}= \frac{1}{w _{-t}^{1/p}},
\end{equation*}
as required. Regarding \eqref{norm of t for p}, we only sketch the proof for positive time shifts. As in the proof of Lemma \ref{smooth projections and time delays},  it suffices to show that $\vertiii{T _{1}} _{p,w}= L _{w,p} $. In order to prove this equality, notice first that for any $t \in \mathbb{Z}_{-} $, the following straightforward inequality holds
\begin{equation*}
\frac{w _{-t}}{w_{-(t-1)}}= \left(\left(\frac{w _{-t}}{w_{-(t-1)}}\right)^{1/p}\right)^{p} \leq \left(\sup_{s \in \mathbb{Z}_{-}}\left\{\left(\frac{w _{-s}}{w_{-(s-1)}}\right)^{1/p}\right\}\right)^{p}= L _{w,p}^p.
\end{equation*}
Now, for any $\mathbf{u} \in \ell_{-}^{p,w}(\mathbb{R}^n) $,
\begin{multline*}
\left\|T _1(\mathbf{u})\right\|_{p,w}= \left(\sum_{t \in \mathbb{Z}_{-}}\left\|\mathbf{u} _{t-1}\right\|^p w _{-t}\right)^{1/p}= \left(\sum_{t \in \mathbb{Z}_{-}}\left\|\mathbf{u} _{t-1}\right\|^p w _{-(t-1)}\frac{w _{-t}}{w _{-(t-1)}}\right)^{1/p}\\
\leq  \left(\sum_{t \in \mathbb{Z}_{-}}\left\|\mathbf{u} _{t-1}\right\|^p w _{-(t-1)}L _{w,p}^p\right)^{1/p}\leq 
\left\|\mathbf{u}\right\|_{p,w} L _{w,p},
\end{multline*}
which proves that $\vertiii{T _{1}} _{p,w}\leq L _{w,p} $. In order to establish the equality we prove the reverse inequality by considering the family of vectors $\mathbf{v}^t \in \ell_{-}^{p,w}(\mathbb{R}^n) $, $t \in \mathbb{Z}_{-}  $ defined by $\mathbf{v}^t _s:= \delta_{s,t} \widetilde{\mathbf{v}}^t/ w _{-(t-1)}^{1/p} $, where $\widetilde{\mathbf{v}}^t \in {\Bbb R}^n $ is such that $\left\|\widetilde{\mathbf{v}}^t\right\|= \left(w_{-(t-1)}/ w _{-t}\right)^{1/p} $. Notice that for all $t \in \mathbb{Z}_{-} $,
\begin{equation*}
\left\|\mathbf{v}^t\right\|^p_{p,w}=\sum_{s \in \mathbb{Z}_{-}}\left\|\mathbf{v}_s ^t\right\|^pw_{-s}= \frac{\left\|\widetilde{\mathbf{v}}^t\right\|^p}{w_{-(t-1)}} w _{-t}= \frac{w_{-(t-1)}}{w _{-t}} \frac{w _{-t}}{w_{-(t-1)}}=1 \quad \mbox{and} \quad \left\|T _1(\mathbf{v} ^t)\right\|^p_{p,w}= \frac{w_{-(t+1)}}{w _{-t}}\geq 1,
\end{equation*}
which implies that
\begin{equation*}
\vertiii{T _{1}} _{p,w}= \sup_{\left\|\mathbf{u}\right\|_{p,w}=1}\left\{\left\|T _1(\mathbf{u})\right\|_{p,w}\right\}
\geq  \sup_{\mbox{$t \in \mathbb{Z}_{-}$}}\left\{\left\|T _1(\mathbf{v} ^t)\right\|_{p,w}\right\}=
\sup_{t \in \mathbb{Z}_{-}} \left\{ \left(\frac{w_{-(t+1)}}{w _{-t}}\right)^{1/p}\right\},
= L _{w,p}
\end{equation*}
which proves the required inequality.
\end{remark}

\begin{remark}
\normalfont
Some of the properties of time delays operators that we just studied have interesting interpretations in a Hilbert space context. See \cite{lindquist:picci} for a detailed study.
\end{remark}

\subsection{The fading memory property and remote past input independence}
\label{The fading memory property and remote past input independence}

The properties of time delay operators that we enunciated in  Lemma \ref{smooth projections and time delays} allow us to show how the fading memory property, defined as the continuity of a filter linking input and output spaces endowed with weighted norms, (see Section \ref{Continuity and differentiability of maps on infinite sequence spaces}) can be interpreted as its asymptotic independence on the remote past input \cite[page 89]{wiener:book}. Analogously, we can see that the FMP amounts to the attribute that, in the words of Volterra \cite[page 188]{volterra:book},  the influence of the input a long time before the given moment fades out. This property has also been characterized as a {\bfi   unique steady-state property} in \cite{Boyd1985} and referred to as the {\bfi input forgetting property} in \cite{jaeger2001}. All these characterizations were proved under various compactness and/or uniformly boundedness hypotheses on the inputs. The next result  shows that property as a straightforward corollary of Lemma \ref{smooth projections and time delays} that, later on in Section \ref{Remote past input independence and the state forgetting property for unbounded inputs}, will be generalized to situations where the inputs are eventually unbounded.

In the following statement we will be using the following notation: given the sequences $\mathbf{u} \in ({\Bbb R}^n)^{\mathbb{Z}_{-}} $ and $\mathbf{v} \in ({\Bbb R}^n) ^t $, $t \in \mathbb{N}$, the symbol $(\mathbf{u}, \mathbf{v}) \in ({\Bbb R}^n)^{\mathbb{Z}_{-}} \times ({\Bbb R}^n) ^t  $ denotes the {\bfi  concatenation} of $ \mathbf{u} $ and $\mathbf{v} $.

\begin{theorem}[FMP and the uniform input forgetting property]
\label{FMP and the uniform input forgetting property}
Let $M , L>0  $, $n , N \in \mathbb{N}^+$  and let $K _M \subset ({\Bbb R}^n)^{\mathbb{Z}_{-}}$, $K _L \subset ({\Bbb R}^M)^{\mathbb{Z}_{-}}$ (respectively, $K _M^+ \subset ({\Bbb R}^n)^{\mathbb{N}^+}$, $K _L^+ \subset ({\Bbb R}^M)^{\mathbb{N}^+}$) be the sets of uniformly bounded left (respectively, right) semi-infinite sequences defined in \eqref{Kset}. Let $U:K _M \longrightarrow U _L $ be a causal and time-invariant fading memory filter. Then, for any $\mathbf{u}, \mathbf{v} \in K _M $ and  ${\bf z} \in K _M^+ $ we have that
\begin{equation}
\label{property past input independence 1}
\lim\limits_{t \rightarrow + \infty} \left\|U(\mathbf{u}, {\bf z})_t-U(\mathbf{v}, {\bf z})_t\right\|=0,
\end{equation}
where in this expression the filter $U$ is defined by time-invariance on positive times using \eqref{why we can restrict to zminus}.  The convergence in \eqref{property past input independence 1} is uniform on $\mathbf{u}, \mathbf{v}$, and  ${\bf z}$ in the sense that there exists a monotonously decreasing sequence $w ^U $ with zero limit such that for all $\mathbf{u}, \mathbf{v} \in K _M $,  ${\bf z} \in K _M^+ $, and $t \in \mathbb{N} $,
\begin{equation}
\label{property past input independence uniform}
\left\|U(\mathbf{u}, {\bf z})_t-U(\mathbf{v}, {\bf z})_t\right\|\leq w ^U _t.
\end{equation}
Filters that satisfy condition \eqref{property past input independence 1} for any $\mathbf{u}, \mathbf{v} \in K _M $ and  ${\bf z} \in K _M^+ $ are said to have the {\bfi  input forgetting property} and we refer to \eqref{property past input independence uniform} as the {\bfi  uniform input forgetting property}.
\end{theorem}

\noindent\textbf{Proof.\ \ } We start by recalling that in the presence of uniformly bounded inputs, the FMP can be characterized as the continuity of the map $U:K _M \longrightarrow K _L $ with the sets $K _M $ and $K _L $  endowed with the relative topology induced either by the product topology on $({\Bbb R}^n)^{\mathbb{Z}_{-}} $ and $({\Bbb R}^N)^{\mathbb{Z}_{-}} $, respectively, or by the weighted norms in the spaces $\ell_{-}^{w}(\mathbb{R}^n) $ and $\ell_{-}^{w}(\mathbb{R}^N)$, with $w$ any weighting sequence (see \cite[Corollary 2.7 and Proposition 2.11]{RC7}). Moreover, the sets $K _M $ and $K _L $  are compact in this topology \cite[Corollay 2.8]{RC7} and hence the FMP filter $U:K _M \longrightarrow K _L $ is not only continuous but also uniformly continuous. Consequently, once we have fixed a weighting sequence $w$, an increasing  modulus of continuity $\omega _U: \mathbb{R}^+\longrightarrow \mathbb{R}^+ $ can be associated to the map $U:(K _M, \left\|\cdot \right\|_w) \longrightarrow (K _L, \left\|\cdot \right\|_w) $. We emphasize that $\omega _U $ depends on $w$ since it is a metric and not a purely topological notion. Now, using \eqref{why we can restrict to zminus} and an arbitrary weighting sequence $w$ that we choose with $D_w< 1 $, we can write for any $t \in \mathbb{N} $
\begin{multline}
\label{first step for independence}
\left\|U(\mathbf{u}, {\bf z})_t-U(\mathbf{v}, {\bf z})_t\right\|= \left\|U \left(\mathbb{P}_{\mathbb{Z}_{-}}\left(T^{\Bbb Z}_{-t}(\mathbf{u}, {\bf z})\right)\right)_0-U \left(\mathbb{P}_{\mathbb{Z}_{-}}\left(T^{\Bbb Z}_{-t}(\mathbf{v},{\bf z})\right)\right)_0\right\|\\
=\left\|p _0\circ U \left(\mathbb{P}_{\mathbb{Z}_{-}}\left(T^{\Bbb Z}_{-t}(\mathbf{u}, {\bf z})\right)\right)-p _0\circ U \left(\mathbb{P}_{\mathbb{Z}_{-}}\left(T^{\Bbb Z}_{-t}(\mathbf{v},{\bf z})\right)\right)\right\|\\
\leq \left\|U \left(\mathbb{P}_{\mathbb{Z}_{-}}\left(T^{\Bbb Z}_{-t}(\mathbf{u}, {\bf z})\right)\right)-U \left(\mathbb{P}_{\mathbb{Z}_{-}}\left(T^{\Bbb Z}_{-t}(\mathbf{v},{\bf z})\right)\right)\right\|_w,
\end{multline}
where we used that $\vertiii{p _0}_w =1$ by the first part of Lemma \ref{smooth projections and time delays}. We now notice that
\begin{equation*}
\mathbb{P}_{\mathbb{Z}_{-}}\left(T^{\Bbb Z}_{-t}(\mathbf{u}, {\bf z})\right)=T _{-t} (\mathbf{u})+ \left(\ldots, {\bf 0}, {\bf z}_1, \ldots, {\bf z} _t\right), \quad \mbox{and} \quad \mathbb{P}_{\mathbb{Z}_{-}}\left(T^{\Bbb Z}_{-t}(\mathbf{v}, {\bf z})\right)=T _{-t} (\mathbf{v})+ \left(\ldots, {\bf 0}, {\bf z}_1, \ldots, {\bf z} _t\right),
\end{equation*}
which substituted in \eqref{first step for independence} and using the second part of Lemma  \ref{smooth projections and time delays} yields 
\begin{multline}
\label{second step for independence}
\left\|U(\mathbf{u}, {\bf z})_t-U(\mathbf{v}, {\bf z})_t\right\|
\leq \omega_U \left( \left\|T _{-t} (\mathbf{u}-\mathbf{v})\right\|_w\right)\\
\leq 
\omega_U \left(\vertiii{T _{-t}}_w  \left\|\mathbf{u}- \mathbf{v} \right\|_w\right)
\leq  \omega_U \left(D_w^{t}  \left\|\mathbf{u}- \mathbf{v} \right\|_w\right)\leq 
\omega_U \left(2MD_w^{t}  \right).
\end{multline}
Now, as $w$ has been chosen so that $D _w<1 $  and $\lim\limits_{t \rightarrow 0} \omega_U (t)=0 $, we set $w ^U _t:=\omega_U \left(2MD_w^{t}  \right) $, and  we have that
\begin{equation}
\label{forget rate}
\lim\limits_{t \rightarrow + \infty}w ^U _t=\lim\limits_{t \rightarrow + \infty}\omega_U \left(2MD_w^{t}  \right)=0,
\end{equation}
which using the inequality \eqref{second step for independence} proves the claim. \quad $\blacksquare$

\subsection{Equivalence of FMP and differentiability in filters and functionals} 
\label{Equivalence of FMP and differentiability in filters and functionals} 

The facts established in  Lemma \ref{smooth projections and time delays} can be used to show the equivalence between the continuity and the differentiability of causal and time-invariant filters and that of their associated functionals. The following result focuses on continuity and the fading memory property and generalizes to the context of eventually unbounded inputs the equivalence between fading memory filters and functionals established in \cite[Propositions 2.11 and 2.12]{RC7} for uniformly bounded inputs. In the results that follow we work in a setup slightly more general than the one that is customary in the literature as we will allow for the weighting sequences considered in the domain and the target of the filters to be different. This degree of generality is needed later on in the text.

\begin{proposition}
\label{continuity of filters and functionals}
Let $V _n\subset \left({\Bbb R}^n\right)^{\mathbb{Z}_{-}} $ and $V _N\subset \left({\Bbb R}^N\right)^{\mathbb{Z}_{-}} $ be time-invariant subsets and let $D_N\subset \mathbb{R}^N $. Let $w^1, w^2 $ be weighting sequences with inverse decay ratios $L _{w^1} $ and $L _{w^2} $, respectively.
\begin{description}
\item [(i)]  Let $U: V _n \subset \ell_{-} ^{w^1}(\mathbb{R}^n) \longrightarrow V _N \subset \ell_{-} ^{w^2}(\mathbb{R}^N)$ be a causal and time-invariant filter. If $U$ has the fading memory property then so does its associated functional $H _U:V _n \longrightarrow p _0(V _N) $. The same conclusion holds for continuous filters $U: V _n \subset \ell_{-} ^{\infty}(\mathbb{R}^n) \longrightarrow V _N \subset \ell_{-} ^{\infty}(\mathbb{R}^N)$.
\item [(ii)] Let $H :V _n\subset \ell_{-} ^{w^1}(\mathbb{R}^n) \longrightarrow D_N$ be a fading memory functional. If $ L_{w^1} $ is finite and $D_N $ is compact then the associated causal and time-invariant filter $U_H :V _n\subset \ell_{-} ^{w^1}(\mathbb{R}^n) \longrightarrow (D_N) ^{\mathbb{Z}_{-}} \subset \ell_{-} ^{w^2}(\mathbb{R}^N)$ has also the fading memory property.
\item [(iii)] Let $H :V _n\subset \ell_{-} ^{w^1}(\mathbb{R}^n) \longrightarrow D_N$  be a fading memory functional and suppose that $V _n  $ contains a point ${\bf z}^0 $ such that $U_H({\bf z}^0)\in \ell_{-}^{w^2}(\mathbb{R}) $, where $U _H $ is the causal and time-invariant filter associated to $H$. If $H $ is Lipschitz, $c_H$ is a Lipschitz constant, and the weighting sequences satisfy one of the following two conditions
\begin{equation}
\label{rw definition}
\mbox{either\ }  R_{w^1, w ^2}:=\sup_{s, t \in \mathbb{N}}\left\{\frac{w^1 _t w^2 _s}{w^1_{t+s}}\right\}< + \infty \mbox{\ or the sequence $\mathcal{L}_{w^1}:=\left(L _{w^1}^{-t}\right)_{t  \in \mathbb{Z}_{-}} \in \ell_{-}^{w^2}(\mathbb{R})$,}
\end{equation}
then $U_H :V _n\subset \ell_{-} ^{w^1}(\mathbb{R}^n) \longrightarrow (D_N) ^{\mathbb{Z}_{-}} \cap  \ell_{-} ^{w^2}(\mathbb{R}^N)$ has also the fading memory property, it is Lipschitz, and $R _{w^1, w ^2} c_H $ or $\|\mathcal{L} _{w^1}\| _{w ^2}  c_H $, respectively, is a Lipschitz constant of $U _H $. The same conclusion holds for continuous functionals $H: V _n \subset \ell_{-} ^{\infty}(\mathbb{R}^n) \longrightarrow D_N $ where the condition \eqref{rw definition} is not needed.
\end{description}
\end{proposition}

\begin{remark}
\normalfont
When in part {\bf (iii)} we consider the same weighting sequence $w$ for the domain and the target, it is easy to see that
\begin{equation*}
R_{w }:=\sup_{s, t \in \mathbb{N}}\left\{\frac{w  _t w  _s}{w _{t+s}}\right\} 
\end{equation*}
satisfies that $R _w\leq \left\|\mathcal{L}_w\right\|_w $ and therefore the second condition in \eqref{rw definition} implies the first one. Indeed,
\begin{equation*}
R_{w }=\sup_{s, t \in \mathbb{N}}\left\{\frac{w  _t w  _s}{w _{t+s}}\right\}=
\sup_{s, t \in \mathbb{N}}\left\{\frac{w _t}{w_{t+1}}\frac{w_{t+1}}{w_{t+2}}\cdots\frac{w_{t+s-1}}{w_{t+s}}w _s \right\} \leq \sup_{s\in \mathbb{N}}\left\{L _w^s w _s
\right\}=\left\|\mathcal{L}_w\right\|_w, \quad \mbox{as required.} 
\end{equation*}
In this setup, the condition \eqref{rw definition} is satisfied by many families of commonly used weighting sequences.  In the two examples considered in  Remark \ref{examples for lw} we have that $R _w=\left\|\mathcal{L}_w\right\|_w=1 $ for the geometric sequence; for the harmonic sequence $\left\|\mathcal{L}_w\right\|_w=+ \infty $ but $R _w=1 $ and hence \eqref{rw definition} is still satisfied.

We emphasize that condition \eqref{rw definition} is not automatically satisfied by all weighting sequences. For example, as we saw in Remark \ref{examples for lw},  the sequence $w _t:=\exp(-t^2) $ is such that  $L _w=+ \infty $ and, additionally, it is easy to see that $R_{w }:=\sup_{s, t \in \mathbb{N}}\left\{\exp(2st)\right\} =+ \infty$.
\end{remark}

\begin{proposition}
\label{diff of filters and functionals}
Let $w ^1 $ and $w ^2 $  be two weighting sequences with inverse decay ratios $L _{w^1} $ and $L _{w^1} $, respectively. Let $V _n\subset \ell ^{w^1}_-({\Bbb R}^n)$ and $V _N\subset \ell ^{w^2}_-({\Bbb R}^N)$ be time-invariant open subsets, and let $D_N$ be an open subset of $\mathbb{R}^N  $. 
\begin{description}
\item [(i)] Let $U: V _n \subset \ell_{-} ^{w^1}(\mathbb{R}^n) \longrightarrow V _N \subset \ell_{-} ^{w^2}(\mathbb{R}^N)$ be a causal and time-invariant filter. If $U$ is of class $C ^r(V _n) $ (respectively, smooth or analytic) when considered as a map ${U}: V _n\subset \left(\ell_{-} ^{w^1}(\mathbb{R}^n), \left\|\cdot \right\|_w\right) \longrightarrow  V _N\subset \left(\ell_{-} ^{w^2}(\mathbb{R}^N), \left\|\cdot \right\|_w\right)$, then so is the associated functional  $H_{U}: V _n\subset \left(\ell_{-} ^{w^1}(\mathbb{R}^n), \left\|\cdot \right\|_w\right) \longrightarrow p _0(V _N) \subset \mathbb{R}^N$.
Moreover,
\begin{equation}
\label{inequality norms derivatives h}
\vertiii{D ^rH _U({\bf z})}_{w^1}\leq \vertiii{D ^rU({\bf z})}_{w ^1,w^2}, \quad \mbox{for any $ {\bf z} \in V _n $.}
\end{equation}
The same conclusion holds when the  weighted sequence spaces are replaced by $\left(\ell_{-} ^{\infty}(\mathbb{R}^n), \left\|\cdot \right\|_\infty\right)$ and $\left(\ell_{-} ^{\infty}(\mathbb{R}^N), \left\|\cdot \right\|_\infty\right)$.
\item [(ii)] Let $H :V _n\subset \ell_{-} ^{w^1}(\mathbb{R}^n) \longrightarrow D_N$ be a functional and suppose that $V _n  $ is convex and contains a point ${\bf z}^0 $ such that $U_H({\bf z}^0)\in \ell_{-}^{w^2}(\mathbb{R}) $, where $U _H $ is the causal and time-invariant filter associated to $H$.  If the functional $H$ is of class $C ^r(V _n) $ and for  any $j \in \left\{1, \ldots, r\right\}$  we have that $c ^j:= \sup_{{\bf z} \in V _n} \left\{\vertiii{D^jH ({\bf z})}_{w^1}\right\}< + \infty$ and the weighting sequences satisfy that 
\begin{equation}
\label{differentiability condition for sequences}
\mathcal{L}_{w^1,j}:=(L _{w^1}^{-jt})_{t  \in \mathbb{Z}_{-}} \in \ell_{-}^{w^2}(\mathbb{R}),
\end{equation}
then the associated causal and time-invariant filter  $U _H $ is differentiable of order $r$ when considered as a map $U _H: V _n\subset \left(\ell_{-} ^{w^1}(\mathbb{R}^n), \left\|\cdot \right\|_{w^1}\right) \longrightarrow (D _N) ^{\mathbb{Z}_{-}}\cap  \left(\ell_{-} ^{w^2}(\mathbb{R}^N), \left\|\cdot \right\|_{w^2} \right)$. Moreover, for any $ {\bf z} \in V _n $,
\begin{equation}
\label{inequality norms derivatives Uh}
\vertiii{D ^rU _H ({\bf z})}_{w^1, w ^2}\leq  c ^r \|\mathcal{L}_{w^1,r}\| _{w^2}. 
\end{equation}
Additionally, $ {U _H} $ is of class $C^{r-1}(V _n) $ and the map
$$D ^{r-1}  {U _H}: (V _n, \left\|\cdot \right\|_{w^1}) \longrightarrow \left(L ^{r-1}\left(\ell_{-} ^{w^1}(\mathbb{R}^n), \ell_{-} ^{w^2}(\mathbb{R}^N)\right ), \vertiii{\cdot }_{w ^1,w^2}\right)  $$ is Lipschitz continuous with Lipschitz constant $c ^r \|\mathcal{L}_{w^1,r}\| _{w^2}$.
The same conclusion holds when the  weighted sequence spaces are replaced by $\left(\ell_{-} ^{\infty}(\mathbb{R}^n), \left\|\cdot \right\|_\infty\right)$ and $\left(\ell_{-} ^{\infty}(\mathbb{R}^N), \left\|\cdot \right\|_\infty\right)$. In that case  the inequality \eqref{inequality norms derivatives Uh} holds with $\|\mathcal{L}_{w^1,r}\| _{w^2}=1 $.
\item [(iii)] Let $H :V _n\subset \ell_{-} ^{w^1}(\mathbb{R}^n) \longrightarrow D_N$ be a functional and suppose that $V _n  $ is convex and contains a point ${\bf z}^0 $ such that $U_H({\bf z}^0)\in \ell_{-}^{w^2}(\mathbb{R}) $, where $U _H $ is the causal and time-invariant filter associated to $H$.  If the functional $H$ is smooth  and $c ^r< + \infty$ for all $r \in \mathbb{N}^+ $, 
then so is the associated causal and time-invariant filter  $U _H: V _n\subset \left(\ell_{-} ^{w^1}(\mathbb{R}^n), \left\|\cdot \right\|_{w^1}\right) \longrightarrow (D _N) ^{\mathbb{Z}_{-}}\cap \left(\ell_{-} ^{w^2}(\mathbb{R}^N), \left\|\cdot \right\|_{w^2}\right)$. The same conclusion holds when the  weighted spaces are  replaced by $\left(\ell_{-} ^{\infty}(\mathbb{R}^n), \left\|\cdot \right\|_\infty\right)$ and $\left(\ell_{-} ^{\infty}(\mathbb{R}^N), \left\|\cdot \right\|_\infty\right)$. In that case, if $H$ is analytic then so is $U _H $ and the radius of convergence of the series expansion of $U _H $ is bigger or equal than that of $H$. 
\end{description}
\end{proposition}

\begin{remark}
\label{higher order derivatives bad}
\normalfont
An important consequence of part {\bf (ii)} in this proposition and, in particular, of the condition \eqref{differentiability condition for sequences} is that, in general, one cannot obtain (higher order) differentiable filters out of differentiable functionals using the same weighted norm in the domain and the target of the filter. The weighted norm in the target needs to be chosen so that it satisfies the nonautomatic condition  \eqref{differentiability condition for sequences} that, additionally, depends on the differentiability degree that we want to preserve. Weighted norms that satisfy that property are relatively easy to find in most cases. For example, if we take as $w ^1 $ the geometric sequence in Remark \ref{examples for lw}, then $\mathcal{L}_{w^1, j}= \left(\lambda^{-jt}\right) _{t \in \mathbb{N}} $ and hence condition \eqref{differentiability condition for sequences} is satisfied if we take as $w ^2 $ any sequence of the type $\left(w ^1\right)^r$ (using the notation in Lemma \ref{inclusions with lw}) with $r\geq j $.
\end{remark}

\section{The fading memory property in reservoir filters with unbounded inputs}
\label{The fading memory property in reservoir filters with unbounded inputs}

Starting in this section we focus on filters defined by reservoir systems of the type introduced in~\eqref{reservoir equation}--\eqref{readout}, but this time we consider  reservoir maps $F:  D _N \times  D _n \longrightarrow  D_N$  where the input variable takes values on a set $D_n\subset \mathbb{R}^N$ that is not necessarily bounded. All along this section, the reservoir map $F$ will be assumed to be continuous and a contraction on the first entry with constant $0<c<1 $, that is, 
\begin{equation*}
\left\|F(\mathbf{x}^1, {\bf z})-F(\mathbf{x}^2 , {\bf z})\right\|\leq c \left\|\mathbf{x}^1 - \mathbf{x}^2\right\|, \quad \mbox{for all $\mathbf{x}^1, \mathbf{x}^2 \in D _N$ and ${\bf z} \in D _n $.}
\end{equation*}

When the inputs are assumed to be uniformly bounded by a constant $M>0 $ and $F$ maps into a ball $\overline{B_{\left\|\cdot \right\|}({\bf 0}, L)}\subset \mathbb{R}^N $, $L>0 $, it has been proved (see \cite[Proposition 2.1 and Theorem 3.1]{RC7}) that we can associate to this system
unique filters $U ^F: K _M \longrightarrow K _L$ and $U ^F_h: K _M \longrightarrow ({\Bbb R}^d) ^{\mathbb{Z}_{-}}$ (the sets $K _M $ and $K _L $ are introduced in \eqref{Kset}) that are causal, time-invariant, continuous and, moreover, satisfy the fading memory property with respect to any weighting sequence $w  $. We recall that $U ^F $ is the filter associated to the solutions of the reservoir equation \eqref{reservoir equation} and assigns to any input sequence  ${\bf z} \in K _M $ the output $U ^F({\bf z})$ that satisfies 
\begin{equation}
\label{relation function filter U}
U^F({\bf z})_t =F(U^F( {\bf z}) _{t-1}, {\bf z} _t), \quad \mbox{for any} \quad t \in \mathbb{Z}_{-}.
\end{equation}
Recall also that $U ^F_h: K _M \longrightarrow ({\Bbb R}^d) ^{\mathbb{Z}_{-}} $ is the filter associated to the full system~\eqref{reservoir equation}--\eqref{readout} and is given by $U ^F_h:=h \circ U ^F $.
We  denote by $H ^F: K _M \longrightarrow \overline{B_{\left\|\cdot \right\|}({\bf 0}, L)} $ and $H ^F _h: K _M \longrightarrow \mathbb{R}^d$ the corresponding reservoir functionals. The reservoir functionals are related to the corresponding reservoir filters via the identities:
\begin{equation}
\label{relation function filter H}
H ^F({\bf z}) =U^F({\bf z})_0=F(U^F( {\bf z}) _{-1}, {\bf z} _0) \quad \mbox{and} \quad H ^F _h({\bf z})=h \left(U ^F({\bf z})\right),
\end{equation}
for all ${\bf z} \in K _M $.

The next theorem is the most important result in this section and shows that the results that we just recalled about the ESP and the FMP for reservoir filters with uniformly bounded inputs remain valid in the presence of unbounded inputs. However, in that case, the fading memory property depends on the weighting sequence that is used to define it. The sufficient condition for the FMP spelled out in the next theorem asserts, roughly speaking, that reservoir systems have the FMP only with respect to weighting sequences that converge to zero faster than the divergence rate of their outputs.

\begin{theorem}[ESP and FMP with continuous reservoir maps]
\label{characterization of fmp unbounded}
Let $F:  D _N \times  D _n \longrightarrow  D_N$  be a continuous reservoir map where $D _n \subset {\Bbb R}^n $, $D _N\subset {\Bbb R}^N $, $n, N  \in \mathbb{N}^+$. Assume, additionally, that it is a contraction on the first entry with constant $0<c<1 $. Let $w$ be a weighting sequence with finite inverse decay ratio $L _w$ and let $V _n \subset (D_n) ^{\mathbb{Z}_{-}} \cap \ell_{-}^{w}(\mathbb{R}^n)$ be a time-invariant set. We consider two situations regarding the target $D_N $ of the reservoir map:
\begin{description}
\item [(i)] $D_N  $ is a compact subset of ${\Bbb R}^N $.
\item [(ii)]  $(D_N)^{\mathbb{Z}_{-}}\cap \ell_{-}^{w}(\mathbb{R}^N)  $ is a complete subset of the Banach space $ \left( \ell_{-}^{w}(\mathbb{R}^N), \left\|\cdot \right\|_w\right) $, $F$ is Lipschitz continuous, and the reservoir system \eqref{reservoir equation} associated to $F$ has a solution $(\mathbf{x}^0, {\bf z}^0) \in  (D_N)^{\mathbb{Z}_{-}}\cap \ell_{-}^{w}(\mathbb{R}^N) \times V _n$, that is, $\mathbf{x}_t^0=F(\mathbf{x}_{t-1}^0, {\bf z}_t ^0)$, for all $t \in \mathbb{Z}_{-} $
\end{description}
In both cases, if
\begin{equation}
\label{fmp condition on w}
c L _w<1
\end{equation}
then the reservoir system associated to $F$ with inputs in $V _n $ has the echo state property and hence determines a unique continuous, causal, and time-invariant reservoir filter 
$
U^F:(V _n, \left\|\cdot \right\|_w) \longrightarrow ((D_N)^{\mathbb{Z}_{-}}\cap \ell_{-}^{w}(\mathbb{R}^N), \left\|\cdot \right\|_w)
$
that has the fading memory property with respect to $w$. Moreover, if $F$ is Lipschitz on the second component (which is always the case under the hypotheses in {\bf (ii)}) with constant $L _z $, that is,
\begin{equation*}
\left\|F(\mathbf{x}, {\bf z} ^1)-F(\mathbf{x}, {\bf z} ^2)\right\|\leq L _z \left\|{\bf z} ^1- {\bf z} ^2\right\|, \quad \mbox{for any} \quad \mathbf{x} \in D_N, \ {\bf z} ^1, {\bf z} ^2 \in D_n,
\end{equation*}
then $U ^F $ is also Lipschitz with constant
\begin{equation}
\label{lips in case continuous Lz}
L_{U ^F}:= \frac{L _z}{1-c L _w}.
\end{equation}

This statement also holds true under the hypotheses in part {\rm {\bf (ii)}} when $(\ell_{-} ^{w}(\mathbb{R}^n), \left\|\cdot \right\|_w) $ is replaced by $(\ell_{-} ^{\infty}(\mathbb{R}^n), \left\|\cdot \right\|_\infty) $. In that case  $L _w  $ is replaced by the constant $1$ and hence condition \eqref{fmp condition on w} is automatically satisfied. The resulting reservoir filter 
$
U^F:(V _n, \left\|\cdot \right\|_\infty) \longrightarrow ((D_N)^{\mathbb{Z}_{-}}\cap \ell_{-}^{\infty}(\mathbb{R}^N), \left\|\cdot \right\|_w)
$ is continuous.
\end{theorem}

\begin{remark}
\normalfont
A very common situation that provides the solution $(\mathbf{x}^0, {\bf z}^0) \in  (D_N)^{\mathbb{Z}_{-}}\cap \ell_{-}^{w}(\mathbb{R}^N) \times V _n$ for the reservoir system needed in part {\bf (ii)}, is the existence of a fixed point $(\overline{\mathbf{x}}^0,\overline{ {\bf z}}^0) \in D _N \times  D _n $ of $F$ that satisfies  $F(\overline{\mathbf{x}}^0, \overline{{\bf z}}^0)=\overline{\mathbf{x}}^0 $. In that case the required solution is given by the constant sequences $\mathbf{x}^0 _t= \overline{\mathbf{x}} ^0 $, $\mathbf{z}^0 _t= \overline{\mathbf{z}} ^0 $, for all $t \in \mathbb{Z}_{-}$.
\end{remark}

\begin{remark}
\normalfont
If the target $D_N  $ of the reservoir map is a closed subset of ${\Bbb R}^N $, that is $\overline{D_N} = D _N $,   then by part {\bf (iii)} of the Corollary \ref{dnn closed and open}, the set $(D_N)^{\mathbb{Z}_{-}}\cap \ell_{-}^{w}(\mathbb{R}^N)  $ is a closed subset of  $ \left( \ell_{-}^{w}(\mathbb{R}^N), \left\|\cdot \right\|_w\right) $ and it is hence necessarily complete.
\end{remark}

\noindent\textbf{Proof of the theorem.\ \ } Consider the map 
\begin{equation}
\label{script f as product definition}
\begin{array}{cccc}
\mathcal{F}: & (D_N)^{\mathbb{Z}_{-}}\cap \ell_{-}^{w}(\mathbb{R}^N) \times V _n&\longrightarrow &(D_N)^{\mathbb{Z}_{-}}\\
	&(\mathbf{x}, {\bf z})&\longmapsto & \left(\mathcal{F}(\mathbf{x}, {\bf z})\right)_t:=F(\mathbf{x} _{t-1}, {\bf z}_t).
\end{array}
\end{equation}
We now show that first, under the two sets of hypotheses in the statement,  $\mathcal{F} $ actually maps into $(D_N)^{\mathbb{Z}_{-}}\cap \ell_{-}^{w}(\mathbb{R}^N) $ and second, that $\mathcal{F} $ is continuous. Suppose first that we are in the hypotheses in {\bf (i)}. Since $D_N  $ is compact then $(D_N)^{\mathbb{Z}_{-}}\subset  \ell_{-}^{w}(\mathbb{R}^N) $ and hence $\mathcal{F} $ obviously maps into $(D_N)^{\mathbb{Z}_{-}}\cap \ell_{-}^{w}(\mathbb{R}^N) $. Regarding the continuity, notice  that $\mathcal{F} $ can be written as 
\begin{equation}
\label{script f as product}
\mathcal{F}=\prod_{t \in \mathbb{Z}_{-}} F  _t  \quad \mbox{with} \quad F _t:= F \circ p _t \circ  \left(T _1\times {\rm id}_{V _n}\right): (D_N)^{\mathbb{Z}_{-}}\cap \ell_{-}^{w}(\mathbb{R}^N) \times V _n\longrightarrow  D _N. 
\end{equation}
The continuity of $F$, the fact that $L _w $ is by hypothesis finite,  and Lemma \ref{smooth projections and time delays} imply that all the functions $F _t :(D_N)^{\mathbb{Z}_{-}}\cap \ell_{-}^{w}(\mathbb{R}^N) \times V _n \subset  \ell_{-} ^{w}(\mathbb{R}^N)\oplus \ell_{-} ^{w}(\mathbb{R}^n) \longrightarrow D _N \subset {\Bbb R}^N$ are continuous and moreover,  they map into a compact subset of 
$\mathbb{R}^N $. An argument mimicking the proof of the first part of Lemma \ref{cr using product maps} allows us to conclude that $\mathcal{F}:  (D_N)^{\mathbb{Z}_{-}}\cap \ell_{-}^{w}(\mathbb{R}^N) \times V _n \longrightarrow  (D_N)^{\mathbb{Z}_{-}}\cap \ell_{-}^{w}(\mathbb{R}^N)$ is a continuous map. 

Suppose now that we are in the hypotheses in part {\bf (ii)}. We now show that since $F$ is Lipschitz then so are all the functions $F_t:= F \circ p _t \circ  \left(T _1\times {\rm id}_{V _n}\right) $, $t \in \mathbb{Z}_{-}$, by Lemma \ref{smooth projections and time delays}, where we  consider the direct sum 
of weighted spaces $\ell_{-} ^{w}(\mathbb{R}^N)\oplus \ell_{-} ^{w}(\mathbb{R}^n) $ as a Banach space with the sum norm $\left\|\cdot \right\|_{w \oplus w} $ defined by $\left\|(\mathbf{x}, \mathbf{z}) \right\|_{w \oplus w} :=\left\|\mathbf{x} \right\|_{w}+\left\|\mathbf{z} \right\|_{w}$, for any $(\mathbf{x}, \mathbf{z}) \in \ell_{-} ^{w}(\mathbb{R}^N)\oplus \ell_{-} ^{w}(\mathbb{R}^n) $. Indeed, let $c_F $ be the Lipschitz constant of $F$  and let $(\mathbf{x}^1, \mathbf{z}^1), (\mathbf{x}^2, \mathbf{z}^2) \in   (D_N)^{\mathbb{Z}_{-}}\cap \ell_{-}^{w}(\mathbb{R}^N) \times V _n $, then:
\begin{multline}
\label{mathcalF is lipschitz1}
\left\|F \circ p _t \circ  \left(T _1\times {\rm id}_{V _n}\right)(\mathbf{x}^1, \mathbf{z}^1)- F \circ p _t \circ  \left(T _1\times {\rm id}_{V _n}\right)(\mathbf{x}^2, \mathbf{z}^2)\right\|
\leq c_F \left\|p _t \circ  \left(T _1\times {\rm id}_{V _n}\right)(\mathbf{x}^1-\mathbf{x}^2, \mathbf{z}^1-\mathbf{z}^2)\right\|\\\leq
\frac{c_F}{w _{-t}}
\left\|\left(T _1\times {\rm id}_{V _n}\right)(\mathbf{x}^1-\mathbf{x}^2, \mathbf{z}^1-\mathbf{z}^2)\right\|_w
\leq \frac{c_F}{w _{-t}} (L _w \left\|\mathbf{x}^1-\mathbf{x}^2\right\|_w + \left\|\mathbf{z}^1-\mathbf{z}^2\right\|_w )\\
\leq 
\frac{c_F}{w _{-t}}L _w(\left\|\mathbf{x}^1-\mathbf{x}^2\right\| _w+ \left\|\mathbf{z}^1-\mathbf{z}^2\right\| _w)=\frac{c_F}{w _{-t}} L _w \left\|(\mathbf{x}^1,\mathbf{z}^1)-( \mathbf{x}^2,\mathbf{z}^2)\right\|_{w\oplus w}.
\end{multline}
This chain of inequalities show that $F _t $ is a Lipschitz continuous function and that ${c_F L _w}/{w _{-t}} $ is a Lipschitz constant. Given that the sequence $c _{\mathcal{F}}:=({c_F L _w}/{w _{-t}})_{t \in \mathbb{Z}_{-}}$ is such that $\left\|c _{\mathcal{F}}\right\|_w =c_F L _w<+ \infty$, the part {\bf (ii)} of Lemma \ref{cr using product maps} guarantees that $\mathcal{F} $ is Lipschitz continuous and that $c_F L _w $ is a Lipschitz constant, that is,
\begin{equation}
\label{mathcalF is lipschitz2}
\left\|\mathcal{F} (\mathbf{x}^1, \mathbf{z}^1)-\mathcal{F} (\mathbf{x}^2, \mathbf{z}^2)\right\|_w\leq\ c_F  L _w \left\|(\mathbf{x}^1,\mathbf{z}^1)-( \mathbf{x}^2,\mathbf{z}^2)\right\|_{w\oplus w}.
\end{equation}
Moreover, let ${\bf u}^0 :=(\mathbf{x} ^0, {\bf z}^0)\in(D_N)^{\mathbb{Z}_{-}}\cap \ell_{-}^{w}(\mathbb{R}^N) \times V _n$. The fact that ${\bf u}^0 $ is a solution of the reservoir system implies that  $\mathcal{F}({\bf u}^0)= \mathbf{x} ^0 \in (D_N)^{\mathbb{Z}_{-}}\cap \ell_{-}^{w}(\mathbb{R}^N)$. An argument mimicking \eqref{maps into lw} in the proof of part {\bf (ii)} in Lemma \ref{cr using product maps} proves that in those conditions $\mathcal{F} $ maps into $(D_N)^{\mathbb{Z}_{-}}\cap \ell_{-}^{w}(\mathbb{R}^N)$.

We now show that in the presence of hypothesis \eqref{fmp condition on w} $\mathcal{F}  $ is a contraction on the first entry with constant $c L _w<1 $. Indeed, for any $\mathbf{x}^1, {\bf x}^2 \in (D_N)^{\mathbb{Z}_{-}}\cap \ell_{-}^{w}(\mathbb{R}^N)$ and any ${\bf z} \in V _n $, we have
\begin{equation}
\label{contraction part 1}
\left\|\mathcal{F}(\mathbf{x}^1, {\bf z})-\mathcal{F}(\mathbf{x}^2 , {\bf z})\right\|_w=
\sup_{t \in \mathbb{Z}_{-}}\left\{\left\|F(\mathbf{x}_{t-1}^1, {\bf z}_t)-F(\mathbf{x}_{t-1}^2, {\bf z}_t)\right\|w_{-t}\right\}\leq 
\sup_{t \in \mathbb{Z}_{-}}\left\{\left\|\mathbf{x}_{t-1}^1-\mathbf{x}^2_{t-1}\right\|c w_{-t}\right\},
\end{equation}
where we used that $F$ is a contraction on the first entry. Now, 
\begin{equation}
\label{contraction part 2}
\sup_{t \in \mathbb{Z}_{-}}\left\{\left\|\mathbf{x}_{t-1}^1-\mathbf{x}^2_{t-1}\right\|c w_{-t}\right\}=
c\sup_{t \in \mathbb{Z}_{-}}\left\{\left\|\mathbf{x}_{t-1}^1-\mathbf{x}^2_{t-1}\right\| w_{-(t-1)}\frac{w _{-t}}{w _{-(t-1)}}\right\}\leq
c L _w \left\|\mathbf{x}^1- \mathbf{x}^2 \right\|_w.
\end{equation}
This shows that $\mathcal{F} $ is a family of contractions with constant $c L _w<1 $ that is continuously parametrized by the elements in $V_n$. Since by hypothesis, the domain $(D_N)^{\mathbb{Z}_{-}}\cap \ell_{-}^{w}(\mathbb{R}^N)$ is complete, Theorem 6.4.1 in \cite{Sternberg:dynamical:book} implies the existence of a continuous map $U^F: \left(V _n, \left\|\cdot \right\|_w\right)\longrightarrow\left((D_N)^{\mathbb{Z}_{-}}\cap \ell_{-}^{w}(\mathbb{R}^N), \left\|\cdot \right\|_w\right) $ that is uniquely determined by the identity
\begin{equation}
\label{defining for filter continuous}
\mathcal{F} \left(U^F({\bf z}), {\bf z}\right)=U^F({\bf z}), \quad \mbox{for all ${\bf z}\in V _n  $}.
\end{equation}
The causality and the time-invariance of $U^F  $ are a consequence of the time invariance of $V _n $ and of Proposition 2.1 in \cite{RC6}. 

We now assume that  $F$ is Lipschitz on the second component and prove \eqref{lips in case continuous Lz}. The relation \eqref{defining for filter continuous} that defines $U ^F $  is equivalent to
\begin{equation*}
U ^F({\bf z})_t=F(U ^F({\bf z})_{t-1}, {\bf z} _t), \quad \mbox{for all} \quad {\bf z} \in V _n,\  t \in \mathbb{Z}_{-}. 
\end{equation*}
Consequently, for any ${\bf z} ^1, {\bf z}^2 \in V _n$, we have,
\begin{multline*}
\left\|U ^F({\bf z}^1)-U ^F({\bf z}^2)\right\|_w=\sup_{t \in \mathbb{Z}_{-}} \left\{\left\|U ^F({\bf z}^1)_t-U ^F({\bf z}^2)_t\right\|w _{-t}\right\}\\
=\sup_{t \in \mathbb{Z}_{-}} \left\{\left\|F(U ^F({\bf z}^1)_{t-1}, {\bf z}^1_t)-F(U ^F({\bf z}^2)_{t-1}, {\bf z}^2_t)\right\|w _{-t}\right\}\\
\leq \sup_{t \in \mathbb{Z}_{-}} \left\{\left( \left\|F(U ^F({\bf z}^1)_{t-1}, {\bf z}^1_t)-F(U ^F({\bf z}^1)_{t-1}, {\bf z}^2_t)\right\| + \left\|F(U ^F({\bf z}^1)_{t-1}, {\bf z}^2_t)-F(U ^F({\bf z}^2)_{t-1}, {\bf z}^2_t)\right\| \right)w _{-t}\right\}\\
\leq \sup_{t \in \mathbb{Z}_{-}} \left\{
L _z \left\|{\bf z} ^1_t-{\bf z} ^2_t \right\|w _{-t}+ c \left\|U ^F({\bf z}^1)_{t-1}-U ^F({\bf z}^2)_{t-1}\right\| w _{-t}
\right\}.
\end{multline*}
If we repeat this procedure $i$ times, it is easy to see that
\begin{multline}
\label{to be chained 0}
\left\|U ^F({\bf z}^1)-U ^F({\bf z}^2)\right\|_w\\
\leq L _z\sup_{t \in \mathbb{Z}_{-}} \left\{
 \sum_{j=0}^{i}c ^j\left\|{\bf z} ^1_{t-j}-{\bf z} ^2_{t-j} \right\|w _{-t}\right\}+ c^{i+1} \sup_{t \in \mathbb{Z}_{-}} \left\{\left\|U ^F({\bf z}^1)_{t-(i+1)}-U ^F({\bf z}^2)_{t-(i+1)}\right\| w _{-t}
\right\}.
\end{multline}
We now study separately the two summands in the right hand side of the previous inequality. First, by Lemma \ref{smooth projections and time delays},
\begin{multline}
\label{to be chained 1}
L _z\sup_{t \in \mathbb{Z}_{-}} \left\{
 \sum_{j=0}^{i}c ^j\left\|{\bf z} ^1_{t-j}-{\bf z} ^2_{t-j} \right\|w _{-t}\right\}=
L _z\sup_{t \in \mathbb{Z}_{-}} \left\{
 \sum_{j=0}^{i}c ^j\left\|\left(T _j({\bf z} ^1)\right)_{t}-\left(T _j({\bf z} ^2)\right)_{t} \right\|w _{-t}\right\}\\
=L _z\sup_{t \in \mathbb{Z}_{-}} \left\{
 \sum_{j=0}^{i}c ^j\left\|T _j({\bf z} ^1-{\bf z} ^2)_{t} \right\|w _{-t}\right\}
\leq L _z \sum_{j=0}^{i}c ^j\sup_{t \in \mathbb{Z}_{-}} \left\{
 \left\|T _j({\bf z} ^1-{\bf z} ^2)_{t} \right\|w _{-t}\right\}\\
=L _z \sum_{j=0}^{i}c ^j
 \left\|T _j({\bf z} ^1-{\bf z} ^2) \right\|_w
\leq L _z \left\|{\bf z} ^1-{\bf z} ^2 \right\|_w\sum_{j=0}^{i}c ^j \vertiii{T _j}_w\\
\leq L _z\left\|{\bf z} ^1-{\bf z} ^2 \right\|_w \sum_{j=0}^{i}(c  L _w)^j= L _z\left\|{\bf z} ^1-{\bf z} ^2 \right\|_w \frac{1-(c  L _w)^{i+1}}{1-c  L _w}, 
\end{multline}
while the second summand can be bounded as follows
\begin{multline}
\label{to be chained 2}
\!\!\!\!\!\!\!c^{i+1} \sup_{t \in \mathbb{Z}_{-}} \left\{\left\|U ^F({\bf z}^1)_{t-(i+1)}-U ^F({\bf z}^2)_{t-(i+1)}\right\| w _{-t}
\right\} =c^{i+1} \sup_{t \in \mathbb{Z}_{-}} \left\{\left\|T_{i+1 }\left(U ^F({\bf z}^1)\right)_{t}-T_{i+1 }\left(U ^F({\bf z}^2)\right)_{t}\right\| w _{-t}
\right\} \\
=c^{i+1} \left\|T_{i+1 }(U ^F({\bf z}^1)-(U ^F({\bf z}^2)) \right\| _w
\leq c^{i+1}  \vertiii{T_{i+1 } } _w \left\|U ^F({\bf z}^1)-U ^F({\bf z}^2) \right\| _w\\
\leq (c L _w)^{i+1}\left\|U ^F({\bf z}^1)-U ^F({\bf z}^2)\right\| _w.
\end{multline}
If we now chain the inequalities \eqref{to be chained 1} and \eqref{to be chained 2} with \eqref{to be chained 0} we can conclude that
\begin{equation}
\label{inequality for subsequent limit}
(1-(c L _w)^{i+1})\left\|U ^F({\bf z}^1)-U ^F({\bf z}^2)\right\| _w\leq L _z\left\|{\bf z} ^1-{\bf z} ^2 \right\|_w \frac{1-(c  L _w)^{i+1}}{1-c  L _w},
\end{equation}
which after simplification using the condition \eqref{fmp condition on w} results in \eqref{lips in case continuous Lz}.
\quad $\blacksquare$ 

\begin{remark}
\label{extension of characterization of fmp to pw}
\normalfont
A slight modification of this proof can be used to extend the statement of Theorem \ref{characterization of fmp unbounded} {\bf (ii)} to reservoir systems with inputs and outputs in $\ell_{-}^{p,w}(\mathbb{R}^n) $ and $\ell_{-}^{p,w}(\mathbb{R}^N) $, respectively. Indeed, assume that we are under the hypotheses of Theorem \ref{characterization of fmp unbounded} {\bf (ii)} with those spaces instead of $\ell_{-}^{w}(\mathbb{R}^n) $ and $\ell_{-}^{w}(\mathbb{R}^N) $. Suppose, additionally, that
\begin{equation}
\label{fmp condition for pw spaces}
c L_{w,p}<1
\end{equation}
where $L_{w,p}$ was defined in \eqref{inverse decay ratio for p}. Then, there exists a unique causal and time-invariant continuous reservoir filter $
U^F:(V _n, \left\|\cdot \right\|_{p,w}) \longrightarrow ((D_N)^{\mathbb{Z}_{-}}\cap \ell_{-}^{p, w}(\mathbb{R}^N), \left\|\cdot \right\|_{p,w})
$. Additionally, $U ^F $ is also Lipschitz with constant
\begin{equation*}
L_{U ^F}:= \frac{L _z}{1-c L _{w,p}}.
\end{equation*}
The proof of this fact is carried out by showing that the map $\mathcal{F} $ in \eqref{script f as product} is Lipschitz continuous when $\ell_{-}^{p,w}(\mathbb{R}^n) $  and $\ell_{-}^{p,w}(\mathbb{R}^N) $ spaces are considered in its domain and target, respectively, with Lipschitz constant $c _F L _{w,p} $ and hence \eqref{mathcalF is lipschitz2} holds in that situation. Indeed,  for any $(\mathbf{x}^1, \mathbf{z}^1), (\mathbf{x}^2, \mathbf{z}^2) \in   (D_N)^{\mathbb{Z}_{-}}\cap \ell_{-}^{p, w}(\mathbb{R}^N) \times V _n $ we can show using the statements in Remark \ref{norms for pw case} that 
\begin{multline*}
\left\|\mathcal{F} (\mathbf{x}^1, \mathbf{z}^1)-\mathcal{F} (\mathbf{x}^2, \mathbf{z}^2)\right\|_{p,w}^p
=\sum_{t \in \mathbb{Z}_{-}} \left\|F _t(\mathbf{x}^1, {\bf z} ^1)-F _t(\mathbf{x}^2, {\bf z} ^2)\right\|^p w _{-t}\\
=\sum_{t \in \mathbb{Z}_{-}} \left\|F (\mathbf{x}^1_{t-1}, {\bf z} ^1_t)-F _t(\mathbf{x}^2_{t-1}, {\bf z} ^2_t)\right\|^p w _{-t}
\leq c _F^p\sum_{t \in \mathbb{Z}_{-}}  \left\|\mathbf{x}^1_{t-1}-\mathbf{x}^2_{t-1}\right\|^p w _{-t}+ c _F^p\sum_{t \in \mathbb{Z}_{-}}  \left\|\mathbf{z}^1_{t}-\mathbf{z}^2_{t}\right\|^p w _{-t}\\
\leq c_F^p \left\|T _1(\mathbf{x}^1- \mathbf{x}^2)\right\|^p_{p,w}+ c _F ^p \left\|{\bf z} ^1- {\bf z} ^2\right\|_{p,w}^p
\leq c_F^p   L _{w,p}^p\left\|\mathbf{x}^1- \mathbf{x}^2\right\|^p_{p,w}+ c _F ^p \left\|{\bf z} ^1- {\bf z} ^2\right\|_{p,w}^p\\
\leq\ c_F^p  L _{w,p}^p \left\|(\mathbf{x}^1,\mathbf{z}^1)-( \mathbf{x}^2,\mathbf{z}^2)\right\|_{p, w\oplus w}^p,
\end{multline*}
where in the last inequality we used that $L _{w,p}> 1 $. We now show that $\mathcal{F} $ is a contraction on the first entry whenever condition \eqref{fmp condition for pw spaces} is satisfied. Indeed,
\begin{multline*}
\left\|\mathcal{F} (\mathbf{x}^1, \mathbf{z}^1)-\mathcal{F} (\mathbf{x}^2, \mathbf{z}^2)\right\|_{p,w}^p
=\sum_{t \in \mathbb{Z}_{-}} \left\|F (\mathbf{x}^1_{t-1}, {\bf z} ^1_t)-F _t(\mathbf{x}^2_{t-1}, {\bf z} ^2_t)\right\|^p w _{-t}\\
\leq c^p\sum_{t \in \mathbb{Z}_{-}}  \left\|\mathbf{x}^1_{t-1}-\mathbf{x}^2_{t-1}\right\|^p w _{-t}=
c^p \left\|T _1(\mathbf{x}^1- \mathbf{x}^2)\right\|^p_{p,w}
\leq\ c^p  L _{w,p}^p \left\|\mathbf{x}^1-\mathbf{x}^2\right\|_{p, w}^p.
\end{multline*}
The rest of the proof can be obtained by mimicking the developments after \eqref{contraction part 2}.
\end{remark}

As a corollary of Theorem \ref{characterization of fmp unbounded} it can be shown that reservoir systems that have by construction uniformly bounded inputs and outputs always have the ESP and FMP properties and that for any weighting sequence $w$. This result was already shown in \cite[Theorem 3.1]{RC7}.

\begin{corollary}
\label{fmp when you restrict to bounded}
Let $M,L>0 $, let $K _M \subset  \left({\Bbb R}^n\right)^{\mathbb{Z}_{-}}$ and $K _L \subset  \left({\Bbb R}^N\right)^{\mathbb{Z}_{-}} $ be subsets of uniformly bounded sequences defined as in \eqref{Kset}, and let $F: \overline{B_{\left\|\cdot \right\|}({\bf 0}, L)} \times \overline{B_{\left\|\cdot \right\|}({\bf 0}, M)} \longrightarrow \overline{B_{\left\|\cdot \right\|}({\bf 0}, L)} $ be a continuous reservoir map. Assume, additionally, that $F$ is a contraction on the first entry with constant $0<c<1 $. Then, the reservoir system associated to $F$ has the echo state property. Moreover, this system has a unique associated causal and time-invariant filter $U^F:K _M \longrightarrow K _L $ that has the fading memory property with respect to any weighting sequence $w$.
\end{corollary}

\noindent\textbf{Proof.\ \ } Given that $\overline{B_{\left\|\cdot \right\|}({\bf 0}, L)}$ is a compact subset of $\mathbb{R}^N$, the hypothesis in part {\bf (i)} of Theorem \ref{characterization of fmp unbounded} and condition \eqref{fmp condition on w} guarantee that there exists a reservoir filter $U^F:K _M \longrightarrow K _L $ associated to $F$ that has the fading memory property with respect to any weighting sequence that satisfies \eqref{fmp condition on w}. Such a sequence always exists as it suffices to take any geometric sequence    $w  _t:= \lambda ^t $, $t \in \mathbb{N} $, with $c< \lambda <1 $. However, as it has been shown in \cite[Corollary 2.7]{RC7}, all the weighted norms induce in the sets $K _M  $ and $K _L $ the same topology, namely, the product topology and hence if $U^F $ is continuous with respect to the topology  induced by the weighted norm $\left\|\cdot \right\|_{w  }  $ then so it is with respect to the norm associated to any other weighting sequence. \quad $\blacksquare$

\begin{remark}
\label{fmp sufficient but not necessary}
\normalfont
This corollary shows that, in general, the condition \eqref{fmp condition on w} is sufficient but not necessary. Indeed, if the hypotheses in the corollary are satisfied, the resulting filter $U^F $ has the fading memory property with respect to any geometric sequence    $w  _t:= \lambda ^t $, with $0<\lambda<1 $, $t \in \mathbb{N} $ for which (see Remark \ref{examples for lw}) $L _{w }=1/ \lambda $.  In particular, this holds true when $\lambda  $ is chosen so that $0< \lambda< c  $ and hence when \eqref{fmp condition on w} is not satisfied since in that case $c L _{w}>1 $. Additional concrete examples that show that  the condition \eqref{fmp condition on w} is sufficient but not necessary  are provided in Section \ref{examples fmp}.
\end{remark}

\begin{remark}
\normalfont
The FMP condition \eqref{fmp condition on w} is sufficient but not necessary even in the absence of boundedness conditions like in Corollary \ref{fmp when you restrict to bounded}
\end{remark}

Another important statement that can be proved when the target of the reservoir map is a compact subset of $\mathbb{R}^N  $ is that the echo state property is in that situation guaranteed {\it for no matter what input\footnote{We thank Lukas Gonon for pointing this out.} in} $\left({\Bbb R}^n\right)^{\mathbb{Z}_{-}}$ even though the FMP may obviously not hold in that case.

\begin{theorem}[ESP for reservoir maps with compact target]
\label{ESP for reservoir maps with compact target}
Let $F:  D _N \times  D _n \longrightarrow  D_N$  be a continuous reservoir map, $D _n \subset {\Bbb R}^n $, $D _N\subset {\Bbb R}^N $, $n, N  \in \mathbb{N}^+$, such that  $D_N $ is a compact subset of ${\Bbb R}^N  $ and $F$ is a contraction on the first entry with constant $0<c<1 $. Then, the reservoir system associated to $F$ has the echo state property for any input in $(D _n)^{\mathbb{Z}_{-}} $. Let $U ^F: (D _n)^{\mathbb{Z}_{-}}  \longrightarrow (D _N)^{\mathbb{Z}_{-}}  $ be the associated reservoir filter. For any weighting sequence $w$ such that  $c L _w<1 $ the map $U ^F: (D _n)^{\mathbb{Z}_{-}}  \longrightarrow ((D _N)^{\mathbb{Z}_{-}}, \left\|\cdot \right\| _w) $ is continuous when in $(D _n)^{\mathbb{Z}_{-}}$ we consider the relative topology induced by the product topology in $({\Bbb R}^N)^{\mathbb{Z}_{-}} $. Moreover, if $(D _n)^{\mathbb{Z}_{-}} \subset \ell_{-}^{w}(\mathbb{R}^n) $ then $U ^F $ has the fading memory property.
\end{theorem}

\noindent\textbf{Proof.\ \ } Consider the map $\mathcal{F}:(D _N)^{\mathbb{Z}_{-}}\times (D _n)^{\mathbb{Z}_{-}} \longrightarrow (D _N)^{\mathbb{Z}_{-}} $ defined in \eqref{script f as product definition} and endow $(D _n)^{\mathbb{Z}_{-}} $ and $(D _N)^{\mathbb{Z}_{-}} $ with the relative topologies induced by the product topologies in $({\Bbb R}^n)^{\mathbb{Z}_{-}} $ and $({\Bbb R}^N)^{\mathbb{Z}_{-}} $, respectively. It is easy to see that the maps $p _t $  and $T _1 $ are continuous with respect to those product topologies and hence $\mathcal{F} $ can be written using \eqref{script f as product} as a Cartesian product of continuous functions, which is always continuous in the product topology. 

Consider now any weighting sequence $w$ such that  $c L _w<1 $. Using an argument similar to the proof of Lemma \ref{cr using product maps} {\bf (i)}, we can conclude that $(D_N)^{\mathbb{Z}_{-}} \subset \ell_{-}^{w}(\mathbb{R}^N) $ and that the product topology on $(D_N)^{\mathbb{Z}_{-}} $ coincides with the norm topology induced by $\left\|\cdot \right\|_w $. Now, following the expressions \eqref{contraction part 1} and \eqref{contraction part 2} it can be shown that $\mathcal{F}  $ is a contraction on the first entry and with respect to $\left\|\cdot \right\|_w $. In view of these facts and given that the product topology in $(D _n)^{\mathbb{Z}_{-}} \subset ({\Bbb R}^n)^{\mathbb{Z}_{-}} $ is metrizable (see \cite[Theorem 20.5]{Munkres:topology}) and that $(D _N)^{\mathbb{Z}_{-}} \subset ({\Bbb R}^N)^{\mathbb{Z}_{-}} $ is compact by  Tychonoff's Theorem (see \cite[Theorem 37.3]{Munkres:topology}) in the product topology and hence complete, Theorem 6.4.1 in \cite{Sternberg:dynamical:book} implies the existence of a unique fixed point of $\mathcal{F} $ for each ${\bf z} \in (D _n)^{\mathbb{Z}_{-}} $, which establishes the ESP. Moreover, that result also shows the continuity of the associated filter $U ^F: (D _n)^{\mathbb{Z}_{-}}  \longrightarrow ((D _N)^{\mathbb{Z}_{-}}, \left\|\cdot \right\| _w) $.

Finally, if $(D _n)^{\mathbb{Z}_{-}} \subset \ell_{-}^{w}(\mathbb{R}^n) $, we know from \cite[Proposition 2.9]{RC7} that the inclusion $\ell_{-}^{w}(\mathbb{R}^n)\hookrightarrow ({\Bbb R}^n)^{\mathbb{Z}_{-}} $ is continuous and hence so is $U ^F  $ when in $(D _n)^{\mathbb{Z}_{-}}  $ we consider the topology generated by the norm $\left\|\cdot \right\| _w $, which establishes the FMP in that situation. \quad $\blacksquare$

\medskip

The following result shows how the FMP of the filter associated to a reservoir map established in Theorem \ref{characterization of fmp unbounded} propagates to the FMP of the filter of the full reservoir system in the readout map is continuous.

\begin{corollary}
\label{fmp full system}
In the conditions of Theorem \ref{characterization of fmp unbounded}, let $h: D_N \longrightarrow \mathbb{R}^d $ be a continuous readout map. Consider the following two cases that correspond to the two sets of hypotheses studied in Theorem \ref{characterization of fmp unbounded}:
\begin{description}
\item [(i)] If $D_N  $ is a compact subset of ${\Bbb R}^N $ then there is a constant $R > 0 $  such that the filter $U ^F_h $ defined by $U ^F_h ({\bf z})_t:=h \left(U^F ({\bf z})_t \right)$, $t \in \mathbb{Z}_{-} $, ${\bf z} \in V _n $ maps $U ^F_h: (V _n, \left\|\cdot \right\|_w) \longrightarrow (K _R, \left\|\cdot \right\|_w)$ and has the fading memory property.
\item [(ii)]  If $(D_N)^{\mathbb{Z}_{-}}\cap \ell_{-}^{w}(\mathbb{R}^N)  $ is a complete subset of $ \left( \ell_{-}^{w}(\mathbb{R}^N, \left\|\cdot \right\|_w\right) $ and $h$ is Lipschitz continuous on $D _N $ such that $U ^F_h({\bf z}^0) \in \ell_{-}^{w}(\mathbb{R}^d) $, then the reservoir filter $U ^F_h: (V _n, \left\|\cdot \right\|_w) \longrightarrow (\ell_{-}^{w}(\mathbb{R}^d), \left\|\cdot \right\|_w)$ has the fading memory property.
\end{description}
This statement also holds true under the hypotheses in part {\rm {\bf (ii)}} when $(\ell_{-} ^{w}(\mathbb{R}^n), \left\|\cdot \right\|_w) $ is replaced by $(\ell_{-} ^{\infty}(\mathbb{R}^n), \left\|\cdot \right\|_\infty) $. The resulting reservoir filter 
$
U ^F_h:(V _n, \left\|\cdot \right\|_\infty) \longrightarrow  (\ell_{-}^{\infty}(\mathbb{R}^d), \left\|\cdot \right\|_\infty)
$ is continuous.
\end{corollary}

\subsection{Examples}
\label{examples fmp}

In the following paragraphs we show how the sufficient condition \eqref{fmp condition on w} explicitly looks like for reservoir systems that are widely used and that have been shown to have universality properties in the fading memory category both with deterministic  and stochastic inputs \cite{RC6, RC7, RC8}.

\paragraph{Linear reservoir maps.}
Consider the reservoir map $F: \mathbb{R}^N \times {\Bbb R}^n \longrightarrow {\Bbb R}^N $ given by 
\begin{equation}
\label{linear reservoir map}
F(\mathbf{x}, {\bf z})=A \mathbf{x}+ {\bf c} {\bf z}, \quad \mbox{with} \quad A \in \mathbb{M}_N, {\bf c} \in \mathbb{M}_{N,n}. 
\end{equation}  
It is easy to see that $F$ is a contraction on the first entry whenever the matrix $A$ satisfies that $\vertiii{A}<1$. In that case, using the notation in Theorem \ref{characterization of fmp unbounded}, $c=\vertiii{A} $. Indeed, for any $\mathbf{x}_1, \mathbf{x}_2 \in \mathbb{R}^N $, ${\bf z} \in {\Bbb R}^n $:
\begin{equation*}
\left\|F(\mathbf{x}_1, {\bf z})-F(\mathbf{x}_2, {\bf z})\right\|= \left\|A(\mathbf{x} _1- \mathbf{x} _2)\right\|\leq \vertiii{A} \left\|\mathbf{x} _1- \mathbf{x}_2\right\|.
\end{equation*} 
We now assume that $\vertiii{A}<1$ and prove the following two statements:

\medskip

\noindent {\bf (i)} The reservoir system associated to \eqref{linear reservoir map} has the echo state property and defines a unique reservoir filter $U^{F}: \ell_{-}^{w}(\mathbb{R}^n) \longrightarrow \ell_{-}^{w}(\mathbb{R}^N)$ that has the fading memory property with respect to any weighting sequence $w$ that satisfies the condition
\begin{equation}
\label{fmp condition for linear}
\sum_{j=0}^{\infty}\frac{\vertiii{A^j} }{w _j}< + \infty.
\end{equation}
The FMP condition \eqref{fmp condition on w} reads in this case as 
\begin{equation}
\label{sufficient condition for fmp in linear case}
\vertiii{A}L _w<1,
\end{equation}
and implies \eqref{fmp condition for linear} but not vice versa.

\medskip

\noindent {\bf (ii)} If the inputs presented to the reservoir system associated to \eqref{linear reservoir map} are uniformly bounded then it has the fading memory property with respect to any weighting sequence. This result was already known as it can be easily obtained by combining \cite[Corollary 11]{RC6} with \cite[Corollary 2.7]{RC7}. We obtain it here directly out of Corollary \ref{fmp when you restrict to bounded} by noting that for any $M>0 $,
\begin{equation}
\label{bounded condition for linear lin}
F(\overline{B_{\left\|\cdot \right\|}({\bf 0}, L)} , \overline{B_{\left\|\cdot \right\|}({\bf 0}, M)} \subset \overline{B_{\left\|\cdot \right\|}({\bf 0}, L)}, \quad \mbox{with} \quad L:= \frac{\vertiii{{\bf c}}M}{1-\vertiii{A}}.
\end{equation}

\medskip

\noindent {\bf Proof of statement (i)} One can  show by mimicking the proof of \cite[Corollary 11]{RC6} that whenever condition \eqref{fmp condition for linear} is satisfied for a given weighting sequence $w$, the reservoir system determined by \eqref{linear reservoir map} has a unique reservoir filter $U^{F}: \ell_{-}^{w}(\mathbb{R}^n) \longrightarrow \ell_{-}^{w}(\mathbb{R}^N)$ associated that is determined by the linear functional $H^{F}: \ell_{-}^{w}(\mathbb{R}^n) \longrightarrow \mathbb{R}^N  $ given by
\begin{equation*}
H^{F}({\bf z}):=\sum_{j=0}^{\infty}A ^j{\bf c} {\bf z}_{-j}.
\end{equation*}
This linear functional is bounded because for any ${\bf z}\in \ell_{-}^{w}(\mathbb{R}^n) $, the hypothesis \eqref{fmp condition for linear} implies that:
\begin{equation*}
\left\|H^{F}({\bf z})\right\|\leq \sum_{j=0}^{\infty}\vertiii{A^j} \vertiii{{\bf c}} \left\|{\bf z}_{-j}\right\|=\vertiii{{\bf c}}\sum_{j=0}^{\infty}\vertiii{A^j}  \left\|{\bf z}_{-j}\right\|\frac{w _j}{w _j}\leq \vertiii{{\bf c}} \left\|{\bf z}\right\|_w\sum_{j=0}^{\infty}\frac{\vertiii{A^j} }{w _j}<+ \infty
\end{equation*}
We now show that for any weighting sequence $w$ that satisfies $\vertiii{A}L _w<1$, the condition \eqref{fmp condition for linear} always holds. Indeed, using \eqref{convergence and divergence rates} we obtain
\begin{equation*}
\sum_{j=0}^{\infty}\frac{\vertiii{A^j} }{w _j}\leq\sum_{j=0}^{\infty}\frac{\vertiii{A} ^j}{w _j}\leq \sum_{j=0}^{\infty}\vertiii{A} ^j L _w ^j= \frac{1}{1-\vertiii{A}L _w}<+ \infty, \quad \mbox{as required.}
\end{equation*}
We finally show that there exist sequences $w$ that satisfy \eqref{fmp condition for linear} but not $\vertiii{A}L _w<1$, which is one more example of the fact, that we already indicated in Remark \ref{fmp sufficient but not necessary}, that the FMP condition \eqref{fmp condition on w} is sufficient but not necessary. Let $w $ be a harmonic weighting sequence as in Remark \ref{examples for lw} given by $w _j:= 1/(1+jd) $, $j \in \mathbb{N} $, with $d>0$. In this case $L _w= 1+d $ so we can choose a value $d$ such that  $\vertiii{A}(1+d)>1 $. However, at the same time, the condition \eqref{fmp condition for linear} holds in this case because
\begin{multline*}
\sum_{j=0}^{\infty}\frac{\vertiii{A^j} }{w _j}\leq \sum_{j=0}^{\infty} \vertiii{A} ^j (1+jd)=
\sum_{j=0}^{\infty} \vertiii{A} ^j+ d(j+1) \vertiii{A} ^j- d\vertiii{A} ^j\\
=\sum_{j=0}^{\infty} (1-d)\vertiii{A} ^j+ d(j+1) \vertiii{A} ^j= \frac{1-d}{1-\vertiii{A}}+ \frac{d}{(1-\vertiii{A})^2}= \frac{1+\vertiii{A}(d-1)}{(1-\vertiii{A}) ^2}
< + \infty.
\end{multline*}

Another example in this direction can be obtained by using a nilpotent matrices. If $A$ is nilpotent then \eqref{fmp condition for linear} is always satisfied for any weighting sequence $w$. At the same time, there are nilpotent matrices with arbitrarily large norm $\vertiii{A} $ which, once more, shows that \eqref{fmp condition for linear} can hold, and hence the FMP, without \eqref{sufficient condition for fmp in linear case} being necessarily true. We notice too that reservoir systems determined by nilpotent matrices always satisfy the echo state property even though they are not necessarily contractions.

\medskip

\noindent {\bf Proof of statement (ii)} We first prove the statement  \eqref{bounded condition for linear lin}. For any $\mathbf{x}\in \overline{B_{\left\|\cdot \right\|}({\bf 0}, L)} $ and ${\bf z}\in \overline{B_{\left\|\cdot \right\|}({\bf 0}, M) }$,
\begin{equation*}
\left\|F(\mathbf{x}, {\bf z})\right\|= \left\|A \mathbf{x}+ {\bf c} {\bf z}\right\|\leq \vertiii{A}L+\vertiii{{\bf c}}M=L, \quad \mbox{as required.}
\end{equation*}
This implies that the reservoir map $F$ in \eqref{linear reservoir map} restricts to a map $F_{L,M}:\overline{B_{\left\|\cdot \right\|}({\bf 0}, L)} \times  \overline{B_{\left\|\cdot \right\|}({\bf 0}, M)} \longrightarrow \overline{B_{\left\|\cdot \right\|}({\bf 0}, L)} $ that is a contraction on the first entry with constant $\vertiii{A}<1 $ and hence satisfies the hypotheses of Corollary \ref{fmp when you restrict to bounded}. This guarantees the existence of a unique associated causal and time-invariant filter $U^F:K _M \longrightarrow K _L $ that has the fading memory property with respect to any weighting sequence $w$.

\paragraph{Echo state networks (ESN).} Let $\sigma: \mathbb{R} \longrightarrow [-1,1] $ be a squashing function, that is, $\sigma $ is non-decreasing, $\lim_{x \rightarrow -\infty} \sigma(x)=-1 $, and $\lim_{x \rightarrow \infty} \sigma(x)=1 $. Moreover, assume that $L _\sigma:=\sup_{x \in \mathbb{R}}\{| \sigma' (x)|\}  < +\infty$. Let $\boldsymbol{\sigma}: \mathbb{R}^N \longrightarrow [-1,1]^N $ be the map obtained by componentwise application of the the squashing function $\sigma $. An echo state network is a reservoir system with linear readout and reservoir map given by 
\begin{equation}
\label{esn reservoir map}
F(\mathbf{x}, {\bf z})=\boldsymbol{\sigma}(A \mathbf{x}+ {\bf c} {\bf z}+ \boldsymbol{\zeta}), \quad \mbox{with} \quad A \in \mathbb{M}_N, {\bf c} \in \mathbb{M}_{N,n}, \boldsymbol{\zeta} \in \mathbb{R}^N. 
\end{equation}  
We notice first that if $\vertiii{A}L _\sigma< 1  $ then $F$ is a contraction on the first component with constant $\vertiii{A}L _\sigma $ (see the second part in \cite[Corollary 3.2]{RC7}). By construction, $F$ maps into the compact space $[-1,1]^N \subset \mathbb{R}^N$ and hence satisfies the hypotheses in the first part of Theorem \ref{characterization of fmp unbounded}. Consequently, for any weighting sequence $w$ that satisfies 
\begin{equation}
\label{condition esn for fmp rc9}
\vertiii{A}L _\sigma L _w< 1
\end{equation}
there exists a unique reservoir filter $U^{F}: \ell_{-}^{w}(\mathbb{R}^n) \longrightarrow\ell_{-}^{w}(\mathbb{R}^N)$ associated to $F$ 
that has the fading memory property with respect to   $w$. By Corollary \ref{fmp when you restrict to bounded} this statement holds true for any $w$ when one considers uniformly bounded inputs.

\paragraph{Non-homogeneous state-affine systems (SAS).} These systems are determined by reservoir maps $F: \mathbb{R}^N \times {\Bbb R}^n \longrightarrow\mathbb{R}^N $ of the form
\begin{equation}
\label{reservoir map SAS general}
F(\mathbf{x}, {\bf z}):=p({\bf z})\mathbf{x}+q({\bf z}),
\end{equation}
where $p $ and $q $ are polynomials with matrix and vector coefficients, respectively, that depending on their nature determine the following two families of SAS systems:
\begin{description}
\item [(i)] {\bf Regular SAS.} $p $ and $q$ are polynomials of degree $r$ and $s$ of the form:
\begin{eqnarray*}
p({\bf z})&=&\sum_{{i _1, \ldots, i _n \in \left\{0, \ldots, r\right\} \above 0 pt i _1+ \cdots + i _n\leq r}}
z _1^{i _1} \cdots z _n^{i _n} A_{{i _1, \ldots, i _n}}, \quad A_{{i _1, \ldots, i _n}} \in \mathbb{M}_{N}, \quad {\bf z} \in D_n\subset{\Bbb R}^n,\\
q({\bf z})&=&\sum_{{i _1, \ldots, i _n \in \left\{0, \ldots, s\right\} \above 0 pt i _1+ \cdots + i _n\leq s}}
z _1^{i _1} \cdots z _n^{i _n} B_{{i _1, \ldots, i _n}}, \quad B_{{i _1, \ldots, i _n}} \in \mathbb{M}_{N,1}, \quad {\bf z} \in D_n\subset{\Bbb R}^n.
\end{eqnarray*}
\item [(ii)] {\bf Trigonometric SAS.} We use trigonometric polynomials instead:
\begin{eqnarray*}
p({\bf z})&=&\sum_{k=1}^r A_k^p \cos(\mathbf{u}_k^p \cdot {\bf z}) + B_k^p \sin(\mathbf{v}_k^p \cdot {\bf z}), \quad A_{k}^p, B _k^p \in \mathbb{M}_{N}, \quad \mathbf{u}_k^p, \mathbf{v} _k^p \in {\Bbb R}^N,\quad {\bf z} \in D_n\subset{\Bbb R}^n,\\
q({\bf z})&=&\sum_{k=1}^s A_k^q \cos(\mathbf{u}_k^q \cdot {\bf z}) + B_k^q \sin(\mathbf{v}_k^q \cdot {\bf z}), \quad A_{k}^q, B _k^q \in \mathbb{M}_{N,1}, \quad \mathbf{u}_k^q, \mathbf{v} _k^q \in {\Bbb R}^N,\quad {\bf z} \in D_n\subset {\Bbb R}^n.
\end{eqnarray*}
\end{description} 
In both cases, define
\begin{equation*}
M _p:=\sup_{{\bf z} \in D_n}\left\{\vertiii{p({\bf z})}\right\} \quad \mbox{and} \quad M _q:=\sup_{{\bf z} \in D_n}\left\{\vertiii{q({\bf z})}\right\}.
\end{equation*}
Note that for regular SAS defined by nontrivial polynomials, the set $D_n$ needs to be bounded in order for $M _p$ and $M _q $ to be finite. Additionally, it is easy to see that $F$ is a contraction on the first  entry with constant $M _p$ whenever $M _p<1 $, which is a condition that we will assume holds true in the rest of this example. Additionally, we  assume that $M _q<+ \infty $. Regular SAS are a generalization of the linear case that we considered in the first part of this section and hence  two statements can be proved that are analogous to the ones in that part, namely:

\noindent {\bf (i)} The reservoir system associated to \eqref{reservoir map SAS general} has the echo state property and defines a unique reservoir filter $U^{F}: \ell_{-}^{w}(\mathbb{R}^n) \longrightarrow \ell_{-}^{w}(\mathbb{R}^N)$ that has the fading memory property with respect to any weighting sequence $w$ that satisfies the condition
\begin{equation}
\label{fmp condition for sas}
\sum_{j=0}^{\infty}\frac{M _p ^j}{w _j}< + \infty.
\end{equation}
The FMP condition \eqref{fmp condition on w} that in this case reads as $M _pL _w<1$ implies \eqref{fmp condition for sas} but not vice versa.

\medskip

\noindent {\bf (ii)} If the inputs presented to the reservoir system associated to \eqref{reservoir map SAS general} are uniformly bounded then it has the fading memory property with respect to any weighting sequence. We obtain this result out of Corollary \ref{fmp when you restrict to bounded} by noting that for any $M>0 $,
\begin{equation*}
F(\overline{B_{\left\|\cdot \right\|}({\bf 0}, L)} , \overline{B_{\left\|\cdot \right\|}({\bf 0}, M)} \subset \overline{B_{\left\|\cdot \right\|}({\bf 0}, L)}, \quad \mbox{with} \quad L:= \frac{M _qM}{1-M _p}.
\end{equation*}
We emphasize that in the case of regular SAS, this is the only situation for which one can have $M _p<1 $ and $M _q<+ \infty $.

\medskip

\noindent We prove only the statement {\bf (i)} since statement {\bf (ii)} can be easily obtained by mimicking the similar statement for the linear case. Indeed, a straightforward generalization of \cite[Proposition 14]{RC6} shows that whenever $M _p<1 $ and $M _q<+ \infty $, the reservoir system determined by \eqref{reservoir map SAS general} has a unique reservoir filter $U^{F}: \ell_{-}^{w}(\mathbb{R}^n) \longrightarrow \ell_{-}^{w}(\mathbb{R}^N)$ associated that is determined by the linear functional $H^{F}: \ell_{-}^{w}(\mathbb{R}^n) \longrightarrow \mathbb{R}^N  $ given by
\begin{equation*}
H^{F}({\bf z}):=\sum_{j=0}^{\infty} \left(\prod_{k=0}^{j-1}p({\bf z}_{-k})\right)q({\bf z}_{-j}).
\end{equation*}
Mimicking the proof of \cite[Proposition 16]{RC6} it can be shown that there exists a constant $C_{p,q} >0$ that depends exclusively of $p $  and $q $ such that  for any ${\bf z}, {\bf s} \in \ell_{-}^{w}(\mathbb{R}^n) $
\begin{equation*}
\left\|H^{F}({\bf z})- H^{F}({\bf s})\right\|\leq C_{p,q}\sum_{j=0}^ \infty M _p^j \left\|{\bf z} _{-j}-{\bf s} _{-j}\right\|=C_{p,q}\sum_{j=0}^ \infty M _p^j \left\|{\bf z} _{-j}-{\bf s} _{-j}\right\| \frac{w _j}{w _j}\leq 
C_{p,q}\left\|{\bf z}- {\bf s}\right\|_w\sum_{j=0}^ \infty \frac{M _p^j}{w _j},
\end{equation*}
which shows that $H^{F}: \ell_{-}^{w}(\mathbb{R}^n) \longrightarrow \mathbb{R}^N  $ is Lipschitz continuous whenever the condition \eqref{fmp condition for sas} holds. The last claim regarding the relation between \eqref{fmp condition for sas} and the FMP condition \eqref{fmp condition on w} is proved by mimicking the similar statement for the linear case.

\section{Differentiability in reservoir filters with unbounded inputs}
\label{Differentiability in reservoir filters}
 
We now extend the  results in the previous section from continuity to differentiability. More specifically, we characterize the situations in which one can prove the existence and obtain the differentiability of reservoir filters out of the  differentiability properties of the maps that define the reservoir system. This approach gives us in passing new techniques to establish the echo state and the fading memory properties of reservoir systems. In particular, differentiability being a local property, we show how systems that do not globally have any of these properties may still have them in a neighborhood of certain types of inputs. A phenomenon of this type has also been explored in \cite{Manjunath:Jaeger}.

It is worth emphasizing that the study of the differentiability properties of fading memory reservoir filters calls naturally for the handling of unbounded inputs since the definition of the Fr\'echet derivative requires them to be defined on open subsets of the Banach space $\ell_{-}^{w}(\mathbb{R}^n) $ that always contain unbounded sequences, as we saw in the first part of Lemma \ref{topological lemma balls etc}.

\subsection{Differentiable reservoir filters associated to differentiable reservoir maps}
\label{Differentiable reservoir filters associated to differentiable reservoir maps}

The first result in this section shows that under certain conditions, the echo state and the fading memory properties associated to differentiable reservoir systems locally persist, that is, if a reservoir system has a unique filter associated to a specific input and it is continuous and differentiable at it, then the same property holds for neighboring inputs.

 \begin{theorem}[Local persistence of the ESP and FMP properties]
\label{Persistence of the ESP and FMP properties}
Let $F:  \mathbb{R}^N \times  \mathbb{R}^n \longrightarrow  \mathbb{R}^N$  be a reservoir map and let $w$ be a weighting sequence with finite inverse decay ratio $L _w$. Suppose that $F$ is of class $C ^1(\mathbb{R}^N \times  \mathbb{R}^n)$ and that the corresponding reservoir system \eqref{reservoir equation} has a solution $(\mathbf{x}^0, {\bf z}^0) \in  \ell_{-}^{w}(\mathbb{R}^N) \times  \ell_{-}^{w}(\mathbb{R}^n)$, that is, $\mathbf{x}_t^0=F(\mathbf{x}_{t-1}^0, {\bf z}_t ^0)$, for all $t \in \mathbb{Z}_{-} $.  Suppose, additionally, that 
\begin{equation}
\label{differentiability condition local}
L_{F}:=\sup_{(\mathbf{x}, {\bf z}) \in \mathbb{R}^N \times  \mathbb{R}^n}\left\{\vertiii{D F (\mathbf{x}, {\bf z})}\right\}<+ \infty. 
\end{equation}
Define $$L_{F _x}(\mathbf{x}^0, {\bf z}^0):=\sup_{t \in \mathbb{Z}_{-}}
\left\{\vertiii{D_xF ({\bf x}^0 _{t-1}, {\bf z}^0_t )}\right\}$$ 
and suppose that 
\begin{equation}
\label{persistence condition}
L_{F _x}(\mathbf{x}^0, {\bf z}^0) L _w<1.
\end{equation}
Then there exist  open time-invariant neighborhoods $V_{\mathbf{x}^0} $ and $V_{\mathbf{z}^0} $ of $\mathbf{x} ^0 $ and $\mathbf{z} ^0 $  in $ \ell_{-}^{w}(\mathbb{R}^N)$ and $ \ell_{-}^{w}(\mathbb{R}^n)$, respectively, such that the reservoir system associated to $F$ with inputs in $V_{\mathbf{z}^0} $ has the echo state property and hence determines a unique causal and time-invariant reservoir filter 
$
U^F:(V_{\mathbf{z}^0}, \left\|\cdot \right\|_w) \longrightarrow (V_{\mathbf{x}^0}, \left\|\cdot \right\|_w)
$. Moreover, $U^F $  is differentiable at all the points of the form $T _{-t}(\mathbf{z}^0) $, $t \in \mathbb{Z}_{-} $, it is locally  Lipschitz continuous on $V_{\mathbf{z}^0} $, and it hence has the fading memory property.
\end{theorem}

\begin{remark}
\normalfont
We  refer to \eqref{persistence condition} as the {\bfi  persistence condition}. We emphasize that this inequality puts into relation the solution $(\mathbf{x}^0, {\bf z}^0) $ whose persistence we are studying with the weighting sequence $w$. In particular, that relation tells us that solutions are more likely to persist with respect to weighting sequences that decay more slowly (that is, $L _w  $ is smaller). 
\end{remark}

\begin{remark}
\normalfont
There is a situation where the persistence condition is particularly easy to verify, namely, when the solution of the reservoir system is constructed as a constant sequence coming from a fixed point of the reservoir map, that is, $(\mathbf{x}^0, {\bf z}^0) \in \mathbb{R}^N \times  \mathbb{R}^n $ such that $F(\mathbf{x}^0, {\bf z}^0)=\mathbf{x}^0 $. In that case $L_{F _x}(\mathbf{x}^0, {\bf z}^0):=
\vertiii{D_xF ({\bf x}^0  , {\bf z}^0  )}$ .
\end{remark}

\begin{remark}
\normalfont
The persistence condition \eqref{persistence condition} can be interpreted as a stability condition for the reservoir system determined by $F$ at the solution $({\bf x}^0, {\bf z}^0) $ with respect to perturbations in $\ell_{-}^{w}(\mathbb{R}^n) $. The persistence of solutions under stability conditions of that type has been thoroughly studied for many types of dynamical systems (see, for instance, \cite{persistence:periodic, mre, paper1:persitence, pascal}).
\end{remark}

\begin{remark}
\normalfont
The derivative $DU^F({\bf z}^0) $ at ${\bf z}^0$ of the locally defined reservoir filter $U^F $ is determined by the differentiation  of the relation \eqref{relation function filter U}. Indeed, for any $\mathbf{u} \in  \ell_{-} ^{w}(\mathbb{R}^n) $, and $t \in \mathbb{Z}_{-} $, the directional derivative $D U^F ({\bf z}^0)\cdot \mathbf{u} $ is determined by the recursions
\begin{eqnarray}
(D U^F ({\bf z}^0)\cdot \mathbf{u}) _t&=&DF \left(U^F({\bf z}^0)_{t-1}, {\bf z}^0 _t\right)\cdot
\left(\left(DU^F ({\bf z}^0)\cdot \mathbf{u}\right) _{t-1}, \mathbf{u} _t\right) \label{recursions U1}\\
	&=&D_xF \left(U^F({\bf z}^0)_{t-1}, {\bf z}^0 _t\right)\cdot  \left(DU^F ({\bf z}^0)\cdot \mathbf{u}\right) _{t-1} +D_{z}F(U({\bf z}^0)_{t-1}, {\bf z}^0 _t)\cdot  \mathbf{u} _t.\label{recursions U2}
\end{eqnarray}
This relation implies, in particular, that $D U^F ({\bf z}^0): \ell_{-}^{w}(\mathbb{R}^n) \longrightarrow \ell_{-}^{w}(\mathbb{R}^N) $ is a bounded linear operator and that
\begin{equation}
\label{norm duU}
\vertiii{D U^F ({\bf z}^0)}_w\leq \frac{L_{F _z}(\mathbf{x}^0, {\bf z}^0)}{1-L_{F _x}(\mathbf{x}^0, {\bf z}^0)L _w},
\end{equation}
where 
$$L_{F _z}(\mathbf{x}^0, {\bf z}^0):=\sup_{t \in \mathbb{Z}_{-}}
\left\{\vertiii{D_zF ({\bf x}^0 _{t-1}, {\bf z}^0_t )}\right\}.$$
Indeed, notice first that for any $ t \in \mathbb{Z}_{-} $,
\begin{equation}
\label{ineqs for lfs derivatives}
\vertiii{D_xF ({\bf x}^0 _{t-1}, {\bf z}^0_t )}\leq \vertiii{DF ({\bf x}^0 _{t-1}, {\bf z}^0_t )}\quad \mbox{and} \quad \vertiii{D_zF ({\bf x}^0 _{t-1}, {\bf z}^0_t )}\leq \vertiii{DF ({\bf x}^0 _{t-1}, {\bf z}^0_t )},
\end{equation}
which, using hypothesis \eqref{differentiability condition local} implies that
\begin{equation}
\label{ineqs for lfs}
L_{F _x}(\mathbf{x}^0, {\bf z}^0)\leq L_{F }<+ \infty\quad \mbox{and } \quad
L_{F _z}(\mathbf{x}^0, {\bf z}^0)\leq L_{F }<+ \infty.
\end{equation}
Now, for any $\mathbf{u} \in  \ell_{-} ^{w}(\mathbb{R}^n) $, and $t \in \mathbb{Z}_{-} $, the relation \eqref{recursions U1} and the inequalities \eqref{ineqs for lfs} imply that
\begin{align*}
\left\|D U^F ({\bf z}^0) \cdot \mathbf{u}\right\|_w&= \sup_{t \in \mathbb{Z}_{-}} \left\{\left\|\left(D U^F ({\bf z}^0) \cdot \mathbf{u}\right) _t\right\|w _{-t}\right\}\\
&= \sup_{t \in \mathbb{Z}_{-}} \left\{\left\|DF \left(U^F({\bf z}^0)_{t-1}, {\bf z}^0 _t\right)\cdot
\left(\left(DU^F ({\bf z}^0)\cdot \mathbf{u}\right) _{t-1}, \mathbf{u} _t\right)\right\|w _{-t}\right\}\\
&= \sup_{t \in \mathbb{Z}_{-}} \left\{\left\|D_xF \left(U^F({\bf z}^0)_{t-1}, {\bf z}^0 _t\right)\cdot  \left(DU^F ({\bf z}^0)\cdot \mathbf{u}\right) _{t-1} +D_{z}F(U({\bf z}^0)_{t-1}, {\bf z}^0 _t)\cdot  \mathbf{u} _t\right\|w _{-t}\right\}\\
&\leq L_{F _x}(\mathbf{x}^0, {\bf z}^0)\sup_{t \in \mathbb{Z}_{-}} \left\{\left\|
\left(DU^F ({\bf z}^0)\cdot \mathbf{u}\right) _{t-1}\right\|w _{-t}\right\}+ L_{F _z}(\mathbf{x}^0, {\bf z}^0)\sup_{t \in \mathbb{Z}_{-}} \left\{\left\|\mathbf{u} _t\right\|w _{-t}\right\}\\
&\leq L_{F _x}(\mathbf{x}^0, {\bf z}^0)\sup_{t \in \mathbb{Z}_{-}} \left\{\left\|
\left(DU^F ({\bf z}^0)\cdot \mathbf{u}\right) _{t-1}\right\|w _{-(t-1)}\frac{w _{-t}}{w _{-(t-1)}}\right\}\\
&\enspace\enspace\enspace\enspace\enspace\enspace +L_{F _z}(\mathbf{x}^0, {\bf z}^0)\sup_{t \in \mathbb{Z}_{-}} \left\{\left\|\mathbf{u} _t\right\|w _{-t}\right\}\\
&\leq L_{F _x}(\mathbf{x}^0, {\bf z}^0)L _w\left\|D U^F ({\bf z}^0) \cdot \mathbf{u}\right\|_w+L_{F _z}(\mathbf{x}^0, {\bf z}^0) \left\|\mathbf{u}\right\|_w,
\end{align*}
which implies \eqref{norm duU}.
\end{remark}

The previous theorem proves that when the persistence condition \eqref{persistence condition} is satisfied at a preexisting solution of a reservoir system then this system has a unique fading memory (and differentiable) filter associated for neighboring inputs. In the next results we show that a global version of that condition ensures first, that globally defined reservoir filters exist, and second, that those filters are differentiable and hence have the fading memory property. 

\begin{theorem}[Characterization of global reservoir filter differentiability]
\label{characterization of reservoir differentiability}
Let  $F:  \mathbb{R}^N \times  {\Bbb R}^n \longrightarrow  \mathbb{R}^N$  be a reservoir map of class $C ^1(\mathbb{R}^N \times  \mathbb{R}^n)$ and let $w$ be a weighting sequence with finite inverse decay ratio $L _w$.
\begin{description}
\item [(i)] Suppose that $F$ satisfies \eqref{differentiability condition local} and define 
\begin{equation*}
L_{F _x}:=\sup_{(\mathbf{x}, {\bf z}) \in \mathbb{R}^N \times  {\Bbb R}^n}
\left\{\vertiii{D_xF (\mathbf{x}, {\bf z})}\right\} \quad \mbox{and } \quad L_{F _z}:=\sup_{(\mathbf{x}, {\bf z}) \in \mathbb{R}^N \times  {\Bbb R}^n}
\left\{\vertiii{D_zF (\mathbf{x}, {\bf z})}\right\}.
\end{equation*}

If the reservoir system \eqref{reservoir equation} associated to $F$ has a solution $(\mathbf{x}^0, {\bf z}^0) \in  \ell_{-}^{w}(\mathbb{R}^N)\times  \ell_{-}^{w}(\mathbb{R}^n) $, that is, $\mathbf{x}_t^0=F(\mathbf{x}_{t-1}^0, {\bf z}_t ^0)$, for all $t \in \mathbb{Z}_{-} $, and 
\begin{equation}
\label{persistence condition global}
L_{F _x} L _w<1
\end{equation}
then it has the echo state property and hence determines a unique  causal and time-invariant reservoir filter 
$
U^F:(\ell_{-}^{w}(\mathbb{R}^n), \left\|\cdot \right\|_w) \longrightarrow (\ell_{-}^{w}(\mathbb{R}^N), \left\|\cdot \right\|_w)
$. Moreover, $U^F $ is differentiable and Lipschitz continuous on  $\ell_{-}^{w}(\mathbb{R}^n)$ with Lipschitz constant $L_{ U^F} $ given by 
\begin{equation}
\label{lip constant uf}
L_{ U^F}:= \frac{L_{F _z}}{1-L_{F _x}L _w} \quad \mbox{and} \quad \vertiii{DU ^F({\bf z})}_w\leq  \frac{L_{F _z}}{1-L_{F _x}L _w}, \quad \mbox{for any ${\bf z} \in \ell_{-}^{w}(\mathbb{R}^n) $.} 
\end{equation}
The filter $U ^F $  has hence the fading memory property.
\item [(ii)] Conversely, let $V _n\subset \ell_{-}^{w}(\mathbb{R}^n) $ be an open and time-invariant subset of $\ell_{-}^{w}(\mathbb{R}^n)  $ and assume that the reservoir system \eqref{reservoir equation} associated to $F$ has a unique causal and time-invariant reservoir filter $
U^F: V _n \longrightarrow  \ell_{-}^{w}(\mathbb{R}^N)$ that is differentiable at ${\bf z}^0 \in \ell_{-}^{w}(\mathbb{R}^n)  $. Then,
\begin{equation}
\label{converse condition smooth 1}
\rho\left((\prod_{t \in \mathbb{Z}_{-}}D_xF \left(U^F({\bf z}^0 )_{t-1}, {\bf z} ^0 _t\right)) \circ T _1\right)< 1,
\end{equation}
where $\rho $ stands for the spectral radius. This in turn implies that
\begin{equation}
\label{converse condition smooth 2}
\lim\limits_{k \rightarrow +\infty} 
\left(\vertiii{D_xF \left(U^F({\bf z}^0 )_{-1}, {\bf z} ^0 _0\right)\circ \cdots \circ 
D_xF \left(U^F({\bf z}^0 )_{-k}, {\bf z} ^0 _{-k+1}\right)}\frac{1}{w_k}\right)=0.
\end{equation}
\end{description}
\end{theorem}

\begin{examples}
\label{differentiability examples}
\normalfont
We briefly examine the form that the hypotheses of Theorem \ref{characterization of reservoir differentiability} take for the three families of reservoir systems that we analyzed in Section \ref{examples fmp}:

\medskip

\noindent {\bf (i) Linear reservoir maps.} In this case, for any $\mathbf{x}\in \mathbb{R}^N   $  and ${\bf z}\in {\Bbb R}^n $, 
\begin{equation*}
DF(\mathbf{x}, {\bf z})=\left. \left(A\,\right| {\bf c}\right), \quad D_xF(\mathbf{x}, {\bf z})=A, \quad \mbox{and } \quad D_zF(\mathbf{x}, {\bf z})= {\bf c}.
\end{equation*}
Consequently $L_{F }=\vertiii{\left. \left(A\,\right| {\bf c}\right)} $, $L_{F _x}=\vertiii{A} $,  $L_{F _z}=\vertiii{{\bf c}} $. The condition \eqref{differentiability condition local} is always satisfied and in this case the sufficient  differentiability condition \eqref{persistence condition global} amounts to $\vertiii{A}L _w<1 $ that, as we saw in \eqref{sufficient condition for fmp in linear case}, is the same as the sufficient condition for the FMP to hold.

\medskip

\noindent{\bf (ii) Echo state networks (ESN).} Consider an ESN constructed using a squashing function $\sigma $ that satisfies that $L _\sigma:=\sup_{x \in \mathbb{R}}\{| \sigma' (x)|\}  < +\infty$. In this case, for any $\mathbf{x}\in \mathbb{R}^N   $  and ${\bf z}\in {\Bbb R}^n $, 
\begin{eqnarray*}
DF(\mathbf{x}, {\bf z})&=&D \boldsymbol{\sigma}(A \mathbf{x}+ {\bf c} {\bf z}+ \boldsymbol{\zeta}) \circ \left. \left(A\,\right| {\bf c}\right), \\ 
D_xF(\mathbf{x}, {\bf z})&=&D \boldsymbol{\sigma}(A \mathbf{x}+ {\bf c} {\bf z}+ \boldsymbol{\zeta}) \circ A, \\ 
D_zF(\mathbf{x}, {\bf z})&=&D \boldsymbol{\sigma}(A \mathbf{x}+ {\bf c} {\bf z}+ \boldsymbol{\zeta}) \circ {\bf c}.
\end{eqnarray*}
Notice that $\vertiii{D\boldsymbol{\sigma}(\mathbf{x})}<L_\sigma <+ \infty$, for any $\mathbf{x}\in {\Bbb R}^N $, and hence
\begin{eqnarray*}
\vertiii{DF(\mathbf{x}, {\bf z})}&\leq &L _\sigma\vertiii{\left. \left(A\,\right| {\bf c}\right)}<+ \infty, \\ 
\vertiii{D_xF(\mathbf{x}, {\bf z})}&\leq &L _\sigma\vertiii{A}<+ \infty, \\ 
\vertiii{D_zF(\mathbf{x}, {\bf z})}&\leq &L _\sigma\vertiii{{\bf c}}<+ \infty,
\end{eqnarray*}
for any $\mathbf{x}\in \mathbb{R}^N   $  and ${\bf z}\in {\Bbb R}^n $. This implies, in particular, that in this case 
\begin{equation*}
L _F<+ \infty, \quad L _{F_x}<+ \infty, \quad L _{F_z}<+ \infty,
\end{equation*}
and the sufficient differentiability condition \eqref{persistence condition global} is implied by the inequality
\begin{equation}
\label{global condition for ESN}
\vertiii{A}L _\sigma L _w<1.
\end{equation}

\medskip

\noindent {\bf (iii) Non-homogeneous state-affine systems (SAS).} A straightforward computations shows that
for any $\mathbf{x}\in \mathbb{R}^N   $  and ${\bf z}\in {\Bbb R}^n $, 
\begin{equation}
\label{derivatives for sas}
\begin{array}{rcl}
DF(\mathbf{x}, {\bf z})&=&\left(p({\bf z}), Dp({\bf z})(\cdot )\mathbf{x}+ Dq ({\bf z})(\cdot )\right), \\ 
D_xF(\mathbf{x}, {\bf z})&=&p({\bf z}), \\ 
D_zF(\mathbf{x}, {\bf z})&=&Dp({\bf z})(\cdot )\mathbf{x}+ Dq ({\bf z})(\cdot ).
\end{array}
\end{equation}
As we already pointed out, for regular SAS defined by nontrivial polynomials the norm $\vertiii{p(z)} $ is not bounded in $\mathbb{R}^n $ and hence $L _{F_x}= \sup_{(\mathbf{x}, {\bf z}) \in \mathbb{R}^N \times  {\Bbb R}^n}\left\{\vertiii{D_xF(\mathbf{x}, {\bf z})}\right\}= \sup_{{\bf z} \in {\Bbb R}^n}\left\{\vertiii{p({\bf z})}\right\}= M _p $  is not finite; the same applies to $L _F $, which implies that in this case neither \eqref{differentiability condition local} nor \eqref{persistence condition global} can be satisfied. 

This is not the case for trigonometric SAS for which the norms of the derivatives in \eqref{derivatives for sas} are bounded on their domains which, in particular, implies that $L _F<+ \infty$, $ L _{F_x}<+ \infty$, and  $ L _{F_z}<+ \infty $.  Moreover, the sufficient differentiability condition \eqref{persistence condition global} in this case reads
\begin{equation*}
M _pL _w<1.
\end{equation*}
\end{examples}

\begin{remark}
\label{everything is sufficient but not necessary}
\normalfont
We recall here an example that we introduced in Section \ref{examples fmp} to show that, as it was already the case with the FMP condition \eqref{fmp condition on w} in Theorem \ref{characterization of fmp unbounded}, the differentiability condition \eqref{persistence condition global} is sufficient but not necessary. Indeed,  consider a linear system with matrix $A$ given by
\begin{equation*}
A=\left(
\begin{array}{cc}
0 &a\\
0 & 0
\end{array}
\right), \quad \mbox{with} \quad a>0.
\end{equation*}
Given that $\vertiii{A}=a $, the reservoir map determined by $A$ is not necessarily a contraction on the first entry. Nevertheless, the nilpotency of $A$ implies that the reservoir system associated to \eqref{linear reservoir map} always has a solution for any input ${\bf z}\in ({\Bbb R}^2)^{\mathbb{Z}_{-}} $ and hence has the ESP and induces a filter $U: ({\Bbb R}^2)^{\mathbb{Z}_{-}} \longrightarrow ({\Bbb R}^2)^{\mathbb{Z}_{-}}$ given by $U({\bf z}) _t:= {\bf z} _t+ A {\bf z} _{t-1} $, $t \in \mathbb{Z}_{-} $ or, equivalently, $U= \mathbb{I}_{({\Bbb R}^2)^{\mathbb{Z}_{-}}}+(\prod_{t \in \mathbb{Z}_{-}}A) \circ  T _1 $. Consider now any weighting sequence $w$ with finite inverse decay ratio $L _w $. Then the restriction of $U$ to $\ell_{-}^{w}(\mathbb{R}^2) $ always maps into $\ell_{-}^{w}(\mathbb{R}^2) $, has the FMP, and it is differentiable. Indeed, it is easy to show using the linearity of the filter that $U=DU( {\bf z}) $ for any ${\bf z} \in \ell_{-}^{w}(\mathbb{R}^2) $ and that 
\begin{equation}
\label{diff nilpotent filter}
\vertiii{U}_w=\vertiii{DU({\bf z})}_w\leq (1+ aL _w).
\end{equation}
Note that in this case $L_{F _x}=\vertiii{A}=a $ and as \eqref{diff nilpotent filter} shows the differentiability of $U$ with respect to any weighting sequence with finite $L _w$, we can conclude that the condition \eqref{persistence condition global} is not necessary for filter differentiability.
\end{remark}

The following corollary puts together the previous theorem and a condition on the readout map that guarantees that the filter associated to the resulting reservoir system is differentiable.

\begin{corollary}
\label{differentiability of full system}
Consider a reservoir system determined by a reservoir map $F:  \mathbb{R}^N \times  {\Bbb R}^n \longrightarrow  \mathbb{R}^N$ of class $C ^1(\mathbb{R}^N \times  \mathbb{R}^n)$ and by a readout map $h: \mathbb{R} ^N\longrightarrow \mathbb{R}^d  $ that is also of class $C ^1(\mathbb{R}^N) $. Assume, additionally that $F$ satisfies the hypotheses in part {\bf (i)} of Theorem \ref{characterization of reservoir differentiability} and that $h$ is such that 
\begin{equation}
\label{readout bounded derivative}
c _h:=\sup_{\mathbf{x} \in {\Bbb R}^N}\left\{\vertiii{Dh(\mathbf{x})}\right\}<+ \infty,
\end{equation}
and the sequence ${\bf y}^0:= \left(h \left(\mathbf{x} ^0\right)\right)_{t \in \mathbb{Z}_{-}}=\left(h \left(U^F(\mathbf{z} ^0)\right)\right)_{t \in \mathbb{Z}_{-}} \in \ell_{-}^{w}(\mathbb{R}^d) $. Then, the reservoir filter $
U ^F_h:(\ell_{-}^{w}(\mathbb{R}^n), \left\|\cdot \right\|_w) \longrightarrow (\ell_{-}^{w}(\mathbb{R}^d), \left\|\cdot \right\|_w)
$ is differentiable at each point in its domain and it hence has the fading memory property.
\end{corollary}

\noindent\textbf{Proof.\ \ } Define first the map 
\begin{equation}
\label{definition funny h}
{\mathcal H}:=\prod_{t \in \mathbb{Z}_{-}}h \circ p _t: \ell_{-}^{w}(\mathbb{R}^N) \longrightarrow(\mathbb{R}^d) ^{\mathbb{Z}_{-}}.
\end{equation}
 Given that $U ^F_h= {\mathcal H} \circ U^F $ and by Theorem \ref{characterization of reservoir differentiability} the filter $U^F $ is differentiable then it suffices to prove that ${\mathcal H} $ is differentiable. This is a consequence of part {\bf (iii)} in Lemma \ref{cr using product maps} and the hypothesis \eqref{readout bounded derivative}. Indeed, let $H _t:=h \circ p _t $, $t \in \mathbb{Z}_{-} $, and notice that by the first part of Lemma \ref{smooth projections and time delays}
\begin{equation*}
\sup_{{\bf x} \in \ell_{-}^{w}(\mathbb{R}^N)} \left\{\vertiii{DH _t({\bf x})}\right\}\leq 
\sup_{{\bf x}_t \in \mathbb{R}^N} \left\{\vertiii{Dh({\bf x}_t)}\right\}\cdot \sup_{{\bf x} \in \ell_{-}^{w}(\mathbb{R}^N)} \left\{\vertiii{p _t({\bf x})}\right\}\leq  \frac{c _h}{w _{-t}}.
\end{equation*}
Now, as $\left\|\left(c _h/w _{-t}\right)_{t \in \mathbb{Z}_{-}}\right\| _w= c _h<+ \infty $ and by hypothesis ${\mathcal H}(\mathbf{x} ^0) \in \ell_{-}^{w}(\mathbb{R}^d) $ it follows from Lemma \ref{cr using product maps} that ${\mathcal H}$ maps into $\ell_{-}^{w}(\mathbb{R}^d) $ and that it is differentiable, as required. \quad $\blacksquare$

\medskip

In some occasions it is  important to determine if a given filter is invertible. The differentiability of reservoir filters associated to reservoir systems associated to differentiable reservoir and readout maps that we established in the previous result allows us to use the inverse function theorem to formulate a sufficient invertibility condition. As we see in the next statement, this criterion can be written down entirely  in terms of the derivatives of the reservoir and the readout maps.

\begin{corollary}
Consider a reservoir system determined by a reservoir map $F:  \mathbb{R}^N \times  {\Bbb R}^n \longrightarrow  \mathbb{R}^N$ and a readout map $h: \mathbb{R} ^N\longrightarrow \mathbb{R}^d  $ that are of class $C ^1(\mathbb{R}^N \times  \mathbb{R}^n)$ and  $C ^1(\mathbb{R}^N) $, respectively, and additionally satisfy the conditions spelled out in the statement of  Corollary \ref{differentiability of full system}. Let ${\bf z} \in \ell_{-}^{w}(\mathbb{R}^n) $, $\mathbf{x} :=U^F({\bf z}) \in \ell_{-}^{w}(\mathbb{R}^N)$, and $\mathbf{y} :=U ^F_h({\bf z}) \in \ell_{-}^{w}(\mathbb{R}^d)$, and suppose that the map
\begin{equation}
\label{condition invertibility}
D{\mathcal H}(\mathbf{x}) \circ \left(\mathbb{I}_{\ell_{-}^{w}(\mathbb{R}^N)}- \left(\prod_{t \in \mathbb{Z}_{-}}D_xF \left({\bf x} _{t-1}, {\bf z}  _t\right)\right) \circ T _1\right) ^{-1}\circ \left( \prod_{t \in \mathbb{Z}_{-}}D_z F \left( {\bf x} _{t-1}, {\bf z}  _t\right)\right): \ell_{-}^{w}(\mathbb{R}^n) \longrightarrow\ell_{-}^{w}(\mathbb{R}^d)
\end{equation}
is a linear homeomorphism (continuous linear bijection with continuous inverse) with  $\mathcal{H} $ as defined in \eqref{definition funny h}. Then there exist open neighborhoods $V _{{\bf z}} \subset \ell_{-}^{w}(\mathbb{R}^n)$ and $V _{{\bf y}} \subset \ell_{-}^{w}(\mathbb{R}^N)$ of ${\bf z}  $ and ${\bf y} $, respectively, such that the restriction of the filter $U ^F_h|_{V _{{\bf z}}}:V _{{\bf z}} \longrightarrow V _{{\bf y}} $ has an inverse $\left(U ^F_h|_{V _{{\bf z}}}\right) ^{-1} $. When the condition \eqref{condition invertibility} is satisfied for all the solutions $({\bf z}, U^F({\bf z})) $ of the reservoir system determined by $F $ then the reservoir filter $U ^F_h $ admits a global inverse $\left(U ^F_h\right) ^{-1}:U ^F_h \left(\ell_{-}^{w}(\mathbb{R}^n)\right)\longrightarrow \ell_{-}^{w}(\mathbb{R}^n) $.
\end{corollary}

\noindent\textbf{Proof.\ \ }It is a straightforward consequence of the inverse function theorem as formulated in \cite[page 670]{Schechter:Handbook} (see also \cite{Eecke:1974}) applied to the Fr\'echet derivative of $U ^F_h= {\mathcal H} \circ U^F $ at the point ${\bf z} \in \ell_{-}^{w}(\mathbb{R}^n)  $. It is easy to see using the chain rule and \eqref{expression derivative as operator} (which is in turn a consequence of \eqref{recursions U2}) that this derivative coincides with the operator in \eqref{condition invertibility} whose invertibility we require. \quad $\blacksquare$

\subsection{The local versus the global echo state property}
\label{The local versus the global echo state property}

Theorem \ref{Persistence of the ESP and FMP properties} emphasizes the local nature of both the echo state and the fading memory properties by providing a sufficient condition that ensures the existence of a locally defined causal and time-invariant filter around a given solution that is shown to have the FMP. In contrast with this local approach, Theorem \ref{characterization of reservoir differentiability} characterizes the existence of a globally defined  differentiable filter associated to a given reservoir system, that hence satisfies the FMP and the ESP for any input.

Even though the conditions in Theorems \ref{Persistence of the ESP and FMP properties} and \ref{characterization of reservoir differentiability} are very alike, the latter is much stronger than the former. In the following paragraphs we illustrate with a family of ESNs of the type introduced in Section \ref{examples fmp} how it is possible to be in violation of the global condition of Theorem \ref{characterization of reservoir differentiability} and nevertheless to find solutions of such reservoir systems around which one can locally define FMP reservoir filters. This example illustrates how {\it the ESP and the FMP are structural features of a reservoir system when considered globally but are mostly input dependent when considered only locally}. This important observation has already been noticed in \cite{Manjunath:Jaeger}  where, using tools coming from the theory of non-autonomous dynamical systems,  sufficient conditions have been formulated (see, for instance, \cite[Theorem 2]{Manjunath:Jaeger}) that ensure the ESP in connection to a given specific input. The differentiability conditions that we impose to our reservoir systems allow us to draw similar conclusions and, additionally,  to automatically conclude the FMP of the resulting locally defined reservoir filters.

Consider the one-dimensional echo state map $F: \mathbb{R}\times \mathbb{R} \longrightarrow\mathbb{R} $, where
\begin{equation}
\label{reservoir for esn example}
F(x,z):=\sigma(a x+z), \quad \mbox{with} \quad a \in \mathbb{R} \quad \mbox{and } \quad
\sigma(x):= \frac{x}{\sqrt{1+ x ^2}}.
\end{equation}
The sigmoid function $\sigma$ in this expression has been chosen so that we can provide algebraic expressions in the following developments. Similar conclusions could nevertheless be drawn using other popular squashing functions.

The  function $\sigma$ maps the real line into the interval $[-1,1] $ and it is easy to see, using the notation introduced in the examples \ref{differentiability examples}, that $L _\sigma:=\sup_{x \in \mathbb{R}}\{| \sigma' (x)|\}  =1$. Moreover, the one-dimensional character of the system makes that, in this case, 
\begin{equation}
\label{lfx for esn example}
L_{F _x}= |a|. 
\end{equation}
Consequently, by Lemma \ref{contraction condition with lfx}, the reservoir map $F$ is a contraction on the first entry if and only if $|a|<1 $, in which case, by Theorem \ref{characterization of fmp unbounded}, the associated ESN has the ESP and the FMP with respect to any input in $\ell_{-}^{w}(\mathbb{R}^n) $, where $w$ is a weighting sequence that satisfies
\begin{equation}
\label{fmp for esn example}
|a|L _w<1.
\end{equation}
The FMP holds with respect to any sequence $w$ if we consider uniformly bounded inputs by Corollary \ref{fmp when you restrict to bounded}. Moreover, a well-known result for ESNs due to H. Jaeger (see \cite[Proposition 3]{jaeger2001}) shows that the ESP cannot be satisfied whenever
\begin{equation}
\label{jaeger condition no esp}
|a|>1.
\end{equation} 
Additionally, the global sufficient differentiability condition \eqref{persistence condition global} in Theorem \ref{characterization of reservoir differentiability} states that the condition \eqref{fmp for esn example} also ensures that the ESP filter is also differentiable. 

We now prove using Theorem \ref{Persistence of the ESP and FMP properties} {\it the existence of locally defined FMP filters associated to this ESN in a neighborhood of certain inputs, even when condition \eqref{jaeger condition no esp} is satisfied which, as we already mentioned, prevents the global existence of such objects.}

Notice first that the solutions of the equation $\sigma(ax)=x $, $x \in \mathbb{R} $, are characterized by the relation
\begin{equation}
\label{fixed point sigma example}
a ^2x ^2=x ^2(a ^2x ^2+1)
\end{equation}
that has as solutions
\begin{equation*}
\left\{
\begin{array}{ccc}
x _0 & = & 0,\\
x _{a}^{\pm} & = &\pm \frac{\sqrt{a ^2-1}}{a},
\end{array}
\right.
\end{equation*}
where the solutions in the second line obviously exist and are different from the first one only when $|a|>1 $, a condition that we assume holds true in the rest of the section. The condition \eqref{fixed point sigma example} implies that the constant sequences $(\mathbf{x} _0, {\bf z} _0)$ and $(\mathbf{x} _a^{\pm}, {\bf z} _0)$ defined by
\begin{equation*}
(\mathbf{x} _0, {\bf z} _0) _t:= (x _0, 0) \quad \mbox{and} \quad (\mathbf{x} _a^{\pm}, {\bf z} _0)_t:=(x _a ^{\pm},0), \quad \mbox{for any} \quad t \in \mathbb{Z}_{-},
\end{equation*}
are solutions of the reservoir system determined by $F$. Moreover, in the notation of Theorem \ref{Persistence of the ESP and FMP properties}, it is easy to see that
\begin{equation*}
L_{F _x}(\mathbf{x} _0, {\bf z} _0)=|a|>1 \quad \mbox{and } \quad L_{F _x}(\mathbf{x} _a^{\pm}, {\bf z} _0)= \frac{1}{a ^2}<1.
\end{equation*}
The persistence condition \eqref{persistence condition} in that result implies that for any weighting sequence that satisfies
\begin{equation*}
\frac{L _w}{a ^2}<1,
\end{equation*}
 there exist  open time-invariant neighborhoods $V_{\mathbf{x} _a^{\pm}} $ and $V_{\mathbf{z}^0} $ of $\mathbf{x} _a^{\pm} $ and $\mathbf{z} ^0 $  in $ \ell_{-}^{w}(\mathbb{R}^N)$ and $ \ell_{-}^{w}(\mathbb{R}^n)$, respectively, such that the reservoir system associated to $F$ with inputs in $V_{\mathbf{z}^0} $ has the echo state property and hence determines a unique causal, time-invariant, and FMP reservoir filter 
$
U^F:(V_{\mathbf{z}^0}, \left\|\cdot \right\|_w) \longrightarrow (V_{\mathbf{x}^0}, \left\|\cdot \right\|_w)
$. 

\subsection{Remote past input independence and the state forgetting property for unbounded inputs}
\label{Remote past input independence and the state forgetting property for unbounded inputs}

In Section \ref{The fading memory property and remote past input independence} we saw how fading memory filters presented with uniformly bounded inputs exhibit  what we called the uniform input forgetting property. An analysis of the proof of the main result in that section, namely Theorem \ref{FMP and the uniform input forgetting property}, shows that the compactness of the space of inputs guaranteed the existence of a modulus of continuity for the filter, which ensured the validity of the input forgetting property and, moreover, it made it uniform. In the context of reservoir systems, we saw in  Theorems \ref{characterization of fmp unbounded} and \ref{characterization of reservoir differentiability} that there are very weak hypotheses that, even when the  inputs are not uniformly bounded, guarantee that the associated reservoir filters are Lipschitz and hence have a modulus of continuity. This allows us to prove an input forgetting property in that more general context.

\begin{theorem}[Input forgetting property for FMP reservoir filters]
\label{Input forgetting property for FMP reservoir filters}
Let $F:  D _N \times  D _n \longrightarrow  D_N$  be a reservoir map where $D _n \subset {\Bbb R}^n $, $D _N\subset {\Bbb R}^N $, $n, N  \in \mathbb{N}^+$. Assume that the hypotheses of Theorem \ref{characterization of fmp unbounded} part {\bf (ii)} or \ref{characterization of reservoir differentiability} part {\bf (i)} are satisfied. Let $
U^F:(V _n, \left\|\cdot \right\|_w) \longrightarrow ((D_N)^{\mathbb{Z}_{-}}\cap \ell_{-}^{w}(\mathbb{R}^N), \left\|\cdot \right\|_w)
$ be the associated causal and time-invariant reservoir filter ($V _n \subset (D_n) ^{\mathbb{Z}_{-}} \cap \ell_{-}^{w}(\mathbb{R}^n)$ under the hypotheses of Theorem \ref{characterization of fmp unbounded}; $V _n= \ell_{-}^{w}(\mathbb{R}^n) $ and $D _N= \mathbb{R}^N $ under the hypotheses of   Theorem \ref{characterization of reservoir differentiability}). Then, for any $\mathbf{u}, \mathbf{v} \in \ell_{-}^{w}(\mathbb{R}^n) $ and  ${\bf z} \in (D_N)^{\mathbb{Z}_{-}}$ we have that
\begin{equation}
\label{property past input independence}
\lim\limits_{t \rightarrow + \infty} \left\|U^F(\mathbf{u}, {\bf z})_t-U^F (\mathbf{v}, {\bf z})_t\right\|=0.
\end{equation}
\end{theorem}

\noindent\textbf{Proof.\ \ } It mimics the proof of Theorem \ref{FMP and the uniform input forgetting property} using as modulus of continuity the map $ \omega_{U ^F}(t):= L_{U ^F} t $, $t \geq 0 $, where $L_{U ^F} $ is the Lipschitz constant whose existence is ensured by the hypotheses of Theorem \ref{characterization of fmp unbounded} or \ref{characterization of reservoir differentiability} and given by \eqref{lips in case continuous Lz} or by \eqref{lip constant uf}. \quad $\blacksquare$

\begin{remark}
\normalfont
As we saw in Remark \ref{extension of characterization of fmp to pw}, Theorem \ref{characterization of fmp unbounded} can be extended to continuous reservoir systems with inputs and outputs in $\ell_{-}^{p,w}(\mathbb{R}^n) $ and $\ell_{-}^{p,w}(\mathbb{R}^N) $, respectively. In particular, we saw that the resulting filters are Lipschitz and hence have a non-trivial modulus of continuity. This implies that a result analogous to Theorem \ref{Input forgetting property for FMP reservoir filters} can be proved for such systems that hence could also be referred to as fading memory from a dynamical point of view.
\end{remark}

When filters are differentiable, there is one more way to measure how they forget inputs simply by looking at their partial derivatives with respect to past input components. The result is a differential input forgetting property that, unlike Theorem \ref{Input forgetting property for FMP reservoir filters}, can be formulated in a uniform way even when the inputs are not uniformly bounded.

\begin{theorem}[Differential uniform input forgetting property]
\label{Differential uniform input forgetting property}
Assume that the hypotheses of Theorem \ref{characterization of reservoir differentiability} {\bf (i)} are satisfied. Let $D_{z ^i _t}H ^F({\bf z}) \in \mathbb{R}^N$ be the partial derivative of the reservoir functional $H ^F: \ell_{-}^{w}(\mathbb{R}^n) \longrightarrow \mathbb{R}^N $ with respect to the $i$-th component of the $t$-th entry of ${\bf z} \in \ell_{-}^{w}(\mathbb{R}^n) $. Then, there exists a monotonously decreasing sequence $w ^F $ with zero limit  such that, for any $t \in \mathbb{Z}_{-}  $,
\begin{equation}
\label{differential input forgetting formula}
\left\|D_{z ^i _t}H ^F({\bf z})\right\|\leq w ^F_{-t}, \quad \mbox{for any} \quad {\bf z} \in \ell_{-}^{w}(\mathbb{R}^n)\quad \mbox{and } \quad i \in \left\{1, \ldots, n\right\}.
\end{equation}
\end{theorem}

\noindent\textbf{Proof.\ \ } Let $\mathbf{e}^{i,t}:= \left(\ldots, {\bf 0}, \mathbf{e}^i, {\bf 0}, \ldots, {\bf 0}\right)\in \ell_{-}^{w}(\mathbb{R}^n)$, where the vector $\mathbf{e}^i $ is the canonical vector in ${\Bbb R}^n $ and it is placed in the $t$-th position. Then, since $\left\|\mathbf{e}^{i,t}\right\|_w = w _{-t} $, we have by \eqref{lip constant uf} and for any ${\bf z} \in \ell_{-}^{w}(\mathbb{R}^n) $ that
\begin{equation*}
\left\|D_{z ^i _t}H ^F({\bf z})\right\|= \left\|DH ^F({\bf z})\cdot \mathbf{e}^{i,t}\right\| \leq \vertiii{DH ^F({\bf z})}_w\left\|\mathbf{e}^{i,t}\right\|_w\leq \vertiii{p _0 \circ DU ^F({\bf z})}_w w _{-t}\leq \frac{L_{F _z}}{1-L_{F _x}L _w} w _{-t},
\end{equation*}
which proves \eqref{differential input forgetting formula} by setting 
\begin{equation*}
w _t^F:= \frac{L_{F _z}}{1-L_{F _x}L _w} w _{t}, \quad t \in \mathbb{N}. \quad \blacksquare
\end{equation*}

Apart from the filters that reservoir maps define when they have the echo state property, we can also use this object to define controlled forward-looking dynamical systems and flows. Indeed, given $F:  D _N \times  D _n \longrightarrow  D_N$ a reservoir map, we denote by $U ^F: (D_n)^{\mathbb{N}^{+}} \times D _N \longrightarrow (D_N)^{\mathbb{N}^{+}}$ the {\bfi  reservoir flow} associated to $F$ that is uniquely determined by the recurrence relations:
\begin{equation}
\label{recursions for reservoir flow}
\left\{
\begin{array}{ccl}
U ^F({\bf z}, \mathbf{x} _0) _1& =&F(\mathbf{x} _0, {\bf z} _1) \quad \mbox{with} \quad {\bf z} \in  (D_n)^{\mathbb{N}^{+}}, \ \mathbf{x} _0 \in D_N,\\
U ^F({\bf z}, \mathbf{x} _0) _t&=&F(U ^F({\bf z}, \mathbf{x} _0) _{t-1}, {\bf z} _t), \quad t > 1.
\end{array}
\right.
\end{equation} 
The value $\mathbf{x} _0 \in D_N $ is called the {\bfi  initial condition} of the {\bfi  path}  $U ^F({\bf z}, \mathbf{x} _0) \in (D_N)^{\mathbb{N}^{+}}$ associated to the {\bfi  input} or  {\bfi  control sequence} ${\bf z} \in  (D_n)^{\mathbb{N}^{+}}$. 

As we saw in Theorems \ref{characterization of fmp unbounded} and \ref{characterization of reservoir differentiability}, the contracting property on the first component in a reservoir map is much related to the ESP and the FMP of the resulting reservoir filter and, in passing, (see Theorem \ref{Input forgetting property for FMP reservoir filters}) to the input forgetting property. The next result shows that something similar happens with reservoir flows associated to contracting reservoir maps as they {\it forget} the influence of initial conditions that are used to create the paths. This feature is referred to as the {\bfi  state forgetting property} in \cite{jaeger2001}.

\begin{theorem}[State forgetting property for contracting reservoir flows]
Let $F:  D _N \times  D _n \longrightarrow  D_N$  be a reservoir map where $D _n \subset {\Bbb R}^n $, $D _N\subset {\Bbb R}^N $, $n, N  \in {\mathbb{N}^{+}}$, and suppose that $F$ is a contraction on the first component. Given an input sequence ${\bf z} \in  (D_n)^{\mathbb{N}^{+}}$, the reservoir flow $U ^F: (D_n)^{\mathbb{N}^{+}} \times D _N \longrightarrow (D_N)^{\mathbb{N}^{+}}$ associated to $F$ satisfies that: 
\begin{equation}
\label{State forgetting property}
\lim\limits_{t \rightarrow + \infty} \left\|U^F({\bf z}, \mathbf{x}_0)_t-U^F ({\bf z}, \overline{\mathbf{x}}_0)_t\right\|=0, \quad \mbox{for any} \quad \mathbf{x} _0, \overline{\mathbf{x}}_0 \in D_N.
\end{equation}
\end{theorem}

\noindent\textbf{Proof.\ \ } Let $c<1 $ be the contraction constant of $F$. Using the  recursions \eqref{recursions for reservoir flow} that define the reservoir flow we can write that for any $t >1 $:
\begin{multline*}
\left\|U^F({\bf z}, \mathbf{x}_0)_t-U^F ({\bf z}, \overline{\mathbf{x}}_0)_t\right\|=
\left\|F(U ^F({\bf z}, \mathbf{x} _0) _{t-1}, {\bf z} _t)- F(U ^F({\bf z}, \overline{\mathbf{x}} _0) _{t-1}, {\bf z} _t)\right\|\\\leq c \left\|U^F({\bf z}, \mathbf{x}_0)_{t-1}-U^F ({\bf z}, \overline{\mathbf{x}}_0)_{t-1}\right\|\leq \cdots
\leq c^{t-1} \left\|U^F({\bf z}, \mathbf{x}_0)_{1}-U^F ({\bf z}, \overline{\mathbf{x}}_0)_{1}\right\|\\\leq 
c^{t-1} \left\|F(\mathbf{x} _0, {\bf z} _1)-F(\overline{\mathbf{x} }_0, {\bf z} _1)\right\|.
\end{multline*}
Taking limits $t \rightarrow + \infty $ on both sides of this inequality yields \eqref{State forgetting property}. \quad $\blacksquare$

\subsection{Analytic reservoir filters associated to analytic reservoir maps}

The results in Section \ref{Differentiable reservoir filters associated to differentiable reservoir maps} characterized the conditions under which reservoir maps of class $C ^1$ yield differentiable reservoir filters with respect to inputs and outputs in weighted sequence spaces. This setup is convenient because it is able to accommodate unbounded signals and allows for an elegant encoding of the fading memory property. However, due to a phenomenon similar to the one already observed in Remark \ref{higher order derivatives bad}, one cannot immediately obtain higher order differentiable reservoir filters out of higher order differentiable reservoir maps because, as we showed in Proposition \ref{diff of filters and functionals}, one needs roughly speaking to modify the weighted norm in the target of the map that defines the filter. This makes impossible the application  in a higher order differentiability context of the Implicit Function Theorem, which is the main tool used in the results in the previous section.  That is why in the following paragraphs we deal with analytic reservoir maps (as real valued functions) and we study the analyticity of the associated reservoir filters with respect to the supremum norm, as opposed to the weighted norms that we considered in the previous section.

Using the supremum norm implies that filter differentiability in that context, when one manages to establish it,  ensures filter continuity and not the fading memory property. In exchange, analyticity allows us to construct Taylor series expansions that, as we see later on, are discrete-time Volterra series representations. 

The next result is the analytic analog of the Local Persistence Theorem \ref{Persistence of the ESP and FMP properties} formulated using the supremum norm that proves that analytic reservoir maps have locally defined analytic reservoir filters associated around constant solutions.

\begin{theorem}[Local persistence of the ESP, continuity, and analyticity]
\label{Persistence of the ESP and FMP properties analytic}
Let $F:  \mathbb{R}^N \times  \mathbb{R}^n \longrightarrow  \mathbb{R}^N$  be a reservoir map. Suppose that $F$ is analytic and that the corresponding reservoir system \eqref{reservoir equation} has a constant solution $(\mathbf{x}^0, {\bf z}^0) \in  \mathbb{R}^N \times \mathbb{R}^n$, that is, $\mathbf{x}^0=F(\mathbf{x}^0, {\bf z} ^0)$.  Suppose, additionally, that for all $r\geq	1 $,
\begin{equation}
\label{differentiability condition local higher analytic}
L_{F,r}:=\sup_{(\mathbf{x}, {\bf z}) \in \mathbb{R}^N \times  \mathbb{R}^n}\left\{\vertiii{D^r F (\mathbf{x}, {\bf z})}\right\}<+ \infty. 
\end{equation}
Suppose that 
\begin{equation}
\label{persistence condition analytic}
L_{F _x}(\mathbf{x}^0, {\bf z}^0):=
\vertiii{D_xF ({\bf x}^0 , {\bf z}^0)} <1.
\end{equation}
Then, there exist  open time-invariant neighborhoods $V_{\mathbf{x}^0} $ and $V_{\mathbf{z}^0} $ of $\mathbf{x} ^0 $ and $\mathbf{z} ^0 $  in $ \ell_{-}^{\infty}(\mathbb{R}^N)$ and $ \ell_{-}^{\infty}(\mathbb{R}^n)$, respectively, such that the reservoir system associated to $F$ with inputs in $V_{\mathbf{z}^0} $ has the echo state property and hence determines a unique causal, time-invariant, and analytic (and hence continuous) reservoir filter 
$
U^F:(V_{\mathbf{z}^0}, \left\|\cdot \right\|_\infty) \longrightarrow (V_{\mathbf{x}^0}, \left\|\cdot \right\|_\infty)
$. 
\end{theorem}

\noindent\textbf{Proof.\ \ } It follows the same scheme as that of Theorem \ref{Persistence of the ESP and FMP properties}. In the following paragraphs we just hint the additional facts that need to be taken into account in order to adapt that proof to this setup.

The first complementary fact has to do with the second part of Lemma \ref{preparation with 1k} which, using the hypothesis \eqref{differentiability condition local higher analytic} allows us to conclude that the map $\mathcal{F}:  \ell_{-}^{\infty}(\mathbb{R}^N) \times \ell^{\infty}_-(\mathbb{R}^n)\longrightarrow \ell_{-}^{\infty}(\mathbb{R}^N)  $ defined in \eqref{script f as product} is smooth. Additionally, it can be easily seen that it is also analytic and that the radii of convergence $\rho_F $ and $\rho _{\mathcal{F}} $ of the Taylor series expansions of $F$ and $\mathcal{F}  $ around $(\mathbf{x}^0, {\bf z}^0) $ and the associated constant sequence (that we denote with the same symbol) satisfy
\begin{equation}
\label{inequality radii ff}
\rho_F\leq \rho_{\mathcal{F}}.
\end{equation}
Indeed, \eqref{expression derivative in components} implies that the Taylor series expansion of $\mathcal{F} $ around the constant sequence $(\mathbf{x}^0, {\bf z}^0) $ can be written, for any $\mathbf{u}^r:=\left(\mathbf{u}, \ldots, \mathbf{u}\right)= \left((\mathbf{u}_{\mathbf{x}}, \mathbf{u}_{\mathbf{z}}), \ldots, (\mathbf{u}_{\mathbf{x}}, \mathbf{u}_{\mathbf{z}})\right) \in \left(\ell_{-} ^{\infty}(\mathbb{R}^N)\oplus \ell_{-} ^{\infty}(\mathbb{R}^n)\right)^r$, as
\begin{multline}
\label{taylor series script ff}
\mathcal{F}(\mathbf{x}^0, {\bf z}^0)+\sum_{r=1}^{\infty} \frac{1}{r!}D^r\mathcal{F}(\mathbf{x}^0, {\bf z}^0)\cdot (\mathbf{u}-(\mathbf{x}^0, {\bf z}^0))^r=\prod_{t \in \mathbb{Z}_{-}} \left(F _t(\mathbf{x}^0, {\bf z}^0)+\sum_{r=1}^{\infty} \frac{1}{r!}D^r F_t (\mathbf{x}^0, {\bf z}^0)\cdot (\mathbf{u}-(\mathbf{x}^0, {\bf z}^0))^r\right)\\
=\prod_{t \in \mathbb{Z}_{-}} \left(F (\mathbf{x}^0, {\bf z}^0)+\sum_{r=1}^{\infty} \frac{1}{r!}D^rF (\mathbf{x}_{t-1}, {\bf z}_t)\circ  \left(p _t \circ(T _{1} \times {\rm id}), \ldots, p _t \circ(T _{1} \times {\rm id})\right)\cdot (\mathbf{u}-(\mathbf{x}^0, {\bf z}^0))^r\right).
\end{multline}
Suppose now that $\mathbf{u}=(\mathbf{u}_{\mathbf{x}}, \mathbf{u}_{\mathbf{z}}) \in  \ell_{-} ^{\infty}(\mathbb{R}^N)\oplus \ell_{-} ^{\infty}(\mathbb{R}^n) $ is chosen such that 
\begin{equation}
\label{choice of rhou}
\left\|\mathbf{u}\right\|_{\infty}=\left\|\mathbf{u}_{\mathbf{x}}\right\|_{\infty}+\left\|\mathbf{u}_{\mathbf{z}}\right\|_{\infty}< \rho_F.
\end{equation} 
Lemma \ref{smooth projections and time delays} implies that for any $t \in \mathbb{Z}_{-}$, we have in that case that
\begin{equation*}
\left\|p _t \circ(T _{1} \times {\rm id})(\mathbf{u})\right\|\leq \left\|\mathbf{u}\right\|_{\infty}< \rho_F
\end{equation*}
and hence we can conclude that all the series labeled by $t \in \mathbb{Z}_{-} $ in each of the factors that make up the last term of \eqref{taylor series script ff} converge for all the elements $\mathbf{u} \in  \ell_{-} ^{\infty}(\mathbb{R}^N)\oplus \ell_{-} ^{\infty}(\mathbb{R}^n) $ that satisfy \eqref{choice of rhou}. This implies that such elements are inside the radius of convergence of the Taylor series expansion of $\mathcal{F} $ around the constant sequence $(\mathbf{x}^0, {\bf z}^0) $ and hence \eqref{inequality radii ff} holds which, as $\rho_F  $ is nontrivial by hypothesis, proves that $\mathcal{F} $ is analytic.

The rest of the proof can be obtained by mimicking that of Theorem \ref{Persistence of the ESP and FMP properties} where, as it is customary, we replace the weighting sequence $w$ by the constant sequence $w ^ \iota$ given by $w ^{\iota} _t:=1 $, for all $t \in \mathbb{N} $, and $L _w  $ is replaced by the constant $1$. 

A technical modification is needed at the time of invoking the Implicit Function Theorem. In Theorem \ref{Persistence of the ESP and FMP properties} we used a version that requires only first order differentiability as hypothesis and produces Lipschitz continuous implicitly defined functions. In this case we can prove that the function ${\cal G} $ is analytic and hence it can be shown that the implicitly defined local filter $
\widetilde{U^F}:(\widetilde{V}_{\mathbf{z}^0}, \left\|\cdot \right\|_w) \longrightarrow (\widetilde{V}_{\mathbf{x}^0}, \left\|\cdot \right\|_w)
$ is analytic by invoking, for instance, \cite[page 175]{valent:elasticity}, and references therein. \quad $\blacksquare$

\section{The Volterra series representation of analytic filters and a universality theorem}
\label{The Volterra series representation of analytic filters and a universality theorem}

In this section we study the Taylor series expansions of analytic causal and time-invariant filters that, as we prove in the next result, coincide with the so called discrete-time Volterra series representations. A very similar result has been formulated in \cite{sandberg:time-delay, sandberg:volterra} for analytic filters with respect to the supremum norm and with inputs with a finite past. The next result extends that statement and characterizes the inputs for which an analytic time-invariant  fading memory filter  with respect to a weighted norm admits a Volterra series representation with semi-infinite past inputs. This generalized result allows this series representation for inputs that are not necessarily bounded.  Additionally, we  use the causality and time-invariance hypotheses to show that the corresponding Volterra series representations have time-independent coefficients.
\begin{theorem}
\label{Volterra series representation}
Let $w $ be a weighting sequence and let $U:B _{\left\|\cdot \right\|_w}({{\bf z}} ^0,M)\subset \ell_{-}^{w}(\mathbb{R})\longrightarrow B _{\left\|\cdot \right\|_w}(U({{\bf z}} ^0),L)\subset \ell_{-}^{w}(\mathbb{R}^N) $ be a causal and time-invariant analytic filter, for some time-invariant ${{\bf z}}^0\in  \ell_{-}^{1, w}(\mathbb{R}) $ (that is, $T _{-t}({\bf z} ^0)= {\bf z}^0 $, for all $t \in \mathbb{Z}_{-} $)  and $M, L > 0 $. Then, for any element in the domain that satisfies  
\begin{equation}
\label{condition domain volterra}
{{\bf z}}\in B _{\left\|\cdot \right\|_w}({{\bf z}} ^0,M)\cap \ell_{-}^{1, w}(\mathbb{R}), \quad \mbox{that is} \quad \sum_{t \in \mathbb{Z}_{-}}|z _t|w _{-t}< + \infty,
\end{equation}
there exists a unique expansion
\begin{equation}
\label{Taylor equals volterra}
U({{\bf z}})_t=U ({{\bf z}}^0)_t+\sum_{j=1}^{\infty}\sum_{m _1=- \infty}^{0} \cdots \sum_{m _j=- \infty}^{0}g _j(m _1, \ldots, m _j)(z_{m _1+t}-z^0_{m _1+t})\cdots (z_{m _j+t}-z^0_{m _j+t}), \quad t \in \mathbb{Z}_{-},
\end{equation}
where the maps $g _j: \mathbb{Z}_{-}^j \longrightarrow\mathbb{R}^N $, $j\geq 1 $, are uniquely determined by the derivatives of the functional $H_U:B _{\left\|\cdot \right\|_w}({{\bf z}} ^0,M)\subset \ell_{-}^{w}(\mathbb{R})\longrightarrow \mathbb{R}^N $ associated to $U$ (that by Proposition \ref{diff of filters and functionals} is analytic) via the relation
\begin{equation}
\label{definition gs}
g _j(m _1, \ldots, m _j):= \frac{1}{j!}D ^jH({\bf z} ^0) \left(e_{m _1}, \ldots, e_{m _j}\right)\quad \mbox{with} \quad \left(e _n\right)_t:=\left\{
\begin{array}{cl}
1 & \mbox{ if } t=n, \\
0 & \mbox{ otherwise. }
\end{array}
\right.
\end{equation}
Moreover, for any $p \in {\mathbb{N}^{+}} $, we have that
\begin{multline}
\label{error estimation volterra}
\left\|U({{\bf z}})_t-U ({{\bf z}}^0)_t-\sum_{j=1}^{p}\sum_{m _1=- \infty}^{0} \cdots \sum_{m _j=- \infty}^{0}g _j(m _1, \ldots, m _j)(z_{m _1+t}-z^0_{m _1+t})\cdots (z_{m _j+t}-z^0_{m _j+t})\right\|\\
\leq \frac{L}{w _{-t}} \left(1- \frac{\left\|{\bf z}\right\|_w}{M}\right)^{-1} \left(\frac{\left\|{\bf z}\right\|_w}{M}\right)^{p+1}.
\end{multline}
These statements also hold true when $\ell_{-} ^{w}(\mathbb{R})$ and $\ell_{-}^{w}(\mathbb{R}^N)$ are replaced by $\ell_{-} ^{\infty}(\mathbb{R})$ and $\ell_{-}^{\infty}(\mathbb{R}^N)$, respectively. In that case, the relation \eqref{Taylor equals volterra} holds whenever ${{\bf z}}\in B _{\left\|\cdot \right\|_\infty}({{\bf z}} ^0,M)\cap \ell_{-}^{1}(\mathbb{R}^n)$ and the inequality \eqref{error estimation volterra} is obtained by taking as the sequence $w$  the constant sequence $w ^ \iota$ given by $w ^{\iota} _t:=1 $, for all $t \in \mathbb{N} $.
\end{theorem}

\begin{remark}
\normalfont
The error estimate \eqref{error estimation volterra} can be reformulated in terms of the weighted norm of the sequence
\begin{equation*}
\mathbf{R }_p({\bf z}):=\left(U({{\bf z}})_t-U ({{\bf z}}^0)_t-\sum_{j=1}^{p}\sum_{m _1=- \infty}^{0} \cdots \sum_{m _j=- \infty}^{0}g _j(m _1, \ldots, m _j)(z_{m _1+t}-z^0_{m _1+t})\right)_{t \in \mathbb{Z}_{-}},
\end{equation*}
as
\begin{equation}
\label{bounds for error remark}
\left\|\mathbf{R} _p({\bf z})\right\|_w\leq L \left(1- \frac{\left\|{\bf z}\right\|_w}{M}\right)^{-1} \left(\frac{\left\|{\bf z}\right\|_w}{M}\right)^{p+1}.
\end{equation}
\end{remark}

\subsection{Finite discrete-time Volterra series are universal in the fading memory category}

In this section we combine the Volterra series representation Theorem \ref{Volterra series representation} with previous universality results  in \cite{RC6}  to show that any fading memory filter with uniformly bounded inputs can be arbitrarily well approximated with a Volterra series with finite terms of the type in \eqref{Taylor equals volterra}. This result provides an alternative proof of a Volterra series universality theorem that was stated for the first time in \cite[Theorems 3 and 4]{Boyd1985}. In particular, this result shows that any time-invariant and causal fading memory filter can be uniformly approximated by a finite memory filter.

\begin{theorem}[Universality of finite discrete-time Volterra series]
\label{Volterra series are universal}
Let $M,L >0 $  and let $K _M \subset \left(\mathbb{R}\right)^{\mathbb{Z}_{-}}, K_L \subset \left(\mathbb{R}^d\right)^{\mathbb{Z}_{-}} $ be as in \eqref{Kset}. Let $U: K _M \longrightarrow K _L $ be  a causal and time-invariant fading memory filter. Then, for any $\epsilon>0 $ there exist $\mathbf{x} ^0 \in K _L$ and $J \in {\mathbb{N}^{+}} $ such that for any $j \in \left\{1, \ldots, J\right\} $ there exist $j$ numbers $M _1 ^j, \ldots, M _j^j \in {\mathbb{N}^{+}}$ and maps $g _j: \mathbb{Z}_{-}^j \longrightarrow\mathbb{R} $ such that the filter determined by the finite Volterra series given by
\begin{equation}
\label{volterra finite later}
V ({\bf z})_t= \mathbf{x} ^0_t+\sum_{j=1}^J\sum_{m _1=-M _1^j}^0 \cdots\sum_{m _j=-M _j^j}^0 g _j(m _1, \ldots, m _j)z_{m _1+t} \cdots z_{m _j+t}
\end{equation}
is such that 
\begin{equation*}
\vertiii{U-V}_ \infty=\sup_{{\bf z}\in K _M}\left\{ \left\|U({\bf z})-V({\bf z})\right\|\right\}< \epsilon.
\end{equation*}
\end{theorem}

\noindent\textbf{Proof.\ \ } The Corollary 11 in \cite{RC6} guarantees that for any $\epsilon>0 $ there exists a linear reservoir system with polynomial readout $h\in \mathbb{R}[\mathbf{x}]$ and nilpotent connectivity matrix $A \in \mathbb{M}_N $, determined by the expressions
\begin{empheq}[left={\empheqlbrace}]{align}
{\bf x } _t & =A\mathbf{x}_{t-1}+ {\bf c}{\bf z} _t,\quad A \in \mathbb{M} _N, {\bf c} \in \mathbb{M} _{N,n},
\label{linear reservoir equation}
\\
y _t & =h (\mathbf{x} _t), \quad h \in \mathbb{R}[\mathbf{x}], \label{linear readout}
\end{empheq}
such that  it has an associated reservoir filter $U^{A, {\bf c}}_h :K _M \longrightarrow K _L$ that satisfies
\begin{equation}
\label{linear approx well}
\vertiii{U-U^{A, {\bf c}}_h}_ \infty< \epsilon.
\end{equation}
Let $J=\deg(h)+1 $ and assume that $A$ is nilpotent of index $p$. In order to prove the theorem it suffices to show that the Volterra series expansion in \eqref{Taylor equals volterra} corresponding to $U^{A, {\bf c}}_h $ has an expression of the type \eqref{volterra finite later}. If that is the case, the statement in \eqref{linear approx well} proves the theorem.

Indeed, recall (see, for instance, \cite[Corollary 11]{RC6}) that the functional $H^{A, {\bf c}}_h $ associated to the filter $U^{A, {\bf c}}_h $ is given by
\begin{equation*}
H^{A, {\bf c}}_h({\bf z})=h\left(\sum_{j=0}^{p-1}A ^j({\bf c} z_{-j})\right),
\end{equation*}
which is a composition of the polynomial $h$ with the functional $H^{A, {\bf c}} $ associated to the reservoir equation \eqref{linear reservoir equation} given by the linear operator
\begin{equation}
\label{linear for nilpotent nil}
H^{A, {\bf c}}({\bf z}):=\sum_{j=0}^{p-1}A ^j({\bf c} z_{-j}).
\end{equation} 
It is easy to see that  $H^{A, {\bf c}}: (\ell_{-}^{\infty}(\mathbb{R}), \left\|\cdot \right\|_ \infty) \longrightarrow\mathbb{R}^N $ has a finite operator norm $\vertiii{H^{A, {\bf c}}} _ \infty$ and that $\vertiii{H^{A, {\bf c}}} _ \infty\leq \vertiii{{\bf c}}/(1-\vertiii{A})$, with $\vertiii{{\bf c}}$ and $\vertiii{A}$ the top singular values of ${\bf c} $ and $A $, respectively. Moreover, it is easy to see that for any $j \in {\mathbb{N}^{+}} $, ${\bf z} \in K _M $, and $\mathbf{v} _1, \ldots, \mathbf{v}_j \in  \ell_{-}^{\infty}(\mathbb{R}) $, we have
\begin{equation*}
D^jH^{A, {\bf c}}_h({\bf z})(\mathbf{v} _1, \ldots, \mathbf{v}_j)=D ^jh(H^{A, {\bf c}}({\bf z})) \left(H^{A, {\bf c}}(\mathbf{v} _1), \ldots, H^{A, {\bf c}}(\mathbf{v}_j))\right),
\end{equation*}
which shows that $H^{A, {\bf c}}_h: (\ell_{-}^{\infty}(\mathbb{R}), \left\|\cdot \right\|_ \infty) \longrightarrow\mathbb{R}^d $ is everywhere analytic.
Using this expression and \eqref{definition gs} we define
\begin{equation}
\label{definition ref taylor linear}
g _j(m _1, \ldots, m _j):= \frac{1}{j!}D ^jh({\bf 0}) \left(H^{A, {\bf c}}(e_{m _1}), \ldots, H^{A, {\bf c}}(e_{m _j})\right).
\end{equation}
As $h$ has finite degree  then $D^jh({\bf 0})=0 $ for any $j>\deg(h)+1=J $. Moreover, since the sum in \eqref{linear for nilpotent nil} is finite by the nilpotency of $A$ it is clear that $g _j(m _1, \ldots, m _j) $ in \eqref{definition ref taylor linear} is nonzero as long as $1\leq j\leq \deg(h)+1=J $ and $-(p-1)\leq m _1, \ldots, m _j\leq 0 $. If we define $M _1^j, \ldots, M _j ^j:= p-1 $ then the Taylor series expansion of $U^{A, {\bf c}}({\bf z}) $ coincides with \eqref{volterra finite later}.

We emphasize that in this case this expansion is valid for any ${\bf z}\in K _M  $ by the finiteness of the number of terms in the sum and that the condition \eqref{condition domain volterra} is hence not necessary. \quad $\blacksquare$

\section{Appendices}

\subsection{Proof of Lemma \ref{topological lemma balls etc}}

\noindent\textbf{(i)} We prove \eqref{balls for the lwtopology} by double inclusion. First, let $\mathbf{x} \in B_{\left\|\cdot \right\|_w}({\bf z}, r) $. By definition $\left\|\mathbf{x} - {\bf z}\right\|_w= \sup _{t \in \mathbb{Z}_{-}}\left\{\left\|\mathbf{x} _t-\mathbf{z} _t\right\|w _{-t}\right\}< r $ and hence for any $\delta _x> 0 $ such that $\left\|\mathbf{x} - {\bf z}\right\|_w< \delta _x < r $ we have that  $\left\|\mathbf{x} _t-\mathbf{z} _t\right\|< \delta _x/ w _{-t}  $, for all $t \in \mathbb{Z}_{-}  $. This implies that
\begin{equation*}
\mathbf{x} \in   \prod_{t \in \mathbb{Z}_{-}}B_{\left\|\cdot \right\|}\left(\mathbf{z} _t, \frac{\delta _x}{w _{-t}}\right)  \subset  \bigcup_{\delta< r} \left(\prod_{t \in \mathbb{Z}_{-}}B_{\left\|\cdot \right\|}\left(\mathbf{z} _t, \frac{\delta}{w _{-t}}\right)\right).
\end{equation*}

Conversely, given an element $\mathbf{x} \in \ell_{-}^{w}(\mathbb{R}^n) $ in the right hand side of \eqref{balls for the lwtopology}, there exists $\delta _x< r  $ such that $\mathbf{x} \in   \prod_{t \in \mathbb{Z}_{-}}B_{\left\|\cdot \right\|}\left(\mathbf{z} _t, \delta _x/w _{-t}\right) $. This implies that $\left\|\mathbf{x} _t- {\bf z} _t\right\|w _{-t}< \delta _x $, for all $t \in \mathbb{Z}_{-} $, and hence  $\sup_{t \in \mathbb{Z}_{-}}\left\{\left\|\mathbf{x} _t- {\bf z} _t\right\|w _{-t}\right\}= \left\|\mathbf{x}- {\bf z}\right\|_w\leq \delta _x < r$, which proves the inclusion.

As to \eqref{inclusions balls}, the first inclusion is a straightforward consequence of \eqref{balls for the lwtopology}. Let now $\mathbf{x} \in \prod_{t \in \mathbb{Z}_{-}}B_{\left\|\cdot \right\|}\left(\mathbf{z} _t, \frac{r}{w _{-t}}\right) $. By definition this  implies that $\left\|\mathbf{x} _t- {\bf z}_t\right\| w _{-t}< r $, for all $t \in \mathbb{Z}_{-} $, and consequently $\sup_{t \in \mathbb{Z}_{-}}\left\{\left\|\mathbf{x} _t- {\bf z}_t\right\| w _{-t}\right\} \leq r $ or, equivalently, $\left\|\mathbf{x}- {\bf z}\right\|_w\leq r $. This implies that $\mathbf{x} \in \overline{B_{\left\|\cdot \right\|_w}({\bf z}, r)} $ and proves the second inclusion.

\medskip

\noindent {\bf (ii)} Let $\mathbf{x} \in \prod_{t \in \mathbb{Z}_{-}}A _t $. Then,
 \begin{equation*}
\left\|\mathbf{x}\right\|_w=\sup _{t\in\mathbb{Z}_{-}}\left\{\left\|\mathbf{x} _t\right\|w _{-t}\right\}\leq \sup _{t\in\mathbb{Z}_{-}}\left\{c _t\right\}<+ \infty,
\end{equation*}
as required.

\medskip

\noindent {\bf (iii)} We first prove that $\prod_{t \in \mathbb{Z}_{-}}\overline{A _t} \subset \overline{\prod_{t \in\mathbb{Z}_{-}}A _t}$. If ${\bf z}\in \prod_{t \in \mathbb{Z}_{-}}\overline{A _t} $, then for any $\epsilon> 0 $  and each $t \in \mathbb{Z}_{-}  $ there exists an element $\mathbf{x}_t \in A _t\cap B_{\left\|\cdot \right\|}\left({\bf z} _t, \frac{\epsilon}{2 w _{-t}}\right)$. Let $\mathbf{x} := \left(\mathbf{x} _t\right)_{t\in\mathbb{Z}_{-}} $. By construction:
\begin{equation*}
\left\|\mathbf{x}- {\bf z} \right\|_w=\sup _{t\in\mathbb{Z}_{-}}\left\{\left\|\mathbf{x} _t- {\bf z} _t\right\|w _{-t}\right\}\leq \frac{ \epsilon}{2}< \epsilon,
\end{equation*}
which implies that $\mathbf{x} \in B_{\left\|\cdot \right\|_w} \left({\bf z}, \epsilon\right)\cap \prod _{t\in\mathbb{Z}_{-}} A _t $ and, as ${\bf z}\in \prod_{t \in \mathbb{Z}_{-}}\overline{A _t} $ is arbitrary, it guarantees that ${\bf z} \in  \overline{\prod_{t \in\mathbb{Z}_{-}}A _t}$.

In order to show the reverse inclusion first note that, as it is proved later on in Lemma \ref{smooth projections and time delays}, the projections $p _t: \ell_{-}^{w}(\mathbb{R}^n)\longrightarrow {\Bbb R}^n $, $t \in \mathbb{Z}_{-}  $, defined by $p _t({\bf z}):= {\bf z} _t $, are continuous. Let ${\bf z} \in  \overline{\prod_{t \in\mathbb{Z}_{-}}A _t}$ arbitrary, let $t \in \mathbb{Z}_{-} $ be arbitrary but fixed,  and let $V _t $ be an open set in ${\Bbb R}^n $ that contains ${\bf z} _t $. The continuity of $p _t $ implies that $p _t^{-1}(V _t  ) $ is an open set in $\ell_{-}^{w}(\mathbb{R}^n) $ that contains ${\bf z} $ and therefore there exists $\mathbf{x} \in \left(\prod_{t \in \mathbb{Z}_{-}}A _t\right)\cap p _t^{-1}(V _t  )$. We consequently have that $\mathbf{x}_t \in A _t $, which guarantees that $ {\bf z} _t \in \overline{A _t} $, as required. \quad $\blacksquare$

\subsection{Proof of Corollary \ref{dnn closed and open}}

\noindent\textbf{(i)} We proceed by contradiction. Suppose that $D_n\neq {\Bbb R}^n $. Let $\mathbf{x} _0 \in {\Bbb R}^n \setminus D _n$ and let ${\bf z} _0 \in D_n $. Define the constant sequences $\mathbf{x}:= \left(\mathbf{x} _0\right)_{t\in\mathbb{Z}_{-}} \in\ell_{-}^{w}(\mathbb{R}^n)\setminus \left((D _n) ^{\mathbb{Z}_{-}}\cap \ell_{-}^{w}(\mathbb{R}^n)\right)$ and $\mathbf{z}:= \left(\mathbf{z} _0\right)_{t\in\mathbb{Z}_{-}} \in  (D _n) ^{\mathbb{Z}_{-}}\cap \ell_{-}^{w}(\mathbb{R}^n) $. Since by hypothesis $(D _n) ^{\mathbb{Z}_{-}}\cap \ell_{-}^{w}(\mathbb{R}^n)  $ is an open subset of $\ell_{-}^{w}(\mathbb{R}^n) $ there exists $\epsilon> 0 $  such that $B_{\left\|\cdot \right\|_w}({\bf z}, 2 \epsilon)\subset (D _n) ^{\mathbb{Z}_{-}}\cap \ell_{-}^{w}(\mathbb{R}^n) $. By the relation \eqref{inclusions balls} in Lemma \ref{topological lemma balls etc} we also have
\begin{equation*}
B_{\left\|\cdot \right\|_w}({\bf z}, \epsilon)\subset \prod_{t \in \mathbb{Z}_{-}}B_{\left\|\cdot \right\|}\left(\mathbf{z} _0, \frac{\epsilon}{w _{-t}}\right)\subset  \overline{B_{\left\|\cdot \right\|_w}({\bf z}, \epsilon)} \subset B_{\left\|\cdot \right\|_w}({\bf z}, 2 \epsilon)\subset (D _n) ^{\mathbb{Z}_{-}}\cap \ell_{-}^{w}(\mathbb{R}^n),
\end{equation*}
and, in particular,
\begin{equation}
\label{inclu produs}
\prod_{t \in \mathbb{Z}_{-}}B_{\left\|\cdot \right\|}\left(\mathbf{z} _0, \frac{\epsilon}{w _{-t}}\right)\subset  (D _n) ^{\mathbb{Z}_{-}}\cap \ell_{-}^{w}(\mathbb{R}^n),\mbox{ which implies $B_{\left\|\cdot \right\|}\left(\mathbf{z} _0, \frac{\epsilon}{w _{-t}}\right)\subset  D _n$, for all $t \in \mathbb{Z}_{-} $.} 
\end{equation}
Let $r_0:= \left\|\mathbf{x} _0- {\bf z}_0\right\|$ and let $t _0  \in \mathbb{Z}_{-}$ be such that  for all $t < t _0 $ we have that $\epsilon /w _{-t_0} > r _0 $. By \eqref{inclu produs} we have that $\mathbf{x} _0 \in B_{\left\|\cdot \right\|}\left(\mathbf{z} _0, \frac{\epsilon}{w _{-t}}\right) \subset D_n $, which contradicts the assumption on the choice of $\mathbf{x}_0$.

\medskip

\noindent {\bf (ii)} By Lemma \ref{topological lemma balls etc} {\bf (iii)} we have that 
\begin{equation}
\label{closure of product is product of closures}
\overline{(D _n) ^{\mathbb{Z}_{-}}}=\left( \overline{D _n} \right)^{\mathbb{Z}_{-}}.
\end{equation} Since by hypothesis ${(D _n) ^{\mathbb{Z}_{-}}}$ is closed and hence it holds true that
\begin{equation}
\label{closed}
\overline{(D _n) ^{\mathbb{Z}_{-}}}={(D _n) ^{\mathbb{Z}_{-}}}.
\end{equation}
Consequently, by \eqref{closure of product is product of closures} and \eqref{closed} we have that $\left( \overline{D _n} \right)^{\mathbb{Z}_{-}} = {(D _n) ^{\mathbb{Z}_{-}}}$ which implies that $\overline{D_n} = D_n$ as required.

\medskip

\noindent {\bf (iii)} Let $\mathbf{x} \in \overline{(D_n)^{\mathbb{Z}_{-}}\cap \ell_{-}^{w}(\mathbb{R}^n) }\subset  \ell_{-}^{w}(\mathbb{R}^n) $ and consider a sequence $\left\{\mathbf{x}^m \right\}_{m \in \mathbb{N}^+} \subset (D_n)^{\mathbb{Z}_{-}}\cap \ell_{-}^{w}(\mathbb{R}^n) $ with  
$\lim_{m \rightarrow \infty} \mathbf{x}^m = \mathbf{x}$, that is for each $\epsilon>0$ there exists such $N(\epsilon)\in \mathbb{N}^+$ such that for all $m>N(\epsilon)$ it holds that $\| \mathbf{x}^m- \mathbf{x}\| _{w} < \epsilon$. Hence for all $s \in {\mathbb{Z}_{-}}$ one has that
\begin{equation*}
w_{-s} \| \mathbf{x}_s^m - \mathbf{x}_s\| \le \sup _{t\in\mathbb{Z}_{-}}\left\{\left\|\mathbf{x} _t^m- {\bf x} _t\right\|w _{-t}\right\} = \|\mathbf{x}^m - \mathbf{x}\|_w \leq  \epsilon,
\end{equation*}
which immediately implies that 
\begin{equation*}
\| \mathbf{x}^m _s- \mathbf{x}_s\| < \dfrac{\epsilon}{w_{-s}}
\end{equation*}
and hence one gets that $\mathbf{x}_s\in \overline{D_n}$ and therefore \eqref{inclusion with bars dn} holds as required.

\noindent The last claim in part {\bf(iii)} follows from  \eqref{inclusion with bars dn}. Indeed, if $D_n = \overline{D_n}$ then by  \eqref{inclusion with bars dn} we have that 
\begin{equation*}
\overline{(D _n) ^{\mathbb{Z}_{-}}\cap \ell_{-}^{w}(\mathbb{R}^n)} \subset \left(\overline{D _n}\right) ^{\mathbb{Z}_{-}}\cap \ell_{-}^{w}(\mathbb{R}^n) =  \left({D _n}\right) ^{\mathbb{Z}_{-}}\cap \ell_{-}^{w}(\mathbb{R}^n).
\end{equation*}
Since the reverse inclusion obviously always holds, we finally have that 
\begin{equation*}
\overline{(D _n) ^{\mathbb{Z}_{-}}\cap \ell_{-}^{w}(\mathbb{R}^n)}=  \left({D _n}\right) ^{\mathbb{Z}_{-}}\cap \ell_{-}^{w}(\mathbb{R}^n).  \quad \blacksquare
\end{equation*}

\subsection{Proof of Lemma \ref{inclusions with lw}}

\noindent The continuity of the inclusions \eqref{inclusions continuous} and \eqref{inclusions continuous powers} is a consequence of the fact that:
\begin{eqnarray}
\left\| \mathbf{z}\right\|_{w^{\frac{1}{k}}} &\leq &\left\| \mathbf{z}\right\|_{w^{\frac{1}{k+1}}}, \quad \mbox{for all $k \in \mathbb{N}^+$ and ${\bf z}\in \ell ^{w^{\frac{1}{k}}}_-({\Bbb R}^n)$,}\label{comparison norm root}\\
\left\| \mathbf{z}\right\|_{w^{k+1}} &\leq &\left\| \mathbf{z}\right\|_{w^{k}}, \quad \mbox{for all $k \in \mathbb{N}^+ $ and ${\bf z}\in \ell ^{w^{k}}_-({\Bbb R}^n)$.}\label{comparison norm power}
\end{eqnarray}
Regarding \eqref{properties sw}, the first inclusion follows from the fact that $\ell_{-} ^{\infty}(\mathbb{R}^n) \subset \ell ^{w}_-({\Bbb R}^n)$ for any weighting sequence. In order to show that this inclusion is in general not an equality it suffices to consider the following example: let ${\bf z} \in (\mathbb{R})^{\mathbb{Z}_-} $ given by $z _t:= -t $, $t \in \mathbb{Z}_{-}  $,  and let $w$ be the weighting sequence defined by $w _t:= \lambda^{t} $, with $t \in \mathbb{N}$ and $0< \lambda<1 $. A simple application of the L'H\^opital  rule shows that, for any $k \in \mathbb{N}^+$,
\begin{equation*}
\lim_{t \rightarrow - \infty} z _t w _{-t}^{1/k}=0,
\end{equation*}
which proves, in particular, that $\left\|{\bf z}\right\|_{w ^{1/k}} < \infty  $ and hence that ${\bf z} \in \ell_{-} ^{w^{1/k}}(\mathbb{R})  $, for any $k \in \mathbb{N}^+ $. This implies that $ {\bf z} \in S _w $. However, $ {\bf z} $ is an unbounded sequence and hence it does not belong to $\ell_{-} ^{\infty}(\mathbb{R})$. In order to show that the second inclusion in \eqref{properties sw} is also strict, take ${\bf z} \in (\mathbb{R})^{\mathbb{Z}_-} $ given by $z _t:= \lambda^{-t} $ with $\lambda> 1  $ and $t \in \mathbb{Z}_{-} $ and let $w$ be the weighting sequence defined by $w _0:=1 $ and  $w _t:= \frac{1}{t}$, for any $t \in \mathbb{N}^+$. The L'H\^opital  rule shows that, for any $k \in\mathbb{N}^+$,
\begin{equation*}
\lim_{t \rightarrow - \infty} |z _t w _{-t}^{k}|=+\infty,
\end{equation*}
and consequently ${\bf z}  $ does not belong to any  of the spaces $\ell_{-} ^{w^{k}}(\mathbb{R}) $ and hence ${\bf z} \not\in S ^w $. \quad $\blacksquare$

\subsection{Proof of Lemma \ref{cr using product maps}}

\noindent\textbf{(i)} The compactness of $D_N $ guarantees \cite[Theorem 27.3]{Munkres:topology} that there exists $L>0 $ such that  $D_N \subset \overline{B_{\left\|\cdot \right\|}({\bf 0}, L)} $ and hence $(D_N)^{\mathbb{Z}_{-}} \subset \ell_{-}^{w}(\mathbb{R}^N) $ necessarily.  It can also be shown (see \cite[Corollary 2.7]{RC7}) that when $D _N  $ is compact, the relative topology $(D_N)^{\mathbb{Z}_{-}} $ induced by the weighted norm $\left\|\cdot \right\|_w  $ in $\ell_{-}^{w}(\mathbb{R}^N) $ coincides with the product topology. This implies (see  \cite[Theorem 19.6]{Munkres:topology}) that if the functions $H _t $ are continuous then so is $\mathcal{H} $.

\medskip

\noindent {\bf (ii)} Let ${\bf z}^1, {\bf z}^2\in W $. Then,
\begin{equation}
\label{Lip ineq qq}
\left\|\mathcal{H}({\bf z}^1)- \mathcal{H}({\bf z}^2)\right\|_w= \sup_{t \in \mathbb{Z}_{-}}\left\{\left\|H _t ({\bf z}^1)-H _t ({\bf z}^2)\right\|w _{-t}\right\}\leq 
\sup_{t \in \mathbb{Z}_{-}}\left\{c_t^0\left\|{\bf z}^1 -{\bf z}^2 \right\|w _{-t}\right\}\leq \|c ^0\| _w\left\|{\bf z}^1 -{\bf z}^2\right\|,
\end{equation}
which proves simultaneously that ${\mathcal H} $ is Lipschitz continuous and that it maps into $\ell_{-}^{w}(\mathbb{R}^N) $. Regarding the last point, recall that by hypothesis there exists a point ${\bf z}^0 $ such that ${\mathcal H}({\bf z}^0)\in \ell_{-}^{w}(\mathbb{R}^N)$ and hence by \eqref{Lip ineq qq} we have, for any ${\bf z} \in W $,
\begin{equation}
\label{maps into lw}
\left\|{\mathcal H}({\bf z})\right\|_w\leq \|c ^0\| _w\left\|{\bf z} -{\bf z}^0\right\|+\left\|{\mathcal H}({\bf z}^0)\right\|_w< + \infty.
\end{equation}

\medskip

\noindent {\bf (iii)} First, it is easy to prove recursively that for any ${\bf z} \in W $, the map $D^r\mathcal{H} ({\bf z}) :=\prod _{t \in \mathbb{Z}_{-}} D^rH_t({\bf z} )$ satisfies the condition \eqref{definition frechet}. In order to prove the first statement of the lemma, it suffices to show that the multilinear map 
\begin{equation*}
D ^r \mathcal{H} ({\bf z}):\underbrace{ \left(V , \left\|\cdot \right\|\right)\times \cdots\times \left(V , \left\|\cdot \right\|\right)}_{\mbox{$r$ times}}   \longrightarrow (\ell_{-} ^{w}(\mathbb{R}^n) , \left\|\cdot \right\|_{w}), 
\end{equation*}
is bounded for any ${\bf z} \in W $. Let $\left(\mathbf{v}^1, \ldots, \mathbf{v} ^r\right) \in V ^r $. Using the $r$-order differentiability of $H_t$ we can write
\begin{multline}
\label{drf for ht}
\left\|D^r \mathcal{H}({\bf z})\cdot \left(\mathbf{v}^1, \ldots, \mathbf{v} ^r\right)\right\|_{w}=\left\|\prod _{t \in \mathbb{Z}_{-}}D^r H_t({\bf z})\cdot \left(\mathbf{v}^1, \ldots, \mathbf{v} ^r\right)\right\|_{w}=\sup_{t \in \mathbb{Z}_{-}}\left\{\left\|D^r H _t({\bf z})\cdot \left(\mathbf{v}^1, \ldots, \mathbf{v} ^r\right)\right\|w_{-t} \right\}\\\leq 
\sup_{t \in \mathbb{Z}_{-}}\left\{\vertiii{D^r H _t({\bf z})}\left\| \mathbf{v}^1 \right\|  \cdots\left\| \mathbf{v} ^r  \right\| w_{-t}\right\}\\\leq
\left\| \mathbf{v}^1 \right\|  \cdots\left\| \mathbf{v} ^r  \right\| 
\sup_{t \in \mathbb{Z}_{-}}\left\{c ^r _tw_{-t}\right\}\leq 
\|c^r\| _w\left\| \mathbf{v}_1\right\|  \cdots \left\| \mathbf{v}_r\right\|  ,
\end{multline}
which proves the boundedness of $D^r \mathcal{H}({\bf z}) $  and the inequality in \eqref{norm of r derivative with wrglobal}.  

We now assume that $ c ^j \in \ell_{-}^{w}(\mathbb{R}) $ for all $j \in \left\{1, \ldots, r\right\}$ and show that $\mathcal{H}$ maps into $\ell_{-}^{w}(\mathbb{R}^N) $ and that it is of class $C^{r-1}(W) $. Notice, first of all, that for any $t \in \mathbb{Z}_{-} $ and any ${\bf z}^1, {\bf z}^2  \in V_n $, we have by the convexity of $W $, the mean value theorem \cite{mta}, and the hypothesis $H _t \in C ^r(W)$, that for all $j \in \left\{1, \ldots, r\right\}$,
\begin{equation}
\label{intermediate ft}
\vertiii{D^{j-1}H _t({\bf z}^1)-D^{j-1}H _t({\bf z}^2)}_w\leq
\sup_{{\bf z} \in W}\left\{\vertiii{D^{j}H _t({\bf z})} \right\} \left\|{\bf z}^1-{\bf z}^2\right\| =
c ^j_t\left\|{\bf z}^1-{\bf z}^2\right\|.
\end{equation}
Taking $j=1$ in the previous inequality, we see that the functions $H _t $ are Lipschitz continuous with constants $c _t^1$ that form a sequence that by hypothesis belongs to $\ell_{-}^{w}(\mathbb{R})$. This guarantees by part {\bf (ii)} that $\mathcal{H}$ maps into $\ell_{-}^{w}(\mathbb{R}^N) $ necessarily.
Now, using the  inequality \eqref{intermediate ft}, we have that for any ${\bf z}^1, {\bf z}^2\in W $,
\begin{multline}
\label{lipschitz differential}
\vertiii{D^{r-1}\mathcal{H}({\bf z}^1)-D^{r-1}\mathcal{H}({\bf z}^2)}_{w}\\
=\sup_{{\mathbf{v}^1, \ldots, \mathbf{v} ^{r-1} \in V \atop \mathbf{v}^1, \ldots, \mathbf{v} ^{r-1} \neq {\bf 0}}} 
\left\{
\frac{\left\|\left(D^{r-1} {\mathcal{H}}({\bf z}^1 )-D^{r-1} {\mathcal{H}}({\bf z}^2)\right)\cdot \left(\mathbf{v}^1, \ldots, \mathbf{v} ^{r-1}\right)\right\|_{w }}{\left\|\mathbf{v}^1\right\|  \cdots \left\|\mathbf{v}^{r-1}\right\| }\right\} \\
= \sup_{{\mathbf{v}^1, \ldots, \mathbf{v} ^{r-1} \in V \atop \mathbf{v}^1, \ldots, \mathbf{v} ^{r-1} \neq {\bf 0}}} 
\left\{
\frac{\sup_{t \in \mathbb{Z}_{-}}\left\{\left\|\left(D^{r-1} {H_t}({\bf z}^1 )-D^{r-1} {H_t}({\bf z}^2)\right)\cdot \left(\mathbf{v}^1, \ldots, \mathbf{v} ^{r-1}\right)\right\|w _{-t}\right\}}{\left\|\mathbf{v}^1\right\|  \cdots \left\|\mathbf{v}^{r-1}\right\| }\right\} \\
\leq \sup_{{\mathbf{v}^1, \ldots, \mathbf{v} ^{r-1} \in V \atop \mathbf{v}^1, \ldots, \mathbf{v} ^{r-1} \neq {\bf 0}}} 
\left\{
\frac{\sup_{t \in \mathbb{Z}_{-}}\left\{ c ^r _t w _{-t}\left\|{\bf z}^1 - {\bf z}^2 \right\| \cdot \|\mathbf{v}^1\|  \cdots\| \mathbf{v} ^{r-1}\|   \right\}}{\left\|\mathbf{v}^1\right\| \cdots \left\|\mathbf{v}^{r-1}\right\| }\right\} =\|c ^r\| _w \left\|{\bf z}^1 - {\bf z}^2 \right\| ,
\end{multline}
which shows that the map
$D ^{r-1} \mathcal{H}: (W, \left\|\cdot \right\|_w) \longrightarrow \left(L ^{r-1}(\ell_{-} ^{w}(\mathbb{R}^n), \ell_{-} ^{w}(\mathbb{R}^N)), \vertiii{\cdot }_{w}\right)  $ is Lipschitz continuous with Lipschitz constant $c _{{\mathcal H}} ^r\leq \|c ^r\| _w $.

\medskip

\noindent {\bf (iv)} The previous part of the lemma together with the hypothesis $c := \sup_{r \in \mathbb{N}^+} \left\{\|c ^r\|_{w}\right\}< + \infty$ guarantees that the differentiability of any order in the functions $H_t$ gets translated into the differentiability of any order of the map $\mathcal{H}: W \subset (V, \left\|\cdot \right\|)\longrightarrow (\ell_{-} ^{w}(\mathbb{R}^N) , \left\|\cdot \right\|_{w}) $. Moreover, let $ \mathbf{u} \in \ell_{-} ^{w}(\mathbb{R}^n)  $  and let $\mathbf{u}^r := (\mathbf{u}, \ldots, \mathbf{u})  \in \left(\ell_{-} ^{w}(\mathbb{R}^n)\right) ^r$, $r \in \mathbb{N}^+ $. The Taylor series expansion of $\mathcal{H}$ around ${\bf 0} \in $ is
\begin{equation}
\label{Taylor fprod}
\mathcal{H}({\bf 0})+\sum_{r=1}^{\infty} \frac{1}{r!}D^r\mathcal{H}({\bf 0})\cdot \mathbf{u}^r=\prod_{t \in \mathbb{Z}_{-}} \left(H _t( {\bf 0})+\sum_{r=1}^{\infty} \frac{1}{r!}D^r H_t ({\bf 0})\cdot \mathbf{u} ^r\right).
\end{equation}
The expansion in the left hand side of this equality is convergent if and only if each of the series in the product in the right hand side is convergent. This is the case when $\left\| \mathbf{u} \right\|_{w}< \rho _t $, for all $t \in \mathbb{Z}_{-} $, which guarantees the convergence of the Taylor series expansion in \eqref{Taylor fprod} for all the elements $\mathbf{u} \in \ell_{-} ^{w}(\mathbb{R}^n) $ that satisfy  $\left\|\mathbf{u}\right\|_ w<\inf_{ t \in \mathbb{Z}_{-}}\left\{\rho _t\right\}= \rho $. Since by hypothesis $\rho>0$, we have proved that $\mathcal{H}   $ is analytic with radius of convergence $\rho_{\mathcal{H}}\geq \rho $.

\medskip

The proof of  the statements in  {\bf (ii)}, {\bf (iii)}, and {\bf (iv)} for the space $\ell_{-}^{\infty}(\mathbb{R}^N) $  is obtained by mimicking the proofs that we just provided, replacing the weighting sequence $w$ by the constant sequence $w^ \iota $ that is equal to $1$ for each $t \in \mathbb{Z}_{-}$. In order to show that part {\bf (i)} is in general false in that situation take $W=(-1,1) $, $D _N=[-1,1] $, and define $H _t(z):= \tanh \left(-t z\right)$, with $t \in \Bbb Z $ and  $ z \in (-1,1) $. Given that $H _t^{-1} \left(- \frac{1}{2}, \frac{1}{2}\right)= \left(-\frac{1}{t}\tanh ^{-1}(- \frac{1}{2}), -\frac{1}{t}\tanh ^{-1}(\frac{1}{2})\right)$ it is clear that
\begin{equation*}
{\mathcal H}^{-1}\left(B_{\left\|\cdot \right\|_ \infty} \left({\bf 0}, \frac{1}{2}\right)\right)=\bigcap_{t \in \mathbb{Z}_{-}}\left(-\frac{1}{t}\tanh ^{-1}\left(- \frac{1}{2}\right), -\frac{1}{t}\tanh ^{-1}\left(\frac{1}{2}\right)\right)= \left\{0\right\}.
\end{equation*}
This equality shows that the preimage by the product map ${\mathcal H}  $ of an open set is not open and hence ${\mathcal H}  $ is not continuous. 
\quad $\blacksquare$

\subsection{Proof of Lemma \ref{smooth projections and time delays}}

\noindent {\bf (i)} The linearity of $p _t  $ is obvious. Let $\mathbf{u} \in \ell_{-}^{w}(\mathbb{R}^n)$ arbitrary. Since $\left\|p _t(\mathbf{u})\right\|= \left\|\mathbf{u}_t\right\|=\left\|\mathbf{u}_t\right\| w _{-t}/  w _{-t} \leq \sup_{j \in \mathbb{Z}_{-}} \left\{\left\|\mathbf{u}_j\right\|w_{-j}\right\}/ w _{-t}= \left\|\mathbf{u}\right\|_w/ w _{-t}$, we can conclude that $\vertiii{p _t}_w\leq 1/  w _{-t} $. Let now $\mathbf{v} \in {\Bbb R}^n  $ such that $\left\| \mathbf{v}\right\|=1  $ and define the element ${\bf z} \in \ell_{-}^{w}(\mathbb{R}^n)  $ by ${\bf z}_t:= \mathbf{v}/ w _{-t} $, for all $t \in \mathbb{Z}_{-} $. It is clear that $\left\|{\bf z}\right\|_w=1 $ and that $\left\|p _t({\bf z})\right\|/ \left\|{\bf z}\right\| _w =1/ w _{-t} $, which shows that $\vertiii{p _t}_w= 1/ w _{-t}$, as required.

\medskip

\noindent {\bf (ii)} We first prove the statements in this part in the case $t<0 $. Suppose that the inverse decay ratio $L _w $ is finite and let $\mathbf{u} \in \ell_{-}^{w}(\mathbb{R}^n)$ arbitrary. Then
\begin{multline}
\label{t-1 bounded}
\left\|T_{1}(\mathbf{u})\right\|_w=\sup_{t \in \mathbb{Z}_{-}}\left\{\left\|\mathbf{u} _{t-1}\right\|w _{-t}\right\}=
\sup_{t \in \mathbb{Z}_{-}}\left\{\left\|\mathbf{u} _{t-1}\right\|w_{-(t-1)} \frac{w_{-t}}{w_{-(t-1)}}\right\}\\
\leq
\sup_{t \in \mathbb{Z}_{-}}\left\{\left\|\mathbf{u} _{t-1}\right\|w_{-(t-1)}\right\} \sup_{t \in \mathbb{Z}_{-}}\left\{ \frac{w_{-t}}{w_{-(t-1)}}\right\} \leq \left\|\mathbf{u}\right\|_w L _w.
\end{multline}
This inequality shows that $T_{1} $ maps $\ell_{-} ^{w}(\mathbb{R}^n) $ into $\ell_{-} ^{w}(\mathbb{R}^n) $ and that $\vertiii{T_{1}}_w\leq  L _w $. Given that for any $t \in \mathbb{Z}_{-} $ we can write
\begin{equation*}
T _{-t} =\underbrace{T_{1}  \circ  \cdots\circ T_{1}}_{\mbox{$-t$ times}},
\end{equation*}
the previous conclusion also proves that $T _{-t}  $ maps $\ell_{-} ^{w}(\mathbb{R}^n) $ into $\ell_{-} ^{w}(\mathbb{R}^n) $ and that $\vertiii{T_{-t}}_w= \vertiii{T_{1}\circ  \cdots \circ T_{1}}_w \leq \vertiii{T_{1}} _w \cdots\vertiii{T_{1}} _w\leq L _w^{- t}$. It remains to be shown that $\vertiii{ T _{1}} _w= L _w $. In order to do so, take an element $\mathbf{v} \in {\Bbb R}^n  $ such that $\left\| \mathbf{v}\right\|=1  $ and define the element ${\bf u} \in \ell_{-}^{w}(\mathbb{R}^n)  $ by ${\bf u}_t:= \mathbf{v}/w_{-t} $, for all $t \in \mathbb{Z}_{-} $. Notice that by construction $\left\|{\bf u}\right\|_w=1 $ and, moreover,
\begin{equation*}
\frac{\|T_{1}({\bf u})\|_w}{\left\|{\bf u}\right\|_w}=\sup_{t \in \mathbb{Z}_{-}}\left\{\left\|\mathbf{u} _{t-1}\right\|w _{-t}\right\}=\sup_{t \in \mathbb{Z}_{-}}\left\{\frac{\left\|\mathbf{v}\right\|}{w _{-(t-1)}}w _{-t}\right\}=L _w,
\end{equation*}
which proves the required identity.

We now show that $T _{-t}: (\ell_{-} ^{w}(\mathbb{R}^n), \left\|\cdot \right\|_w) \longrightarrow (\ell_{-} ^{w}(\mathbb{R}^n), \left\|\cdot \right\|_w) $ is surjective. Indeed, it is clear that for any $ {\bf u}  \in  \ell_{-}^{w}(\mathbb{R}^n) $, the element 
\begin{equation*}
\widetilde{ \mathbf{u}}:= (\mathbf{u}, \underbrace{{\bf 0}  ,  \ldots, {\bf 0}}_{\mbox{$-t$ times}})  \quad \mbox{is such that } \quad T _{-t} \left(\widetilde{\mathbf{u}}\right) =\mathbf{u}.
\end{equation*}
We hence just need to show that $\widetilde{ \mathbf{u}} \in \ell_{-}^{w}(\mathbb{R}^n) $. This is the case because
\begin{multline}
\label{belongs to right space}
\left\|\widetilde{\mathbf{u}}\right\|_w=\sup_{s \in \mathbb{Z}_{-}}\left\{\left\|\widetilde{\mathbf{u}_s}\right\| w_{-s}\right\}=
\sup_{s \in \mathbb{Z}_{-}}\left\{\left\|{\mathbf{u}_{s+t}}\right\| w_{-s}\right\}=
\sup_{s \in \mathbb{Z}_{-}}\left\{\left\|{\mathbf{u}_{s+t}}\right\| w_{-(s+t)}\frac{w_{-s}}{w_{-(s+t)}}\right\} \\
=\sup_{s \in \mathbb{Z}_{-}}\left\{\left\|{\mathbf{u}_{s+t}}\right\| w_{-(s+t)}\frac{w_{-s}}{w_{-(s+1)}}\frac{w_{-(s+1)}}{w_{-(s+2)}} \cdots \frac{w_{-(s+t-1)}}{w_{-(s+t)}}\right\}\leq \left\|\mathbf{u}\right\|_w L _w^{-t}< + \infty,
\end{multline}
because $ {\bf u}  \in  \ell_{-}^{w}(\mathbb{R}^n) $ and by hypothesis $L _w< + \infty $. Now, since we already showed that $T _{-t}: (\ell_{-} ^{w}(\mathbb{R}^n), \left\|\cdot \right\|_w) \longrightarrow (\ell_{-} ^{w}(\mathbb{R}^n), \left\|\cdot \right\|_w) $ is continuous, then the Banach-Schauder Open Mapping Theorem \cite[Theorem 2.2.15]{mta} implies that $T _{-t} $ is necessarily an open map.

It remains to be shown that $T _{-t}: (\ell_{-} ^{w}(\mathbb{R}^n), \left\|\cdot \right\|_w) \longrightarrow (\ell_{-} ^{w}(\mathbb{R}^n), \left\|\cdot \right\|_w) $ is a submersion (see \cite[Section 3.5]{mta} for context and definitions). First, it is obvious that
\begin{equation*}
\ker T _{-t}= \left\{ \left(\ldots, {\bf 0}, {\bf 0}, \mathbf{v}\right)\mid \mathbf{v} \in (\mathbb{R}^n)^{-t} \right\}.
\end{equation*}
Since $T _{-t}$ is linear and bounded, in order to show that it is a submersion it suffices to show that $\ker T _{-t} $ is split, that is, it has a closed complement in $(\ell_{-} ^{w}(\mathbb{R}^n), \left\|\cdot \right\|_w) $. We now prove that such a complement is given by the subspace
\begin{equation}
\label{complement kernel tt}
C _{-t}:= \left\{  (\mathbf{u}, \underbrace{{\bf 0}  ,  \ldots, {\bf 0}}_{\mbox{$-t$ times}}) \mid \mathbf{u} \in \ell_{-}^{w}(\mathbb{R}^n)\right\}.
\end{equation}
The inequality \eqref{belongs to right space} implies that
\begin{equation}
\label{inclusion in lw}
C _{-t}\subset \ell_{-}^{w}(\mathbb{R}^n).
\end{equation}
Additionally, $C _{-t}$ is clearly closed in $\ell_{-}^{w}(\mathbb{R}^n)$. We conclude by showing by double inclusion that
\begin{equation}
\label{split condition to show}
\ell_{-}^{w}(\mathbb{R}^n)=\ker T _{-t}\oplus C _{-t}.
\end{equation}
Let first $\mathbf{u}\in \ell_{-}^{w}(\mathbb{R}^n) $ and define
\begin{equation*}
\mathbf{u}_1:= (\ldots, \mathbf{u}_{t-2}, \mathbf{u}_{t-1}, \mathbf{u}_t, \underbrace{{\bf 0}  ,  \ldots, {\bf 0}}_{\mbox{$-t$ times}}) \quad \mbox{and} \quad
\mathbf{u}_2:=(\ldots, {\bf 0}, {\bf 0}, \mathbf{u}_{t+1}, \mathbf{u}_{t+2}, \ldots, \mathbf{u}_{0}).
\end{equation*}
It is clear that $\mathbf{u} = \mathbf{u} _1+ \mathbf{u} _2$. Additionally, the sequence $\mathbf{u}_2$ is obviously in $\ker T _{-t} $ and using an argument similar to the one in \eqref{belongs to right space} it is easy to show that $\mathbf{u}_1 \in C _{-t} $, which proves the inclusion $\ell_{-}^{w}(\mathbb{R}^n)\subseteq\ker T _{-t}\oplus C _{-t} $. 

Conversely, let $\mathbf{u}_1 \in C _{-t} $ and $\mathbf{u}_2 \in \ker T _{-t} $. By \eqref{inclusion in lw} we have that $\left\|\mathbf{u}_1\right\|_w< + \infty $ and it is also clear that $\left\|\mathbf{u}_2\right\|_w< + \infty $. Therefore $\left\|\mathbf{u}_1+ \mathbf{u} _2\right\|_w\leq \left\|\mathbf{u}_1\right\|_w+ \left\|\mathbf{u} _2\right\|_w<+ \infty $ and hence $\mathbf{u}_1+ \mathbf{u} _2 \in \ell_{-}^{w}(\mathbb{R}^n) $, which shows that $T _{-t} $ is a submersion. 

Finally, the statements in the case $t>0 $ are proved in a similar fashion. In particular, it is easy to see that 
\begin{equation*}
T _{-t} \circ T _{t}= \mathbb{P}_{C _t}, \quad \mbox{for any $t>0 $,} \quad
\end{equation*}
where $\mathbb{P}_{C _t} $ is the projection onto the subspace $C _t $  defined in \eqref{complement kernel tt} according to the splitting \eqref{split condition to show}. Moreover, it is easy to see that  $T _{-t} $ is injective and that its image ${\rm Im} \,T _{-t} $ is split because ${\rm Im} \, T _{-t} =C _t$ and by \eqref{split condition to show}
\begin{equation*}
\ell_{-}^{w}(\mathbb{R}^n)=\ker T _{t}\oplus {\rm Im}\, T _{-t},\quad t>0,
\end{equation*}
which proves that $T _{-t} $ is an immersion.

\medskip

\noindent {\bf (iii)} Straightforward consequence of the definitions.

\medskip

The proofs for the space $(\ell_{-} ^{\infty}(\mathbb{R}^n), \left\|\cdot \right\|_\infty) $ can be obtained by replacing in the previous arguments the weighting sequence $w$ by the constant sequence $w ^{\iota} $. \quad $\blacksquare$

\subsection{Proof of Proposition \ref{continuity of filters and functionals}}

\noindent\textbf{(i)} As $H _U $ is given by  $H _U=p _0\circ U $, the FMP (respectively, continuity) of $U$ and the first part of Lemma \ref{smooth projections and time delays} prove the statement.

\medskip

\noindent {\bf (ii)} Notice first that as 
\begin{equation}
\label{writing of uh as product}
U _H=\prod_{t \in \mathbb{Z}_{-}}H \circ T_{-t}
\end{equation}
then, as $L_{w^1}$ is finite,  $U _H $ is by the second part of Lemma \ref{smooth projections and time delays} the Cartesian product of continuous functions $H _t:=H \circ T _{-t}: V _n \subset \ell_{-} ^{w^1}(\mathbb{R}^n) \longrightarrow D_N $. Since $D_N  $ is by hypothesis compact, the result follows from the first part of Lemma \ref{cr using product maps}.

\medskip

\noindent {\bf (iii)} Let ${\bf z}^1, {\bf z}^2 \in V _n $ arbitrary. Then by \eqref{writing of uh as product} and the Lipschitz hypothesis on $H$, we have that
\begin{multline}
\label{to bound rw}
\left\|U _H({\bf z}^1)-U _H({\bf z}^2)\right\|_{w^2}=\sup_{t \in \mathbb{Z}_{-}} \left\{\left\|H(T _{-t}({\bf z} ^1))-H(T _{-t}({\bf z} ^2))\right\|w^2 _{-t}\right\}\\
\leq c _H\sup_{t \in \mathbb{Z}_{-}} \left\{\left\|T _{-t}({\bf z} ^1)-T _{-t}({\bf z} ^2)\right\|_{w^1} w^2 _{-t}\right\}.
\end{multline}
If the first condition in \eqref{rw definition} is satisfied, this expression is bounded above by 
\begin{multline*}
c _H\sup_{t,s \in \mathbb{Z}_{-}} \left\{\left\|T _{-t}({\bf z} ^1)_s-T _{-t}({\bf z} ^2)_s\right\|w^2 _{-t} w^1 _{-s}\right\}
=c _H\sup_{t,s \in \mathbb{Z}_{-}} \left\{\left\|{\bf z} ^1_{t+s}-{\bf z} ^2_{t+s}\right\|w^1_{-(t+s)}\frac{w^2 _{-t} w^1 _{-s}}{w^1_{-(t+s)}}\right\}\\
\leq R _{w ^1, w ^2}c _H \left\| {\bf z} ^1- {\bf z} ^2\right\|_{w^1}
\end{multline*}
which proves that in that case $U_H$ has  the fading memory property, it is Lipschitz, and $R _{w ^1, w ^2}c _H $ is a Lipschitz constant. If the second condition in \eqref{rw definition} is satisfied then the inverse decay ratio $L_{w ^1}$ is necessarily finite and hence
\eqref{to bound rw} can be bounded using the second part of Lemma \ref{smooth projections and time delays} as 
\begin{equation*}
c _H\sup_{t \in \mathbb{Z}_{-}} \left\{\left\|T _{-t}({\bf z} ^1)-T _{-t}({\bf z} ^2)\right\|_{w^1} w^2 _{-t}\right\}\leq c _H  \left\| {\bf z} ^1-{\bf z} ^2\right\|_{w^1}\sup_{t \in \mathbb{Z}_{-}} \left\{L _{w^1}^{-t} w^2 _{-t}\right\}= \|\mathcal{L} _{w^1}\| _{w^2}  c_H \left\| {\bf z} ^1-{\bf z} ^2\right\|_{w^1},
\end{equation*}
which proves that in that case $\|\mathcal{L} _{w^1}\| _{w^2}  c_H$ is a Lipschitz constant of $U _H $. The Lipschitz continuity of $U _H $ together with the hypothesis on the existence of a point ${\bf z}^0 $ such that $U_H({\bf z}^0)\in \ell_{-}^{w^2}(\mathbb{R}) $ guarantee that $U _H $ maps into $\ell_{-}^{w^2}(\mathbb{R}^N) $ using a strategy similar to the one followed in \eqref{maps into lw}.

The proof for the spaces $\ell_{-}^{\infty}(\mathbb{R}^n)$ and $\ell_{-}^{\infty}(\mathbb{R}^N) $ is obtained by taking as weighting sequences the constant sequence $w ^ \iota$ given by $w ^{\iota} _t:=1 $, for all $t \in \mathbb{N} $, that automatically satisfies any of the two conditions in  \eqref{rw definition}. \quad $\blacksquare$

\subsection{Proof of Proposition \ref{diff of filters and functionals}}

\noindent\textbf{(i)} Recall first that $H _U $ can be written as $H _U=p _0\circ U $. The chain rule and the linearity of the projection $p _0 $ imply that $D^rH _U ({\bf z})= p _0 \circ D ^rU ({\bf z}) $ for any ${\bf z} \in V _n$. The first part of Lemma \ref{smooth projections and time delays} guarantees then that $H _U $ is of class $C ^r(V _n) $ and that
\begin{equation*}
\vertiii{D ^rH _U({\bf z})}_{w ^1}= \vertiii{p _0 \circ D ^rU ({\bf z})}_{w ^1}\leq \vertiii{p _0 }_{w ^2} \cdot  \vertiii{  D ^rU ({\bf z})}_{w ^1, w ^2}=\vertiii{D ^rU({\bf z})}_{w ^1, w ^2}, \mbox{ for any $ {\bf z} \in V _n$, }
\end{equation*}
as required. The proof for the spaces $\ell_{-}^{\infty}(\mathbb{R}^n)$ and $\ell_{-}^{\infty}(\mathbb{R}^N) $ is obtained by taking as sequence $w$ the constant sequence $w ^ \iota$ given by $w ^{\iota} _t:=1 $, for all $t \in \mathbb{N} $.

\medskip

\noindent {\bf (ii)} First of all, notice that the hypothesis on $c ^1 $ and the convexity of $V _n $ imply via the mean value theorem \cite{mta} that $H$ is Lipschitz. Moreover, the hypothesis on $\mathcal{L}_{w,1} $ in the statement implies that condition \eqref{rw definition} is satisfied and hence the third part in Proposition \ref{continuity of filters and functionals} guarantees that $U _H $ maps into $\ell_{-}^{w^2}(\mathbb{R}^N) $.

Now, the expression \eqref{writing of uh as product} implies that for any ${\bf z}\in V _n $,
\begin{equation}
\label{derivative of functional}
D^rU _H({\bf z})=\prod_{t \in \mathbb{Z}_{-}}D ^r H (T _{-t}({\bf z})) \circ \underbrace{\left(T_{-t} , \ldots, T_{-t}\right)}_{\mbox{$r$ times}}, \quad r\geq 1.
\end{equation}
In order to prove \eqref{inequality norms derivatives h} consider $\mathbf{u}^1, \ldots, \mathbf{u} ^r \in \ell ^{w^1}_-({\Bbb R}^n ) $  arbitrary and notice that by the second part of Lemma \ref{smooth projections and time delays} we have
\begin{multline*}
\left\|D^rU _H({\bf z})\left( \mathbf{u}^1, \ldots, \mathbf{u} ^r\right)\right\|_{w^2}=
\sup_{t \in \mathbb{Z}_{-}}\left\{\left\|D ^r H (T _{-t}({\bf z})) \cdot \left(T_{-t}(\mathbf{u}^1), \ldots, T_{-t}(\mathbf{u}^r)\right)\right\|w^2 _{-t}\right\}\\\leq
c ^r\sup_{t \in \mathbb{Z}_{-}}\left\{\left\|T_{-t}(\mathbf{u}^1)\right\|_{w^1} \cdots\left\|T_{-t}(\mathbf{u}^r)\right\|_{w^1}w^2 _{-t}\right\}\\\leq
c ^r\sup_{t \in \mathbb{Z}_{-}}\left\{\left\| \mathbf{u}^1 \right\|_{w^1} \cdots\left\| \mathbf{u}^r \right\|_{w^1}L _{w^1}^{-rt}w^2 _{-t}\right\}\leq
c ^r \left\|\mathcal{L}_{w^1,r}\right\|_{w^2} \left\|\mathbf{u}^1\right\| _{w^1} \cdots \left\|\mathbf{u}^r\right\| _{w^1},
\end{multline*}
as required. 
We now show that $ {U _H} $ is of class $C^{r-1}(V _n) $. Let  ${\bf z}^1, {\bf z}^2\in V _n $ arbitrary. Then, using a strategy similar to that one in the last inequality in the previous expression, we have
\begin{multline*}
\vertiii{D^{r-1}U _H({\bf z}^1)-D^{r-1}U _H({\bf z}^2)}_{w ^1, w ^2}\\
=\sup_{{\mathbf{u}^1, \ldots, \mathbf{u} ^{r-1} \in \ell ^{w^1}_-({\Bbb R}^n) \atop \mathbf{u}^1, \ldots, \mathbf{u} ^{r-1} \neq {\bf 0}}} 
\left\{
\frac{\left\|\left(D^{r-1} {U _H}({\bf z}^1 )-D^{r-1} {U _H}({\bf z}^2)\right)\cdot \left(\mathbf{u}^1, \ldots, \mathbf{u} ^{r-1}\right)\right\|_{w^2 }}{\left\|\mathbf{u}^1\right\|_{w^1} \cdots \left\|\mathbf{u}^{r-1}\right\|_{w^1}}\right\} \\
= \sup_{{\mathbf{u}^1, \ldots, \mathbf{u} ^{r-1} \in \ell ^{w^1}_-({\Bbb R}^n) \atop \mathbf{u}^1, \ldots, \mathbf{u} ^{r-1} \neq {\bf 0}}} 
\left\{
\frac{\sup_{t \in \mathbb{Z}_{-}}\left\{\left\|\left(D^{r-1} {H}(T _{-t}({\bf z}^1) )-D^{r-1} {H}(T _{-t}({\bf z}^2) )\right)\cdot \left(T _{-t}(\mathbf{u}^1), \ldots, T _{-t}(\mathbf{u} ^{r-1})\right)\right\|w^2 _{-t}\right\}}{\left\|\mathbf{u}^1\right\|_{w^1} \cdots \left\|\mathbf{u}^{r-1}\right\|_{w^1}}\right\} \\
\leq \sup_{{\mathbf{u}^1, \ldots, \mathbf{u} ^{r-1} \in \ell ^{w^1}_-({\Bbb R}^n) \atop \mathbf{u}^1, \ldots, \mathbf{u} ^{r-1} \neq {\bf 0}}} 
\left\{
\frac{\sup_{t \in \mathbb{Z}_{-}}\left\{ c ^r  \left\|T _{-t}({\bf z}^1)- T _{-t}({\bf z}^2)\right\|_{w^1}\cdot \|T _{-t}(\mathbf{u}^1)\|_{w^1} \cdots\| T _{-t}(\mathbf{u} ^{r-1})\|_{w^1}  \cdot w^2 _{-t}\right\}}{\left\|\mathbf{u}^1\right\|_{w^1} \cdots \left\|\mathbf{u}^{r-1}\right\|_{w^1}}\right\} \\
=\sup_{{\mathbf{u}^1, \ldots, \mathbf{u} ^{r-1} \in \ell ^{w^1}_-({\Bbb R}^n) \atop \mathbf{u}^1, \ldots, \mathbf{u} ^{r-1} \neq {\bf 0}}} 
\left\{
\frac{\sup_{t \in \mathbb{Z}_{-}}\left\{ c ^r\left\|{\bf z}^1-  {\bf z}^2 \right\|_{w^1}\cdot \| \mathbf{u}^1 \|_{w^1} \cdots\|  \mathbf{u} ^{r-1} \|_{w^1}  L _{w^1}^{-r t}w^2 _{-t} \right\}}{\left\|\mathbf{u}^1\right\|_{w^1} \cdots \left\|\mathbf{u}^{r-1}\right\|_{w^1}}\right\} \\
\leq c ^r \left\|\mathcal{L}_{w^1,r}\right\|_{w^2}\left\|{\bf z}^1 - {\bf z}^2 \right\|_{w^1},
\end{multline*}
which shows that the map
$D ^{r-1} U _H: (V _n, \left\|\cdot \right\|_{w^1}) \longrightarrow \left(L ^{r-1}(\ell_{-} ^{w^1}(\mathbb{R}^n), \ell_{-} ^{w^2}(\mathbb{R}^N)), \vertiii{\cdot }_{w^1,w ^2}\right)  $ is Lipschitz continuous with Lipschitz constant $c ^r \left\|\mathcal{L}_{w^1,r}\right\| _{w ^2}$.

\medskip

\noindent {\bf (iii)} First, the condition $c ^r < + \infty$ for all $r \in \mathbb{N}^+ $ implies by part {\bf (ii)} that $U _H $ is smooth if $H$ is. Suppose now that we work with the supremum norm. The expression \eqref{derivative of functional} shows that the point ${\bf z} \in V _n $ belongs to the domain of convergence of the series expansion of $U _H $ if and only if all the points $T _{-t}({\bf z}  ) $ belong to the domain of convergence of the series expansion of $H$. Finally, suppose that ${\bf z} \in V _n $ belongs to the domain of convergence of the series expansion of $H$. Since $\vertiii{T _{-t}}_{\infty}\leq 1 $ for all $t \in \mathbb{Z}_{-} $ by Lemma \ref{smooth projections and time delays}, we have that $\left\|T _{-t}({\bf z})\right\|_{\infty} \leq \left\|{\bf z}\right\|_{\infty} $, which guarantees that all the points $T _{-t}({\bf z}  ) $ belong to the domain of convergence of the series expansion of $H$ and hence, by the argument above, ${\bf z} \in V _n $ belongs to the domain of convergence of the series expansion of $U _H $, which proves the statement. \quad $\blacksquare$

\subsection{Proof of Corollary \ref{fmp full system}}

\noindent Under the hypothesis in part {\bf (i)}, the continuity of $h$ implies that $h \left(D_N\right)$ is compact and hence there exists a constant $R>0 $ such that $h \left(D _N \right)\subset  \overline{B_{\left\|\cdot \right\|}({\bf 0}, R)}$. The first part of Lemma \ref{cr using product maps} guarantees that the map ${\mathcal H} :=\prod _{t \in \mathbb{Z}_{-}} h 
:((D_N)^{\mathbb{Z}_{-}}, \left\|\cdot \right\| _w)\longrightarrow (K _R, \left\|\cdot \right\| _w)$  is continuous and as $U ^F_h= {\mathcal H}\circ U^F  $ and we proved that under the hypotheses {\bf (i)} in the theorem that $U^F: \left(V _n, \left\|\cdot \right\|_w\right)\longrightarrow\left(K _L, \left\|\cdot \right\|_w\right) $ is continuous, the claim follows.

We now prove the statement under the hypotheses in part {\bf (ii)}. First, we show that if $h$ is Lipschitz continuous in $D^N$ with constant $c_h  $  then so is the map ${\mathcal H} $ in $(D_N)^{\mathbb{Z}_{-}}\cap \ell_{-}^{w}(\mathbb{R}^N)  $. Indeed, let $\mathbf{x}^1, \mathbf{x} ^2 \in (D_N)^{\mathbb{Z}_{-}}\cap \ell_{-}^{w}(\mathbb{R}^N)  $, then
\begin{equation*}
\left\|{\mathcal H}({\bf x} ^1)- {\mathcal H}({\bf x} ^2)\right\|_w=
\sup_{t\in\mathbb{Z}_{-}}\left\{\left\|h({\bf x} ^1_t)-h({\bf x} ^2_t)\right\|w _{-t}\right\}\leq c_h \left\| \mathbf{x} ^1- \mathbf{x} ^2\right\|_w.
\end{equation*}
The hypothesis $U ^F_h({\bf z}^0) \in \ell_{-}^{w}(\mathbb{R}^d) $ amounts to the fact that the point $U^F ({\bf z} ^0)\in  (D_N)^{\mathbb{Z}_{-}}\cap \ell_{-}^{w}(\mathbb{R}^N)  $ is such that ${\mathcal H} (U^F ({\bf z} ^0))\in \ell_{-}^{w}(\mathbb{R}^d) $. An argument mimicking \eqref{maps into lw} in the proof of part {\bf (ii)} in Lemma \ref{cr using product maps} proves that in those conditions ${\mathcal H}  $ maps into $\ell_{-}^{w}(\mathbb{R}^d)$.
 \quad $\blacksquare$

\subsection{Proof of Theorem \ref{Persistence of the ESP and FMP properties}}

\noindent  We start with a preliminary result whose proof mimics that of Lemma \ref{cr using product maps} and is also a consequence of Lemma \ref{smooth projections and time delays}.
As we already did in the proof of Theorem \ref{characterization of fmp unbounded}, in the statement we  consider the direct sum 
of weighted spaces $\ell_{-} ^{w}(\mathbb{R}^N)\oplus \ell_{-} ^{w}(\mathbb{R}^n) $ as a Banach space with the sum norm $\left\|\cdot \right\|_{w \oplus w} $ defined by $\left\|(\mathbf{u}, \mathbf{v}) \right\|_{w \oplus w} :=\left\|\mathbf{u} \right\|_{w}+\left\|\mathbf{v} \right\|_{w}$, for any $(\mathbf{u}, \mathbf{v}) \in \ell_{-} ^{w}(\mathbb{R}^N)\oplus \ell_{-} ^{w}(\mathbb{R}^n) $. Additionally, in all that follows $V _n$ stands for any open convex subset of the Banach space $ \left( \ell_{-}^{w}(\mathbb{R}^n), \left\|\cdot \right\|_w\right) $.

\begin{lemma}
\label{preparation with 1k}
In the hypotheses of the theorem, consider the map 
\begin{equation}
\begin{array}{cccc}
\label{definition of fcal}
\mathcal{F}: & \ell_{-}^{w}(\mathbb{R}^N) \times V _n&\longrightarrow &(\mathbb{R}^N)^{\mathbb{Z}_{-}}\\
	&(\mathbf{x}, {\bf z})&\longmapsto & \left(\mathcal{F}(\mathbf{x}, {\bf z})\right)_t:=F(\mathbf{x} _{t-1}, {\bf z}_t),
\end{array}
\end{equation}
where $V _n$ is an open convex subset of $ \ell_{-}^{w}(\mathbb{R}^n)  $. Then,
\begin{description}
\item [(i)] $\mathcal{F}  $ is Lipschitz continuous with constant $L_{F} L _w $ and maps into $\ell_{-}^{w}(\mathbb{R}^N) $.
\item [(ii)]  If $F$ is of class $C ^r(\mathbb{R} ^N \times  \mathbb{R} ^n)$, $r\geq	§1 $, suppose that 
\begin{equation}
\label{differentiability condition local higher}
L_{F,r}:=\sup_{(\mathbf{x}, {\bf z}) \in \mathbb{R}^N \times  \mathbb{R}^n}\left\{\vertiii{D^r F (\mathbf{x}, {\bf z})}\right\}<+ \infty. 
\end{equation} 
and let $w' $  be any  weighting sequence such that 
\begin{equation}
\label{condition on bounding sequence}
c_{w', w ^r}=\sup _{t \in \mathbb{Z}_{-}} \left\{\frac{w' _{-t}}{w _{-t}^r}\right\}< + \infty.
\end{equation}
Then the map $\mathcal{F} $ is a functor between the  sets
\begin{equation*}
\mathcal{F}:  \ell_{-}^{w}(\mathbb{R}^N) \times V _n \subset \ell_{-} ^{w}(\mathbb{R}^N)\oplus \ell_{-} ^{w}(\mathbb{R}^n)
\longrightarrow  \ell_{-}^{w'}(\mathbb{R}^N)
\end{equation*}
and is differentiable of order $r$ and of class $C^{r-1}( \ell_{-}^{w}(\mathbb{R}^N) \times V _n) $. Moreover,
\begin{equation}
\label{bound for derivative scriptf}
\vertiii{D ^r \mathcal{F}(\mathbf{x}, {\bf z})}_{w, w'}\leq L_{F,r}L _w^r c_{w', w ^r}, \mbox{ for all $(\mathbf{x}, {\bf z}) \in \ell_{-}^{w}(\mathbb{R}^N) \times V _n$} 
\end{equation}
and the map $D ^{r-1} \mathcal{F}: \ell_{-}^{w}(\mathbb{R}^N) \times V _n \longrightarrow L ^{r-1}(\ell_{-} ^{w}(\mathbb{R}^n)\oplus \ell_{-} ^{w}(\mathbb{R}^N), \ell_{-} ^{w'}(\mathbb{R}^N))$ is Lipschitz continuous with Lipschitz constant $L_{F,r}L _w^r c_{w', w ^r}$.
\item [(iii)] The linear map $D _x\mathcal{F}(\mathbf{x}^0, {\bf z} ^0): (\ell_{-}^{w}(\mathbb{R}^N), \left\|\cdot \right\|_{w})\longrightarrow \left(\ell_{-}^{w}(\mathbb{R}^N), \left\|\cdot \right\|_{w}\right)$ is a contraction with constant $L_{F _x}(\mathbf{x}^0, {\bf z}^0) L _w<1 $.
\end{description}
These results also hold when the spaces $\left(\ell_{-} ^{w}(\mathbb{R}^n), \left\|\cdot \right\|_w\right) $ and $\left(\ell_{-} ^{w}(\mathbb{R}^N), \left\|\cdot \right\|_w\right) $ are replaced by $\left(\ell_{-} ^{\infty}(\mathbb{R}^n), \left\|\cdot \right\|_\infty\right)$ and $\left(\ell_{-} ^{\infty}(\mathbb{R}^N), \left\|\cdot \right\|_\infty\right)$, respectively. In that case, the statement is obtained by taking as the sequences $w$ and $w' $ the constant sequence $w ^ \iota$ given by $w ^{\iota} _t:=1 $, for all $t \in \mathbb{N} $.  The inequality \eqref{bound for derivative scriptf} holds true with $L _w=c_{w', w ^r} =1$. 
\end{lemma}

\noindent\textbf{Proof of the lemma.\ \ (i)} Notice first that, as we pointed out in \eqref{script f as product}, and using the notation in Lemma \ref{cr using product maps},  
\begin{equation}
\label{script f as product bis}
\mathcal{F}=\prod _{t \in \mathbb{Z}_{-}} F_t, \quad \mbox{where} \quad F _t:= F \circ p _t \circ  \left(T _1\times {\rm id}_{V _n}\right):  \ell_{-}^{w}(\mathbb{R}^N) \times V _n\longrightarrow  \mathbb{R}^N.
\end{equation}
Also, the hypothesis \eqref{differentiability condition local}, the mean value theorem, and the convexity of the set 
$$
p _t \circ(T _{1} \times {\rm id}_{V _n}) \left( \ell_{-}^{w}(\mathbb{R}^N) \times V _n\right)
$$ 
imply that $F$ is a Lipschitz function with constant $L_{F}$. A development identical to \eqref{mathcalF is lipschitz1} guarantees that the maps $F _t $
are Lipschitz and that $L_{F}L _w/ w _{-t}   $ is a Lipschitz constant of $F _t $, $t \in \mathbb{Z}_{-}$. Given that the sequence $c _{\mathcal{F}}:=({L_{F} L _w}/{w _{-t}})_{t \in \mathbb{Z}_{-}}$ is such that $\left\|c _{\mathcal{F}}\right\|_w =L_{F} L _w<+ \infty$ and $\mathcal{F}= \prod _{t \in \mathbb{Z}_{-}} F_t $, the part {\bf (ii)} of Lemma \ref{cr using product maps} guarantees that $\mathcal{F} $ is Lipschitz continuous and that $L_{F} L _w $ is a Lipschitz constant of $\mathcal{F} $.

Since by hypothesis the reservoir system has a solution $(\mathbf{x}^0, {\bf z}^0) \in \ell_{-}^{w}(\mathbb{R}^N)  \times V _n $, we have that $\mathcal{F}(\mathbf{x}^0, {\bf z}^0)=\mathbf{x}^0 \in  \ell_{-}^{w}(\mathbb{R}^N) $. This implies that 
$\mathcal{F} $ maps into  $\ell_{-}^{w}(\mathbb{R}^N) $ since the Lipschitz condition that we just proved shows  that for any $(\mathbf{x}, {\bf z}) \in   \ell_{-}^{w}(\mathbb{R}^N)  \times V _n$
\begin{equation*}
\|\mathcal{F}(\mathbf{x}, {\bf z})\| _w \leq 
L_{F}L _w\left\|(\mathbf{x}, {\bf z})-(\mathbf{x}^0, {\bf z}^0)\right\|_{w\oplus w}+\|\mathcal{F}(\mathbf{x}^0, {\bf z}^0 )\| _w,
\end{equation*}
which shows that $\|\mathcal{F}(\mathbf{x}, {\bf z})\| _w <+ \infty $ and hence that $\mathcal{F}(\mathbf{x}, {\bf z}) \in \ell_{-}^{w}(\mathbb{R}^N)$.

\medskip

\noindent {\bf (ii)} The expression \eqref{script f as product bis}, the chain rule, the finiteness of $L _w $, and the linearity of $p _t $ and $T _1  $ imply that for any $(\mathbf{x}, {\bf z}) \in  \ell_{-}^{w}(\mathbb{R}^N) \times V _n$:
\begin{equation}
\label{expression derivative in components}
D^rF _t(\mathbf{x}, {\bf z}) = D^rF (\mathbf{x}_{t-1}, {\bf z}_t)\circ  \left(p _t \circ(T _{1} \times {\rm id}_{V _n}), \ldots, p _t \circ(T _{1} \times {\rm id}_{V _n})\right): \left(\ell_{-} ^{w}(\mathbb{R}^N)\oplus \ell_{-} ^{w}(\mathbb{R}^n)\right)^r\longrightarrow  \mathbb{R}^N.
\end{equation}
We now prove \eqref{bound for derivative scriptf}. Notice first that for $\mathbf{u}=\left(\mathbf{u}^1, \ldots, \mathbf{u}^r\right)= \left((\mathbf{u}^1_{\mathbf{x}}, \mathbf{u}^1_{\mathbf{z}}), \ldots, (\mathbf{u}^r_{\mathbf{x}}, \mathbf{u}^r_{\mathbf{z}})\right) \in \left(\ell_{-} ^{w}(\mathbb{R}^N)\oplus \ell_{-} ^{w}(\mathbb{R}^n)\right)^r$ we can write using \eqref{differentiability condition local} and Lemma \ref{smooth projections and time delays}:
\begin{multline*}
\left\|D^rF _t(\mathbf{x}, {\bf z})\cdot \mathbf{u}\right\|=
\left\|D^rF  (\mathbf{x}_{t-1}, {\bf z}_t)\circ  \left(p _t \circ(T _{1} \times {\rm id}_{V _n})(\mathbf{u}^1), \ldots, p _t \circ(T _{1} \times {\rm id}_{V _n})(\mathbf{u}^r)\right)\right\|\\\leq
\vertiii{D^rF  (\mathbf{x}_{t-1}, {\bf z}_t)} \left\|p _t \circ(T _{1} \times {\rm id}_{V _n})(\mathbf{u}^1)\right\| \cdots \left\|p _t \circ(T _{1} \times {\rm id}_{V _n})(\mathbf{u}^r)\right\|\\\leq
\frac{L_{F,r}}{ w _{-t} ^r}\left\|\left(T _1(\mathbf{u}_{\mathbf{x}} ^1), \mathbf{u}_{\mathbf{z}} ^1\right)\right\|_{w\oplus w} \cdots \left\|\left(T _1(\mathbf{u}_{\mathbf{x}} ^r), \mathbf{u}_{\mathbf{z}} ^r\right)\right\|_{w\oplus w}\\\leq 
\frac{L_{F,r}}{ w _{-t} ^r} \left(\left\|T _1(\mathbf{u}_{\mathbf{x}} ^1)\right\| _w+\left\| \mathbf{u}_{\mathbf{z}} ^1\right\|_{w}\right) \cdots
\left(\left\|T _1(\mathbf{u}_{\mathbf{x}} ^r)\right\| _w+\left\| \mathbf{u}_{\mathbf{z}} ^r\right\|_{w}\right)\\\leq 
\frac{L_{F,r}}{ w _{-t} ^r} \left(L _w\left\|\mathbf{u}_{\mathbf{x}} ^1\right\| _w+\left\| \mathbf{u}_{\mathbf{z}} ^1\right\|_{w}\right) \cdots
\left(L _w\left\|\mathbf{u}_{\mathbf{x}} ^r\right\| _w+\left\| \mathbf{u}_{\mathbf{z}} ^r\right\|_{w}\right)\\\leq 
\frac{L_{F,r} L _w ^r }{ w _{-t} ^r}\left( \left\|\mathbf{u}_{\mathbf{x}} ^1\right\| _w+\left\| \mathbf{u}_{\mathbf{z}} ^1\right\|_{w}\right) \cdots
\left( \left\|\mathbf{u}_{\mathbf{x}} ^r\right\| _w+\left\| \mathbf{u}_{\mathbf{z}} ^r\right\|_{w}\right)=
\frac{L_{F,r} L _w ^r }{ w _{-t} ^r} \left\|\mathbf{u}^1\right\|_{w\oplus w} \cdots \left\|\mathbf{u}^r\right\|_{w\oplus w},
\end{multline*}
which shows that 
\begin{equation}
\label{inie with fT}
\vertiii{D^rF _t(\mathbf{x}, {\bf z})} _w\leq \frac{L_{F,r} L _w ^r }{ w _{-t} ^r}. 
\end{equation}
Since, as we saw in part {\bf (i)}  $\mathcal{F}$ maps into $\ell_{-}^{w}(\mathbb{R}^N)  $,   and by Lemma \ref{inclusions with lw} $\ell_{-}^{w}(\mathbb{R}^N) \subset  \ell_{-}^{w^r}(\mathbb{R}^N)$, then $\mathcal{F}$ also maps into $\ell_{-}^{w^r}(\mathbb{R}^N)  $. Additionally, since the sequence $c ^r:= \left(L_{F,r} L _w ^r / w _{-t} ^r\right)_{t \in \mathbb{Z}_{-}}$ is such that $\left\|c ^r\right\|_{w ^r}=L_{F,r} L _w ^r <+ \infty $, the part {\bf (iii)} of Lemma \ref{cr using product maps} guarantees that the map 
\begin{equation*}
\mathcal{F}:  \ell_{-}^{w}(\mathbb{R}^N) \times V _n
\longrightarrow  \ell_{-}^{w^r}(\mathbb{R}^N)
\end{equation*}
is differentiable of order $r$ and that 
\begin{equation}
\label{bound drr with wr}
\vertiii{D^r \mathcal{F}(\mathbf{x}, {\bf z})}_{w,w ^r}\leq L_{F,r} L _w ^r <+ \infty. 
\end{equation}
This argument can be reproduced with the power sequence $w ^r $ replaced by any other sequence $w' $ that satisfies \eqref{condition on bounding sequence}, in which case, it is easy to see that $\ell_{-}^{w^r}(\mathbb{R}^N) \subset \ell_{-}^{w'}(\mathbb{R}^N) $, and we can conclude the differentiability of the map $\mathcal{F}:  \ell_{-}^{w}(\mathbb{R}^N) \times V _n
\longrightarrow  \ell_{-}^{w'}(\mathbb{R}^N) $ for which the relation \eqref{bound drr with wr} is replaced by 
\begin{equation}
\label{bound drr with wr1}
\vertiii{D^r \mathcal{F}(\mathbf{x}, {\bf z})}_{w,w'}\leq L_{F,r} L _w ^r c_{w', w ^r}<+ \infty. 
\end{equation}
The rest of the statement is a consequence of part {\bf (iii)} of Lemma \ref{cr using product maps} applied in this setup.

\medskip

\noindent {\bf (iii)} A computation similar to the one that was used to establish \eqref{expression derivative in components} leads to the following expression for the partial derivatives $D_x\mathcal{F} $ of $\mathcal{F} $: 
\begin{equation}
\label{expression partial derivative in components}
D^r_x\mathcal{F}(\mathbf{x}, {\bf z})=\prod _{t \in \mathbb{Z}_{-}}D^r_xF _t(\mathbf{x}, {\bf z})= \prod _{t \in \mathbb{Z}_{-}} D^r_xF  (\mathbf{x}_{t-1}, {\bf z}_t)\circ  \left(p _t \circ T _{1}, \ldots, p _t \circ T _{1}  \right).
\end{equation}
Using this expression for $r= 1 $  and Lemma \ref{smooth projections and time delays} we can write, for any $\mathbf{u} \in \ell_{-}^{w}(\mathbb{R}^N) $,
\begin{multline*}
\left\|D_x \mathcal{F}(\mathbf{x}^0, {\bf z} ^0) \cdot \mathbf{u}\right\|_w
= \sup_{t \in \mathbb{Z}_{-}}\left\{
\|D_xF  (\mathbf{x}_{t-1}^0, {\bf z}_t^0)\circ  \left(p _t \circ T _{1}\right)(\mathbf{u})\|w _{-t}
\right\}\\
\leq L_{F _x}(\mathbf{x}^0, {\bf z}^0)\sup_{t \in \mathbb{Z}_{-}}\left\{
\vertiii{p _t}_{w} \| T _{1}(\mathbf{u})\|_ww _{-t}
\right\}\\
\leq L_{F _x}(\mathbf{x}^0, {\bf z}^0)\sup_{t \in \mathbb{Z}_{-}}\left\{
\frac{1}{w _{-t}} \| T _{1}(\mathbf{u})\|_ww _{-t}
\right\}
\leq L_{F _x}(\mathbf{x}^0, {\bf z}^0) L _w \left\|\mathbf{u}\right\|_w,
\end{multline*}
as required. $\blacktriangledown $

\medskip

We now proceed with the proof of the theorem in which we obtain the persistence result as a consequence of the Implicit Function Theorem and of the Lemma \ref{preparation with 1k} that we just proved. Using the same notation as in that result we define the map
\begin{equation*}
\begin{array}{cccc}
{\cal G} :&  \ell_{-}^{w}(\mathbb{R}^N) \times \ell_{-}^{w}(\mathbb{R}^n) & \longrightarrow &  \ell_{-}^{w}(\mathbb{R}^N)\\
	&(\mathbf{x}, {\bf z})&\longmapsto & \mathcal{F}(\mathbf{x}, {\bf z})- {\bf x},
\end{array}
\end{equation*}
or equivalently, ${\cal G}= \mathcal{F}- \pi_N $, where $\pi_N: \ell_{-}^{w}(\mathbb{R}^N) \times \ell_{-}^{w}(\mathbb{R}^n) \longrightarrow  \ell_{-}^{w}(\mathbb{R}^N)$ is just the projection onto the first factor.

Notice that by construction and the hypothesis on the point $(\mathbf{x}^0, {\bf z}^0)$ we have that
\begin{equation}
\label{first condition IFT}
{\cal G}(\mathbf{x}^0, {\bf z}^0)={\bf 0}.
\end{equation}
Since the projection $\pi_N $ is linear and by Lemma \ref{preparation with 1k} $\mathcal{F} $ is Lipschitz continuous and differentiable of order $1$,  then so is ${\cal G}= \mathcal{F}- \pi_N $. This implies in particular that the partial derivative $D_x\mathcal{G}(\mathbf{x}^0, {\bf z}^0): \ell_{-} ^{w}(\mathbb{R}^N) \longrightarrow  \ell_{-} ^{w}(\mathbb{R}^N) $ is a bounded operator that we now set to prove that it is an isomorphism. We proceed in two stages that show how the hypotheses in the statement of the theorem imply that this linear map is both injective and surjective. 

\medskip

\noindent {\bf The partial derivative $D_x\mathcal{G}(\mathbf{x}^0, {\bf z}^0):  \ell_{-} ^{w}(\mathbb{R}^N) \longrightarrow  \ell_{-} ^{w}(\mathbb{R}^N) $ is injective.} Notice first that,
\begin{equation*}
D_x\mathcal{G}(\mathbf{x}^0, {\bf z}^0) \cdot  \mathbf{u} = D_x\mathcal{F}(\mathbf{x}^0, {\bf z}^0)\cdot \mathbf{u}- \mathbf{u},\quad \mbox{for any $\mathbf{u} \in \ell_{-} ^{w}(\mathbb{R}^N)  $}.
\end{equation*}
Consequently, the points $ \mathbf{u} \in \ell_{-} ^{w}(\mathbb{R}^N) $ such that $D_x\mathcal{G}(\mathbf{x}^0, {\bf z}^0)\cdot \mathbf{u}= {\bf 0}$ coincide with the fixed points of the map $D_x\mathcal{F}(\mathbf{x}^0, {\bf z}^0):  \ell_{-} ^{w}(\mathbb{R}^N) \longrightarrow  \ell_{-} ^{w}(\mathbb{R}^N) $. Since by part {\bf (iii)} of Lemma \ref{preparation with 1k} $D_x\mathcal{F}(\mathbf{x}^0, {\bf z}^0) $ is a contracting linear map in $\ell_{-} ^{w}(\mathbb{R}^N) $ it has hence only zero  as unique fixed point and  the claim follows.

\medskip

\noindent {\bf The partial derivative $D_x\mathcal{G}(\mathbf{x}^0, {\bf z}^0):  \ell_{-} ^{w}(\mathbb{R}^N) \longrightarrow  \ell_{-} ^{w}(\mathbb{R}^N) $ is surjective.} We prove that for any $\mathbf{v} \in \ell_{-} ^{w}(\mathbb{R}^N) $ there exists $\mathbf{u} \in \ell_{-} ^{w}(\mathbb{R}^N) $ such that $D_x\mathcal{G}(\mathbf{x}^0, {\bf z}^0) \cdot \mathbf{u}= \mathbf{v} $. By the definition of $\mathcal{F} $ in \eqref{definition of fcal} and the expression of its partial derivative in \eqref{expression partial derivative in components}, this equation is equivalent to the recursions, 
\begin{equation}
\label{recursion to be solved}
\mathbf{v} _t= D_xF(\mathbf{x} ^0_{t-1}, {\bf z}_t^0)\cdot \mathbf{u}_{t-1}- \mathbf{u} _t, \quad \mbox{for all  $t \in \mathbb{Z}_{-}$.}
\end{equation}
This equation has a unique solution given by the series
\begin{equation}
\label{series with solution}
\mathbf{u}_t=- \mathbf{v} _t+\sum_{j=1}^{\infty}D_xF(\mathbf{x}^0_{t-1}, {\bf z}_t^0)\cdot D_xF(\mathbf{x}^0_{t-2}, {\bf z}_{t-1}^0)\cdots D_xF(\mathbf{x}^0_{t-j}, {\bf z}_{t-j+1}^0)(-\mathbf{v}_{t-j}), \quad t \in \mathbb{Z}_{-}.
\end{equation}
Indeed, it is straightforward to show that \eqref{series with solution} satisfies \eqref{recursion to be solved}. It remains then to be shown that the sequence $\mathbf{u} $ determined by \eqref{series with solution} belongs to $\ell_{-} ^{w}(\mathbb{R}^N) $. In order to do so we first show that the series in \eqref{series with solution} is convergent by proving that for any $t \in \mathbb{Z}_{-}$, the sequence $\left\{S _n\right\}_{n \in \mathbb{N}^+} $ defined by 
\begin{equation}
\label{to be proved s is convergent}
S _n:= \sum_{j=1}^{n}D_xF(\mathbf{x}^0_{t-1}, {\bf z}_t^0)\cdot D_xF(\mathbf{x}^0_{t-2}, {\bf z}_{t-1}^0)\cdots D_xF(\mathbf{x}^0_{t-j}, {\bf z}_{t-j+1}^0)(-\mathbf{v}_{t-j})w _{-t}, 
\end{equation} 
is a Cauchy sequence. This is so because for any $m,n \in \mathbb{N}^+ $, $m\geq n $,
\begin{multline}
\label{ineqs for surjectivity}
\left\|S _m- S _n\right\|=\\
\left\|\sum_{j=n+1}^{m}D_xF(\mathbf{x}^0_{t-1}, {\bf z}_t^0)\cdot D_xF(\mathbf{x}^0_{t-2}, {\bf z}_{t-1}^0)\cdots D_xF(\mathbf{x}^0_{t-j}, {\bf z}_{t-j+1}^0)(-\mathbf{v}_{t-j}) \right\|w_{-(t-j)}\frac{w_{-(t-j+1)}}{w_{-(t-j)}} \cdots \frac{w_{-t}}{w_{-(t-1)}}\\
\leq \sum_{j=n+1}^{m}\vertiii{D_xF(\mathbf{x}^0_{t-1}, {\bf z}_t^0)}\cdot \vertiii{D_xF(\mathbf{x}^0_{t-2}, {\bf z}_{t-1}^0)}\cdots \vertiii{D_xF(\mathbf{x}^0_{t-j}, {\bf z}_{t-j+1}^0)}\left\|-\mathbf{v}_{t-j}\right\| w_{-(t-j)}L _w^j\\
\leq\sum_{j=n+1}^{m} L_{F _x}(\mathbf{x}^0, {\bf z}^0)^j L _w^j \left\|\mathbf{v}\right\|_w= \frac{\left(L_{F _x}(\mathbf{x}^0, {\bf z}^0) L _w\right)^{n+1}-\left(L_{F _x}(\mathbf{x}^0, {\bf z}^0) L _w\right)^{m+1}}{1-L_{F _x}(\mathbf{x}^0, {\bf z}^0) L _w}\left\|\mathbf{v} \right\|_w,
\end{multline}
which can be made as small as we want because the sequence $\{\left(L_{F _x}(\mathbf{x}^0, {\bf z}^0) L _w\right)^j\}_{j \in \mathbb{N}^+ }$ is convergent and hence Cauchy due to the hypothesis $L_{F _x}(\mathbf{x}^0, {\bf z}^0) L _w<1 $. This implies that $\{S _n\}_{n \in \mathbb{N}^+ }$ is convergent and hence so is the series that defines $\mathbf{u} _t  $ in \eqref{series with solution}.

It remains to be shown that the sequence $\mathbf{u}:=(\mathbf{u} _t)_{t \in \mathbb{Z}_{-}}$ defined by \eqref{series with solution} is an element of $\ell_{-}^{w}(\mathbb{R}^N) $. Following the same strategy that we used to construct the inequalities \eqref{ineqs for surjectivity} it is easy to see that
\begin{equation*}
\left\|\mathbf{u} _t\right\|w _{-t}\leq \frac{1}{1-L_{F _x}(\mathbf{x}^0, {\bf z}^0) L _w} \left\|\mathbf{v} \right\|_w, \mbox{\, for all $t \in \mathbb{Z}_{-}$.} 
\end{equation*}
Consequently,
\begin{equation*}
\left\|\mathbf{u}\right\|_w=\sup_{t \in \mathbb{Z}_{-}}\left\{\left\|\mathbf{u}_t\right\|w _{-t}\right\}\leq \frac{1}{1-L_{F _x}(\mathbf{x}^0, {\bf z}^0) L _w} \left\|\mathbf{v} \right\|_w< + \infty,
\end{equation*}
as required.

\medskip

\noindent {\bf The partial derivative $D_x\mathcal{G}(\mathbf{x}^0, {\bf z}^0):  \ell_{-} ^{w}(\mathbb{R}^N) \longrightarrow  \ell_{-} ^{w}(\mathbb{R}^N) $ is a linear homeomorphism.} This fact is a consequence of the Banach Isomorphism Theorem (see for instance \cite{mta}) that states that any continuous linear isomorphism of Banach spaces has necessarily a continuous inverse.

\medskip

Using all the facts that we just proved, we can invoke the the Implicit Function Theorem as formulated in \cite[page 671]{Schechter:Handbook} (see also \cite{Eecke:1974}) to show the existence of two  open neighborhoods $\widetilde{V}_{\mathbf{x}^0} $ and $\widetilde{V}_{\mathbf{z}^0} $ of $\mathbf{x} ^0 $ and $\mathbf{z} ^0 $  in $ \ell_{-}^{w}(\mathbb{R}^N)$ and  $\ell_{-}^{w}(\mathbb{R}^n)$, respectively, and a unique Lipschitz continuous map   $
\widetilde{U^F}:(\widetilde{V}_{\mathbf{z}^0}, \left\|\cdot \right\|_w) \longrightarrow (\widetilde{V}_{\mathbf{x}^0}, \left\|\cdot \right\|_w)
$
that is differentiable at $\mathbf{z}^0 $ and  satisfies 
\begin{equation*}
{\cal G}(\widetilde{U^F}({\bf z}), {\bf z})={\bf 0},  \mbox{\,for all ${\bf z} \in \widetilde{V}_{\mathbf{z}^0}$}, 
\end{equation*}
which is equivalent to $\mathcal{F}(\widetilde{U^F}({\bf z}), {\bf z})=U^F({\bf z}) $. In view of the identities \eqref{relation function filter U} this means, in other words, that $\widetilde{U^F} $ is the unique reservoir filter with inputs in $\widetilde{V}_{\mathbf{z}^0}$ associated to the reservoir system determined by $F$. This filter is clearly causal and its Lipschitz continuity implies that it has  the fading memory property.  

We conclude the proof by showing that the filter $\widetilde{U^F} $ can be extended to a time-invariant filter $U^F $ defined on the time-invariant saturations $ {V}_{\mathbf{x}^0} $ and $ {V}_{\mathbf{z}^0} $ of the sets  $\widetilde{V}_{\mathbf{x}^0} $ and $\widetilde{V}_{\mathbf{z}^0} $, respectively, and that has the properties listed in the statement. Indeed, define
\begin{equation*}
{V}_{\mathbf{x}^0}:=\bigcup_{t \in \mathbb{Z}_{-}}T _{-t}\left(\widetilde{V}_{\mathbf{x}^0} \right) \quad \mbox{and} \quad
{V}_{\mathbf{z}^0}:=\bigcup_{t \in \mathbb{Z}_{-}}T _{-t}\left(\widetilde{V}_{\mathbf{z}^0} \right).
\end{equation*}
The sets $ {V}_{\mathbf{x}^0} $ and $ {V}_{\mathbf{z}^0} $ are by construction time-invariant and open by the openness of the maps $T _{-t} $ that we established in part {\bf (ii)} of Lemma \ref{smooth projections and time delays}. Define now the map $U^F: {V}_{\mathbf{z}^0} \longrightarrow {V}_{\mathbf{x}^0} $ as
\begin{equation}
\label{definition extension}
U^F\left(T _{-t}({\bf z})\right):= T _{-t} \left(\widetilde{U^F}({\bf z})\right), \quad \mbox{for some} \quad t \in \mathbb{Z}_{-} \mbox{ and } {\bf z} \in \widetilde{V}_{\mathbf{z}^0}. 
\end{equation}
We first show that $U^F  $ is well-defined and time-invariant. Let $t _1, t _2 \in \mathbb{Z}_{-} $ and ${\bf z}_1, {\bf z}_2 \in \widetilde{V}_{\mathbf{z}^0} $ be such that $T _{-t _1}({\bf z}_1)=T _{-t _2}({\bf z}_2) $. Let us now show that
\begin{equation}
\label{well define UF}
U^F \left(T _{-t _1}({\bf z}_1)\right)=U^F \left(T _{-t _2}({\bf z}_2)\right).
\end{equation}
Indeed, for any $t \in \mathbb{Z}_{-} $, the definition \eqref{definition extension} and the causality of $\widetilde{U^F }$ imply that
\begin{multline*}
\left(U^F \left(T _{-t _1}({\bf z}_1)\right)\right)_t= \left(T _{-t_1} \left(\widetilde{U^F}({\bf z}_1)\right)\right)_t\\
=
\widetilde{U^F}({\bf z}_1)_{t+ t _1}=
\widetilde{U^F}({\bf z}_2)_{t+ t _2}=\left(T _{-t_2} \left(\widetilde{U^F}({\bf z}_2)\right)\right)_t=\left(U^F \left(T _{-t _2}({\bf z}_2)\right)\right)_t,
\end{multline*}
which proves \eqref{well define UF}. The time-invariance of $U^F$, as defined in \eqref{definition extension}, is straightforward.

We conclude by showing that $U^F $  is differentiable at all the points of the form $T _{-t}(\mathbf{z}^0) $, $t \in \mathbb{Z}_{-} $ and that it is locally  Lipschitz continuous on $V_{\mathbf{z}^0} $. Since differentiability is a local property, it suffices to prove this property for the restriction of $U^F $   to  open sets. Before we do that, we note that since by part {\bf (ii)} of Lemma \ref{smooth projections and time delays} the map $T _{-t}: \widetilde{V}_{{\bf z}^0}\longrightarrow T _{-t}\left(\widetilde{V}_{\mathbf{z}^0} \right) $ is a submersion, the Local Onto Theorem (see \cite[Theorem 3.5.2]{mta}) guarantees that for  ${\bf z} ':=T _{-t}({\bf z} ^0)\in T _{-t}\left(\widetilde{V}_{\mathbf{z}^0} \right) $ there exists an open neighborhood $V_{{\bf z} '}\subset  T _{-t}\left(\widetilde{V}_{\mathbf{z}^0} \right) $ and a smooth section $\sigma_{{\bf z}'}:V_{{\bf z} '}\longrightarrow\widetilde{V}_{\mathbf{z}^0} $ of $T _{-t}  $ that  satisfies that 
\begin{equation}
\label{expression of section}
\sigma_{{\bf z}'}({\bf z}')= {\bf z} ^0 \quad \mbox{ and} \quad T _{-t} \circ \sigma_{{\bf z}'}= {\rm id}_{V_{{\bf z} '}}.
\end{equation}
The section $\sigma_{{\bf z}'} $ allows us to write down the restriction $U^F|_{V_{{\bf z} '}} $ of $U^F $ to the open subset $V_{{\bf z} '} $ as
\begin{equation}
\label{local expression uff}
U^F|_{V_{{\bf z} '}}({\bf z})=T _{-t}\circ \widetilde{U^F} \left(\sigma_{{\bf z}'}({\bf z})\right), \quad \mbox{for all} \quad {\bf z} \in V_{{\bf z} '}.
\end{equation}
This is so because by \eqref{expression of section} we have that ${\bf z}=T _{-t} \left(\sigma_{{\bf z}'}({\bf z})\right)$, with $\sigma_{{\bf z}'}({\bf z}) \in \widetilde{V}_{\mathbf{z}^0} $, as well as by \eqref{definition extension}. Consequently, since by \eqref{local expression uff} the restriction $U^F|_{V_{{\bf z} '}} $ is a composition of Lipschitz continuous functions then so is $U^F|_{V_{{\bf z} '}} $. The differentiability of $U^F $ at the point ${\bf z}'=T _{-t}({\bf z} ^0) $ can also be concluded using \eqref{local expression uff} by invoking the differentiability of $T _{-t} $ and $\sigma_{{\bf z}'} $ on their domains and the differentiability of $\widetilde{U^F } $ at $\sigma_{{\bf z}'}({\bf z}')= {\bf z}^0 $.
\quad $\blacksquare$

\subsection{Proof of Theorem \ref{characterization of reservoir differentiability}}

\noindent\textbf{(i)} We start with a lemma that shows how condition \eqref{persistence condition global} guarantees the existence of a globally defined filter $U^F: (\ell_{-}^{w}(\mathbb{R}^n), \left\|\cdot \right\|_w) \longrightarrow  (\ell_{-}^{w}(\mathbb{R}^N), \left\|\cdot \right\|_w)$.

\begin{lemma}
\label{contraction condition with lfx}
Let $F:  \mathbb{R}^N \times  {\Bbb R}^n \longrightarrow  \mathbb{R}^N$  be a reservoir map of class $C ^1(\mathbb{R}^N \times  {\Bbb R}^n)$ and let $w$ be a weighting sequence with finite inverse decay ratio $L _w$.
The reservoir map $F$ is a contraction on the first entry if and only if
\begin{equation}
\label{existence global condition contraction}
L_{F _x} <1.
\end{equation}
Moreover, whenever conditions  \eqref{differentiability condition local} and  \eqref{persistence condition global} are satisfied and $(\mathbf{x}^0, {\bf z}^0) \in  (\mathbb{R}^N)^{\mathbb{Z}_{-}} \times  ({\Bbb R}^n) ^{\mathbb{Z}_{-}} $  is a solution of the reservoir system determined by $F$, then there exists a unique causal, time-invariant, and fading memory filter $
U^F: (\ell_{-}^{w}(\mathbb{R}^n), \left\|\cdot \right\|_w) \longrightarrow  (\ell_{-}^{w}(\mathbb{R}^N), \left\|\cdot \right\|_w).
$
\end{lemma}
 
\noindent\textbf{Proof of the lemma.\ \ }
We first show that $F$ is a contraction on the first entry if and only if $L_{F _x} <1 $. Suppose first that $F$ is a contraction with contraction rate $0<c<1 $. Then for any $(\mathbf{x}, {\bf z}) \in  \mathbb{R}^N \times  \mathbb{R}^n$ and any $\mathbf{u} \in \mathbb{R}^N $, the partial derivative $D_xF (\mathbf{x}, {\bf z}) : \mathbb{R}^N \longrightarrow \mathbb{R} ^N$ satisfies that
\begin{equation*}
\left\|D_xF (\mathbf{x}, {\bf z}) \cdot \mathbf{u}\right\|=\lim _{t \rightarrow 0}
\frac{\left\|F (\mathbf{x}+ t \mathbf{u}, {\bf z})- F (\mathbf{x}, {\bf z})\right\|}{t}\leq
\lim _{t \rightarrow 0}
\frac{c t\left\|\mathbf{u}\right\|}{t}= c \left\|\mathbf{u}\right\|,
\end{equation*}
\label{def script lfx}
which implies that $\vertiii{D_xF (\mathbf{x}, {\bf z})}\leq c $ and hence
\begin{equation*}
L_{F _x}:=\sup_{(\mathbf{x}, {\bf z}) \in D_N \times  D_n}
\left\{\vertiii{D_xF (\mathbf{x}, {\bf z})}\right\}\leq c<1.
\end{equation*}
Conversely, suppose that $L_{F _x} <1 $. Since $F$ is of class $C ^1(\mathbb{R}^N \times  \mathbb{R}^n)$, the mean value theorem guarantees that for any $(\mathbf{x} ^1, {\bf z}), (\mathbf{x} ^2, {\bf z}) \in \mathbb{R}^N \times  \mathbb{R}^n $:
\begin{equation*}
\left\|F(\mathbf{x} ^1, {\bf z})-F(\mathbf{x} ^2, {\bf z})\right\|\leq \sup_{(\mathbf{x}, {\bf z}) \in \mathbb{R}^N \times  \mathbb{R}^n}
\left\{\vertiii{D_xF (\mathbf{x}, {\bf z})}\right\}\left\|\mathbf{x} ^1-\mathbf{x} ^2\right\|=L_{F _x}\left\|\mathbf{x} ^1-\mathbf{x} ^2\right\|,
\end{equation*}
and $F$ is hence is a contraction on the first entry.

Suppose now that conditions  \eqref{differentiability condition local} and \eqref{persistence condition global} are satisfied and that $(\mathbf{x}^0, {\bf z}^0) \in  (\mathbb{R}^N)^{\mathbb{Z}_{-}} \times  ({\Bbb R}^n) ^{\mathbb{Z}_{-}} $  is a solution of the reservoir system determined by $F$. Notice first that since $L _w> 1$ then the condition \eqref{persistence condition global} implies that $L_{F _x}<1 $, necessarily, and hence, as we just proved, $F$ is a contraction on the first entry with constant $L_{F _x}$. 
Additionally, as \eqref{differentiability condition local} is satisfied, the mean value theorem
implies that $F$ is Lipschitz continuous with constant $L_{F}$. 
All these facts allow us to invoke part {\bf (ii)} of Theorem \ref{characterization of fmp unbounded} to conclude the existence of the filter $U^F  $ in the statement, since in this situation, the condition \eqref{fmp condition on w} coincides with \eqref{persistence condition global}.  $\blacktriangledown $

\medskip

The proof of the first part of the theorem can now be obtained by applying Theorem \ref{Persistence of the ESP and FMP properties} to each point of the form 
\begin{equation*}
(U^F({\bf z}),{\bf z}) \in \ell_{-}^{w}(\mathbb{R}^N) \times \ell_{-}^{w}(\mathbb{R}^n)
\end{equation*}
for which, according to its statement, there exist  open neighborhoods $V_{U^F({\bf z}) } $ and $V_{\mathbf{z} } $ of $U^F({\bf z})$ and $\mathbf{z} $  in $ \ell_{-}^{w}(\mathbb{R}^N)$ and $ \ell_{-}^{w}(\mathbb{R}^n)$, as well as a unique locally defined causal reservoir filter 
$
\widetilde{U} _F:V_{\mathbf{z} }  \longrightarrow V_{U^F({\bf z}) }
$ associated to $F$. The uniqueness feature implies that $\widetilde{U} _F= U^F|_{V_{\mathbf{z} }}$. Moreover, since $\widetilde{U} _F $  is differentiable at $\mathbf{z} $ and we can repeat this construction for any point ${\bf z} \in \ell_{-}^{w}(\mathbb{R}^n) $ we can conclude that $U^F  $ is differentiable at any point in $\ell_{-}^{w}(\mathbb{R}^n) $.

Finally, the Lipschitz continuity on $\ell_{-}^{w}(\mathbb{R}^n)$ of $U^F $ is a consequence of the mean value theorem, the inequality \eqref{norm duU}, and the fact that 
\begin{equation*}
\sup_{{\bf z} \in \ell_{-}^{w}(\mathbb{R}^n)}\left\{\vertiii{D U^F ({\bf z})}_w\right\}\leq \sup_{{\bf z} \in \ell_{-}^{w}(\mathbb{R}^n)}\left\{
\frac{L_{F _z}(U^F({\bf z}), {\bf z})}{1-L_{F _x}(U^F({\bf z}), {\bf z}))L _w}\right\}\leq  \frac{L_{F _z}}{1-L_{F _x}L _w},
\end{equation*}
which proves \eqref{lip constant uf}.

\medskip

\noindent {\bf (ii)} First of all, the existence of the filter $
U^F: V _n \longrightarrow  \ell_{-}^{w}(\mathbb{R}^N)$ and its differentiability at ${\bf z}^0 \in V _n $ imply that for any $\mathbf{u} \in \ell_{-}^{w}(\mathbb{R}^n) $ and $t \in \mathbb{Z}_{-} $ it satisfies \eqref{relation function filter U} as well as \eqref{recursions U2}, that is,
\begin{equation*}
(D U^F ({\bf z}^0 )\cdot \mathbf{u}) _t=D_xF \left(U^F({\bf z}^0 )_{t-1}, {\bf z }^0  _t\right)\cdot  \left(DU^F ({\bf z }^0 )\cdot \mathbf{u}\right) _{t-1} +D_{z}F(U({\bf z }^0 )_{t-1}, {\bf z }^0  _t)\cdot  \mathbf{u} _t.
\end{equation*}
This identity  can be rewritten in terms of operators on sequences as
\begin{equation*} 
D U^F ({\bf z }^0 )=\left(\prod_{t \in \mathbb{Z}_{-}}D_xF \left(U^F({\bf z }^0 )_{t-1}, {\bf z }^0  _t\right)\right) \circ T _1\circ   DU^F ({\bf z }^0 )+ \prod_{t \in \mathbb{Z}_{-}} D_z F \left(U^F({\bf z }^0 )_{t-1}, {\bf z }^0  _t\right),
\end{equation*}
or equivalently as
\begin{equation} 
\label{expression derivative as operator}
\left(\mathbb{I}_{\ell_{-}^{w}(\mathbb{R}^N)}- \left(\prod_{t \in \mathbb{Z}_{-}}D_xF \left(U^F({\bf z }^0 )_{t-1}, {\bf z }^0  _t\right)\right) \circ T _1\right)D U^F ({\bf z }^0 )= \prod_{t \in \mathbb{Z}_{-}}D_z F \left(U^F({\bf z }^0 )_{t-1}, {\bf z }^0  _t\right).
\end{equation}
This identity determines $D U^F ({\bf z }^0 ) $ that by hypothesis exists if and only if the operator on the left hand side is invertible, which is in turn equivalent to the condition \eqref{converse condition smooth 1}. We finally show that \eqref{converse condition smooth 1} implies \eqref{converse condition smooth 2}. 

We first notice that by Gelfand's formula \cite[page 195]{lax:functional:analysis} the condition \eqref{converse condition smooth 1} is equivalent to
\begin{equation*}
\lim\limits_{k \rightarrow +\infty} 
\vertiii{ \left((\prod_{t \in \mathbb{Z}_{-}}D_xF \left(U^F({\bf z}^0 )_{t-1}, {\bf z} ^0 _t\right)) \circ T _1\right)^k} _w=0.
\end{equation*}
This in turn implies that for any $\mathbf{u}\in \ell_{-}^{w}(\mathbb{R}^n) $, we have that
\begin{equation*}
\lim\limits_{k \rightarrow +\infty} 
\left\| \left((\prod_{t \in \mathbb{Z}_{-}}D_xF \left(U^F({\bf z}^0 )_{t-1}, {\bf z} ^0 _t\right)) \circ T _1\right)^k (\mathbf{u})\right\| _w=0,
\end{equation*}
or, equivalently, that
\begin{equation}
\label{to take sup afterwards}
\lim\limits_{k \rightarrow +\infty} \left(
\sup_{t \in \mathbb{Z}_{-}}\left\{
\left\| \left(D_xF \left(U^F({\bf z}^0 )_{t-1}, {\bf z} ^0 _t\right)\circ \cdots \circ 
D_xF \left(U^F({\bf z}^0 )_{t-k}, {\bf z} ^0 _{t-k+1}\right)
\right)  (\mathbf{u}_{t-k})\right\| 
\right\}\right)=0.
\end{equation}
If we now take vectors $\mathbf{u} \in \ell_{-}^{w}(\mathbb{R}^n)$ in \eqref{to take sup afterwards} of the form $\mathbf{u} _t:= \widetilde{\mathbf{u}}/w _{-t} $, $t \in \mathbb{Z}_{-} $, with $\widetilde{ \mathbf{u} } \in {\Bbb R}^n $ such that $\left\|\widetilde{\mathbf{u}}\right\|=1 $, and we take the supremum in \eqref{to take sup afterwards} with respect to all those vectors $\widetilde{ \mathbf{u}} $, we obtain that
\begin{equation}
\label{to take sup afterwards }
\lim\limits_{k \rightarrow +\infty} \left(
\sup_{t \in \mathbb{Z}_{-}}\left\{
\vertiii{ D_xF \left(U^F({\bf z}^0 )_{t-1}, {\bf z} ^0 _t\right)\circ \cdots \circ 
D_xF \left(U^F({\bf z}^0 )_{t-k}, {\bf z} ^0 _{t-k+1}\right) }\frac{w _{-t}}{w _{-(t-k)}}
\right\}\right)=0.
\end{equation}
Given that 
\begin{multline*}
\vertiii{D_xF \left(U^F({\bf z}^0 )_{-1}, {\bf z} ^0 _0\right)\circ \cdots \circ 
D_xF \left(U^F({\bf z}^0 )_{-k}, {\bf z} ^0 _{-k+1}\right)}\frac{1}{w_k}\\
\leq 
\sup_{t \in \mathbb{Z}_{-}}\left\{
\vertiii{ D_xF \left(U^F({\bf z}^0 )_{t-1}, {\bf z} ^0 _t\right)\circ \cdots \circ 
D_xF \left(U^F({\bf z}^0 )_{t-k}, {\bf z} ^0 _{t-k+1}\right) } \frac{w _{-t}}{w _{-(t-k)}}
\right\},
\end{multline*}
the condition \eqref{converse condition smooth 2} follows. \quad $\blacksquare$

\subsection{Proof of Theorem \ref{Volterra series representation}}

\noindent Since by hypothesis $U$ is analytic in $B _{\left\|\cdot \right\|_w}({{\bf z}} ^0,M) $ then 
\begin{equation}
\label{Taylor for U volterra}
U({\bf z})=U({\bf z} ^0)+\sum_{j=1}^{\infty}\frac{1}{j!}D^jU({\bf z} ^0) (\underbrace{{\bf z}- {\bf z} ^0 ,  \ldots, {\bf z}- {\bf z} ^0}_{\mbox{$j$ times}}), \quad \mbox{for any} \quad {\bf z} \in B _{\left\|\cdot \right\|_w}({{\bf z}} ^0,M).
\end{equation}
We now show that for the elements that satisfy \eqref{condition domain volterra} the series expansion \eqref{Taylor for U volterra} amounts to the discrete-time Volterra series expansion \eqref{Taylor equals volterra}.
Let $m \in \mathbb{Z}_{-}  $ and let $\delta _m \in \ell_{-}^{w}(\mathbb{R}) $ be the sequence defined by
\begin{equation}
\label{delta sequences}
\left(\delta _m\right)_t:=
\left\{
\begin{array}{cl}
\frac{1}{w_{-m}} & \mbox{ if } t=m, \\
0 & \mbox{ otherwise. }
\end{array}
\right.
\end{equation}
Note that $\left\|\delta _m\right\|_w=1 $ for all $m \in \mathbb{Z}_{-}  $. Moreover, for any ${\bf z}\in \ell_{-}^{w}(\mathbb{R}) $ we can write
\begin{equation*}
{\bf z}- {\bf z}^0=\sum_{t \in \mathbb{Z}_{-}}\widetilde{z} _t \delta _t, \quad \mbox{with} \quad \widetilde{z} _t= (z _t- z _t ^0) w _{-t},
\end{equation*}
and hence by the multilinearity of the derivatives $D^jU({\bf z} ^0) ({\bf z}- {\bf z} ^0 ,  \ldots, {\bf z}- {\bf z} ^0) $ and the causality of the filter $U$ we have that
\begin{equation}
\label{terms taylor expansion}
D^jU({\bf z} ^0) ({\bf z}- {\bf z} ^0 ,  \ldots, {\bf z}- {\bf z} ^0)_t= \sum_{m _1=- \infty}^{t} \cdots \sum_{m _j=- \infty}^{t}\widetilde{z}_{m _1}\cdots \widetilde{z}_{m _j}D^jU({\bf z} ^0)(\delta _{m _1}, \ldots, \delta_{m _j})_t,  \mbox{ for all } t \in \mathbb{Z}_{-}.
\end{equation}
We first show that for the elements that satisfy \eqref{condition domain volterra} the sum in the right hand side of \eqref{terms taylor expansion} is finite. Indeed, for any $t \in \mathbb{Z}_{-} $:
\begin{multline}
\left|\sum_{m _1=- \infty}^{t} \cdots \sum_{m _j=- \infty}^{t}\widetilde{z}_{m _1}\cdots \widetilde{z}_{m _j}D^jU({\bf z} ^0)(\delta _{m _1}, \ldots, \delta_{m _j})_t\right|\\
=\left|\sum_{m _1=- \infty}^{t} \cdots \sum_{m _j=- \infty}^{t}\widetilde{z}_{m _1}\cdots \widetilde{z}_{m _j}\frac{1}{w _{-t}}w _{-t}D^jU({\bf z} ^0)(\delta _{m _1}, \ldots, \delta_{m _j})_t\right|\\
\leq \sum_{m _1=- \infty}^{t} \cdots \sum_{m _j=- \infty}^{t}\left|\widetilde{z}_{m _1}\cdots \widetilde{z}_{m _j}\right|\frac{1}{w _{-t}}\|D^jU({\bf z} ^0)(\delta _{m _1}, \ldots, \delta_{m _j})\| _w\\
\leq \sum_{m _1=- \infty}^{t} \cdots \sum_{m _j=- \infty}^{t}\frac{\left|\widetilde{z}_{m _1}\cdots \widetilde{z}_{m _j}\right|}{w _{-t}}\vertiii{D^jU({\bf z} ^0)} _w\|(\delta _{m _1}, \ldots, \delta_{m _j})\| _w\\
=j\frac{\vertiii{D^jU({\bf z} ^0)} _w}{w _{-t}}\sum_{m _1=- \infty}^{t} \cdots \sum_{m _j=- \infty}^{t}\left|\widetilde{z}_{m _1}\right|\cdots \left|\widetilde{z}_{m _j}\right|\\
=
j\frac{\vertiii{D^jU({\bf z} ^0)} _w}{w _{-t}}\left(\sum_{m =- \infty}^{t} \left|{z}_{m }-{z}_{m }^0\right|w _{-m} \right)^j<+ \infty,
\end{multline}
where the last equality is a consequence of, for example, \cite[Theorem 8.44]{Apostol:analysis}, and the last inequality follows from two facts. First, as $U$ is analytic, it is in particular smooth and hence $\vertiii{D^jU({\bf z} ^0)} _w<+ \infty $ for all $j \in \mathbb{Z}_{-}  $. Second, since by hypothesis ${\bf z},{{\bf z}}^0 \in \ell_{-}^{1, w}(\mathbb{R}^n) $ then ${\bf z}-{{\bf z}}^0\in  \ell_{-}^{1, w}(\mathbb{R}^n) $ and hence $\sum_{m =- \infty}^{t} \left|{z}_{m }-{z}_{m }^0\right|w _{-m} <+ \infty $.

We now show that \eqref{Taylor for U volterra} can be rewritten as \eqref{Taylor equals volterra}. Notice first that for any $t,m \in \mathbb{Z}_{-}  $ such that $m\leq t $, the sequences \eqref{delta sequences} satisfy
\begin{equation}
\label{properties delta sequ}
T _{-t} \left(\delta _m\right)= \frac{w_{-(m-t)}}{w_{-m}}\delta _{m-t}.
\end{equation}
Second, the time-invariance of $U$ and of the sequence ${\bf z} ^0 $, imply that for any $j\in {\mathbb{N}^{+}}  $, $t \in \mathbb{Z}_{-} $, and ${\bf z} ^1, \ldots, {\bf z} ^j \in \ell_{-}^{w}(\mathbb{R}) $, we have that 
\begin{equation*}
T _{-t} \left(D^jU({\bf z} ^0) \left({\bf z} ^1, \ldots, {\bf z} ^j\right) \right)=
D^jU(T _{-t} \left({\bf z} ^0\right)) \left(T _{-t} \left({\bf z} ^1\right), \ldots, T _{-t} \left({\bf z} ^j\right)\right) =
D^jU({\bf z} ^0) \left(T _{-t} \left({\bf z} ^1\right), \ldots, T _{-t} \left({\bf z} ^j\right)\right).
\end{equation*}
These two relations imply that for any $t \in \mathbb{Z}_{-}  $
\begin{multline*}
D^jU({\bf z} ^0)(\delta _{m _1}, \ldots, \delta_{m _j})_t= \left(T _{-t}\left(D^jU({\bf z} ^0)(\delta _{m _1}, \ldots, \delta_{m _j})\right)\right)_0=D^jU({\bf z} ^0)(T _{-t}(\delta _{m _1}), \ldots, T _{-t}(\delta_{m _j}))_0\\
=D^jU({\bf z} ^0)(\delta _{m _1-t}, \ldots, \delta_{m _j-t})_0 \frac{w_{-(m _1-t)}}{w_{-m _1}} \cdots\frac{w_{-(m _j-t)}}{w_{-m _j}}.
\end{multline*}
If we substitute this relation in the summands of \eqref{terms taylor expansion}, we obtain that 
\begin{multline}
\label{almost for g}
\widetilde{z}_{m _1}\cdots \widetilde{z}_{m _j}D^jU({\bf z} ^0)(\delta _{m _1}, \ldots, \delta_{m _j})_t\\
=(z_{m _1}-z_{m _1}^0)\cdots (z_{m _j}-z_{m _j}^0)\cdot w_{-(m _1-t)} \cdots w_{-(m _j-t)}\cdot D^jU({\bf z} ^0)(\delta _{m _1-t}, \ldots, \delta_{m _j-t})_0.
\end{multline}
Define now
\begin{multline}
\label{definition of g}
g _j(n _1, \ldots, n _j):=w_{-n _1}\cdots w_{-n _j} \frac{1}{j!}D^jU({\bf z} ^0)(\delta_{n _1}, \ldots, \delta_{n _j})_0\\=
w_{-n _1}\cdots w_{-n _j} \frac{1}{j!}D^jH({\bf z} ^0)(\delta_{n _1}, \ldots, \delta_{n _j})_0
= \frac{1}{j!}D^jH({\bf z} ^0)(e_{n _1}, \ldots, e_{n _j})_0,
\end{multline}
where  $e _m \in \ell_{-}^{w}(\mathbb{R}) $ is the sequence defined in \eqref{definition gs}.
If we make  the change of variables $n _i:=m _i-t $ in \eqref{almost for g}, we use \eqref{definition of g}, and we insert the resulting expression in \eqref{terms taylor expansion} and subsequently in \eqref{Taylor for U volterra} we obtain \eqref{Taylor equals volterra}. 
The uniqueness of this series expansion follows from the same argument as in \cite[Theorem 1]{sandberg:volterra}.

We now prove the error estimates \eqref{error estimation volterra} with the same strategy as in \cite{sandberg:volterra}. Using the Cauchy bounds for analytic functions (see, for instance, the last expression in \cite[page 112]{Hille:Phillips}) and the analyticity hypothesis on $U:B _{\left\|\cdot \right\|_w}({{\bf z}} ^0,M)\subset \ell_{-}^{w}(\mathbb{R})\longrightarrow B _{\left\|\cdot \right\|_w}(U({{\bf z}} ^0),L)\subset \ell_{-}^{w}(\mathbb{R}^N) $, we have that for any $j \in {\mathbb{N}^{+}} $ and $ t \in \mathbb{Z}_{-} $
\begin{equation}
\label{bound for one derivative}
\|D^jU({\bf z} ^0)({\bf z}, \ldots, {\bf z}) _t\|=\|p _t \circ D^jU({\bf z} ^0)({\bf z}, \ldots, {\bf z}) \|\leq \vertiii{p _t}_w\|D^jU({\bf z} ^0)({\bf z}, \ldots, {\bf z}) \| _w\leq \frac{j! L}{w _{-t}} \left(\frac{\left\| {\bf z}\right\|_w}{M}\right)^j,
\end{equation}
where we also used the first part of Lemma \ref{smooth projections and time delays}. Now, as we saw in the previous paragraphs,
\begin{multline}
\label{error estimation volterra proof}
\left\|U({{\bf z}})_t-U ({{\bf z}}^0)_t-\sum_{j=1}^{p}\sum_{m _1=- \infty}^{0} \cdots \sum_{m _j=- \infty}^{0}g _j(m _1, \ldots, m _j)(z_{m _1+t}-z^0_{m _1+t})\cdots (z_{m _j+t}-z^0_{m _j+t})\right\|\\
\leq \left\|U({{\bf z}})_t-U ({{\bf z}}^0)_t-\sum_{j=1}^{p}\frac{1}{j!}D^jU({\bf z} ^0) ({\bf z}- {\bf z} ^0 ,  \ldots, {\bf z}- {\bf z} ^0)\right\|=
\sum_{j=p+1}^{\infty}\frac{1}{j!}D^jU({\bf z} ^0) ({\bf z}- {\bf z} ^0 ,  \ldots, {\bf z}- {\bf z} ^0)\\
\leq \frac{L}{w _{-t}}\sum_{j=p+1}^{\infty} \left(\frac{\left\| {\bf z}\right\|_w}{M}\right)^j
\leq \frac{L}{w _{-t}} \left(1- \frac{\left\|{\bf z}\right\|_w}{M}\right)^{-1} \left(\frac{\left\|{\bf z}\right\|_w}{M}\right)^{p+1},
\end{multline}
where the inequalities in the last line follow from \eqref{bound for one derivative}. \quad $\blacksquare$

\subsection{Time invariance of the solutions of a reservoir system}

The filters studied in this paper are those determined by reservoir systems of the type introduced in~\eqref{reservoir equation}--\eqref{readout}. As we already pointed out, in that case we can associate unique reservoir filters $U ^F  $ and  $U ^F  _h$ to the reservoir map $F$ and the reservoir system, respectively, whenever \eqref{reservoir equation} satisfies the echo state property. In that case, it has been shown in \cite[Proposition 2.1]{RC7} that both $U ^F  $ and  $U ^F  _h$ are necessarily causal and time-invariant. We complement this fact with a similar elementary statement that does not require the echo state property or the existence reservoir filters.

\begin{lemma}
Let $(\mathbf{x} ^0, {\bf z} ^0) \in \left({\Bbb R}^N\right)^{\mathbb{Z}_{-}} \times \left({\Bbb R}^n\right)^{\mathbb{Z}_{-}} $ be a solution of the reservoir system determined by the map $F: {\Bbb R}^N \times {\Bbb R}^n \longrightarrow {\Bbb R}^N  $. Then, for any $\tau \in \mathbb{Z}_{-}  $, the pair $(T _\tau(\mathbf{x} ^0), T _\tau({\bf z} ^0)) \in \left({\Bbb R}^N\right)^{\mathbb{Z}_{-}} \times \left({\Bbb R}^n\right)^{\mathbb{Z}_{-}} $ is also a solution.
\end{lemma}

\noindent\textbf{Proof.\ \ } By hypothesis, for any $t \in \mathbb{Z}_{-} $ we have that
\begin{equation*}
F(\mathbf{x} ^0_{t-1}, {\bf z} ^0 _t)=\mathbf{x} ^0_{t},
\end{equation*}
and hence
\begin{equation*}
F \left(T _\tau(\mathbf{x} ^0)_{t-1}, T _\tau({\bf z} ^0)_t\right)=F(\mathbf{x} ^0_{t- \tau-1}, {\bf z} ^0 _{t- \tau})=\mathbf{x} ^0_{t-\tau}=T _\tau(\mathbf{x} ^0)_{t}, \quad \mbox{as required.} \quad \blacksquare
\end{equation*}

\medskip

\noindent {\bf Acknowledgments:} We thank Lukas Gonon for fruitful discussions. The authors acknowledge partial financial support of the French ANR ``BIPHOPROC" project (ANR-14-OHRI-0002-02) as well as the hospitality of the Centre Interfacultaire Bernoulli of the Ecole Polytechnique F\'ed\'erale de Lausanne during the program ``Stochastic Dynamical Models in Mathematical Finance, Econometrics, and Actuarial Sciences" that made possible the collaboration that lead to some of the results included in this paper. LG acknowledges partial financial support of the Graduate School of Decision Sciences of the Universit\"at Konstanz. JPO acknowledges partial financial support  coming from the Research Commission of the Universit\"at Sankt Gallen and the Swiss National Science Foundation (grant number 200021\_175801/1).

\noindent
\addcontentsline{toc}{section}{Bibliography}
\bibliographystyle{wmaainf}
\bibliography{/Users/JP17/Dropbox/Public/GOLibrary}
\end{document}